\documentclass[10pt,twocolumn,letterpaper]{article}

\usepackage{cvpr}
\usepackage{xcolor}
\usepackage{times}
\usepackage{epsfig}
\usepackage{graphicx}
\usepackage{amsmath}
\usepackage{amssymb}
\usepackage{algorithm}
\usepackage{algpseudocode}
\usepackage{amssymb}
\usepackage{amsthm}
\usepackage{bm}
\usepackage{subfigure}
\usepackage{verbatim}
\usepackage{graphicx}
\usepackage{bbding}
 \usepackage[misc]{ifsym}

\cvprfinalcopy 

\def\cvprPaperID{851} 

\ifcvprfinal\fi

\DeclareMathOperator{\Tr}{Tr}
\newtheorem{prop}{Proposition}
\begin{document}

\title{Joint Cuts and Matching of Partitions in One Graph}
\author{Tianshu Yu\\
Arizona State University\\
85281 Tempe AZ USA\\
{\tt\small tianshuy@asu.edu}
\and
Junchi Yan {$^\text{(\Letter)}$}\\
Shanghai Jiao Tong University\\
IBM Research -- China\\
{\tt\small yanesta13@163.com}
\and
Jieyi Zhao\\
University of Texas at Houston\\
77030 Houston TX USA\\
{\tt\small jieyi.zhao@uth.tmc.edu}
\and
Baoxin Li\\
Arizona State University\\
85281 Tempe AZ USA\\
{\tt\small baoxin.li@asu.edu}
}
\maketitle

\begin{abstract}
As two fundamental problems, graph cuts and graph matching have been investigated over decades, resulting in vast literature in these two topics respectively. However the way of jointly applying and solving graph cuts and matching receives few attention. In this paper, we first formalize the problem of simultaneously cutting a graph into two partitions i.e. graph cuts and establishing their correspondence i.e. graph matching. Then we develop an optimization algorithm by updating matching and cutting alternatively, provided with theoretical analysis. The efficacy of our algorithm is verified on both synthetic dataset and real-world images containing similar regions or structures.
\end{abstract}


\section{Introduction}
\label{sec:intro}
\begin{figure*}[t]
\begin{center}
    \includegraphics[width = 0.195\textwidth]{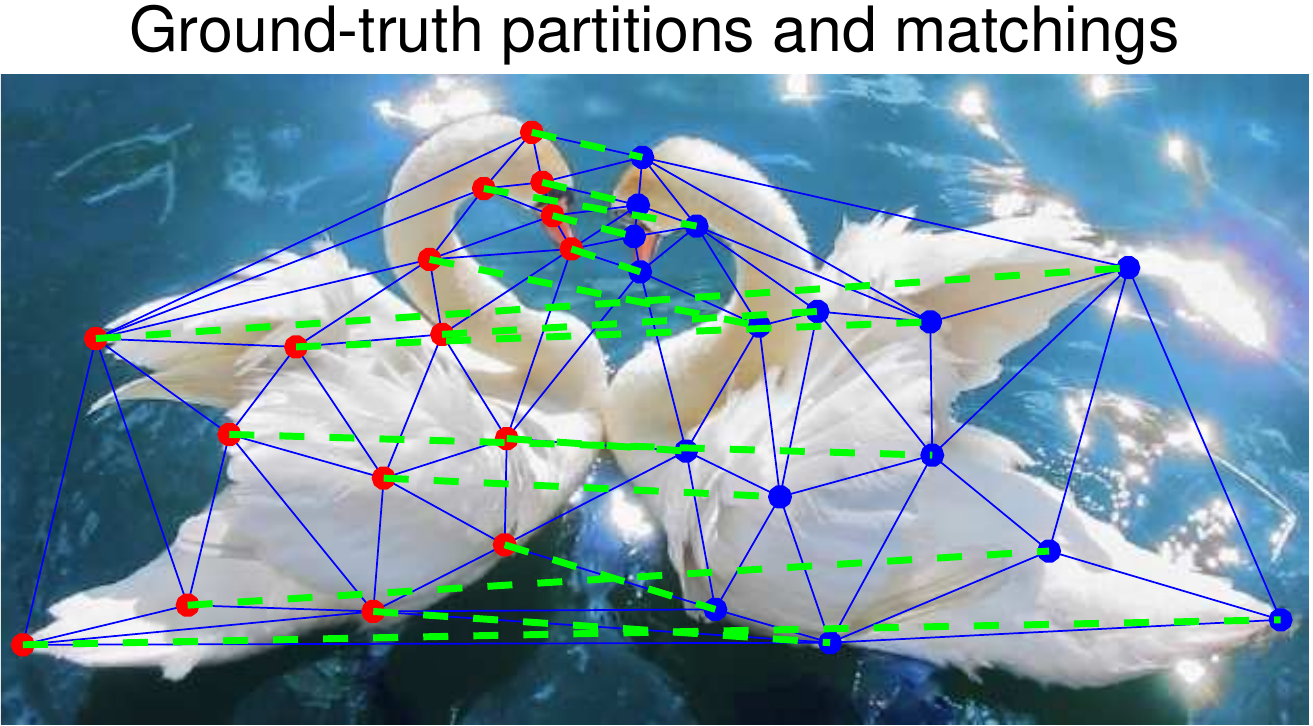}
    \includegraphics[width = 0.195\textwidth]{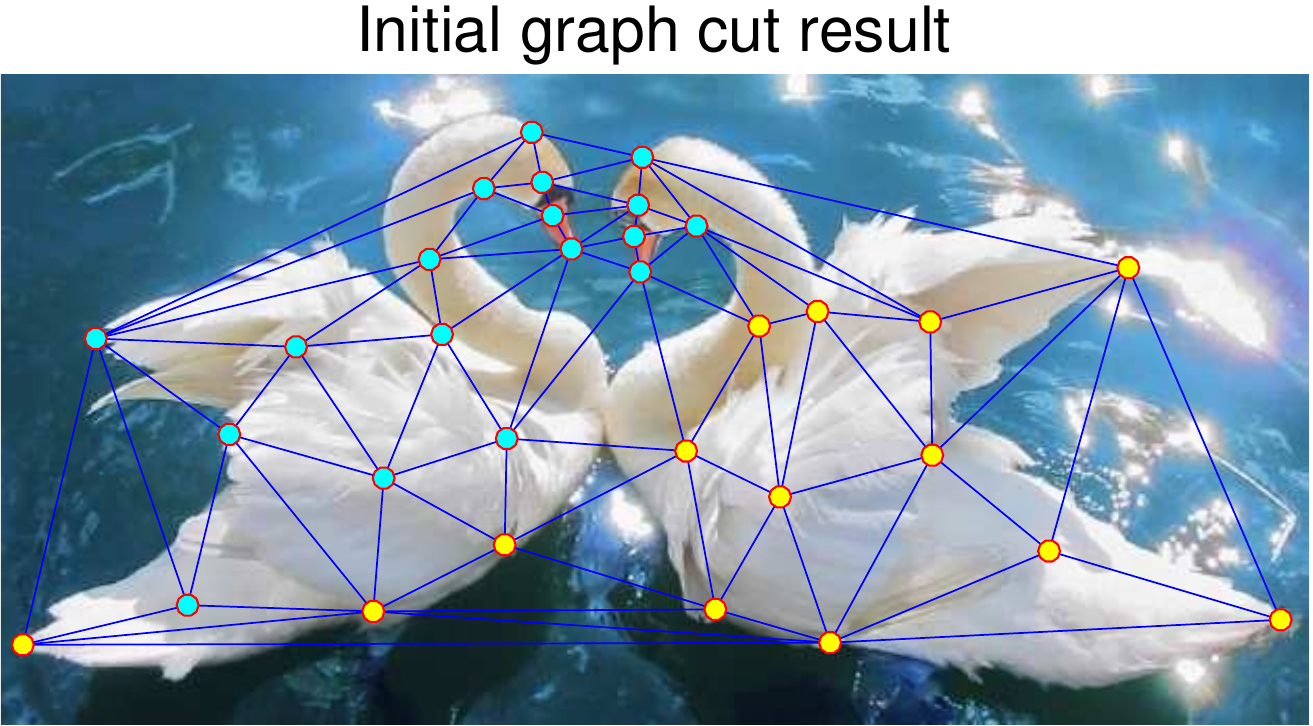}
    \includegraphics[width = 0.195\textwidth]{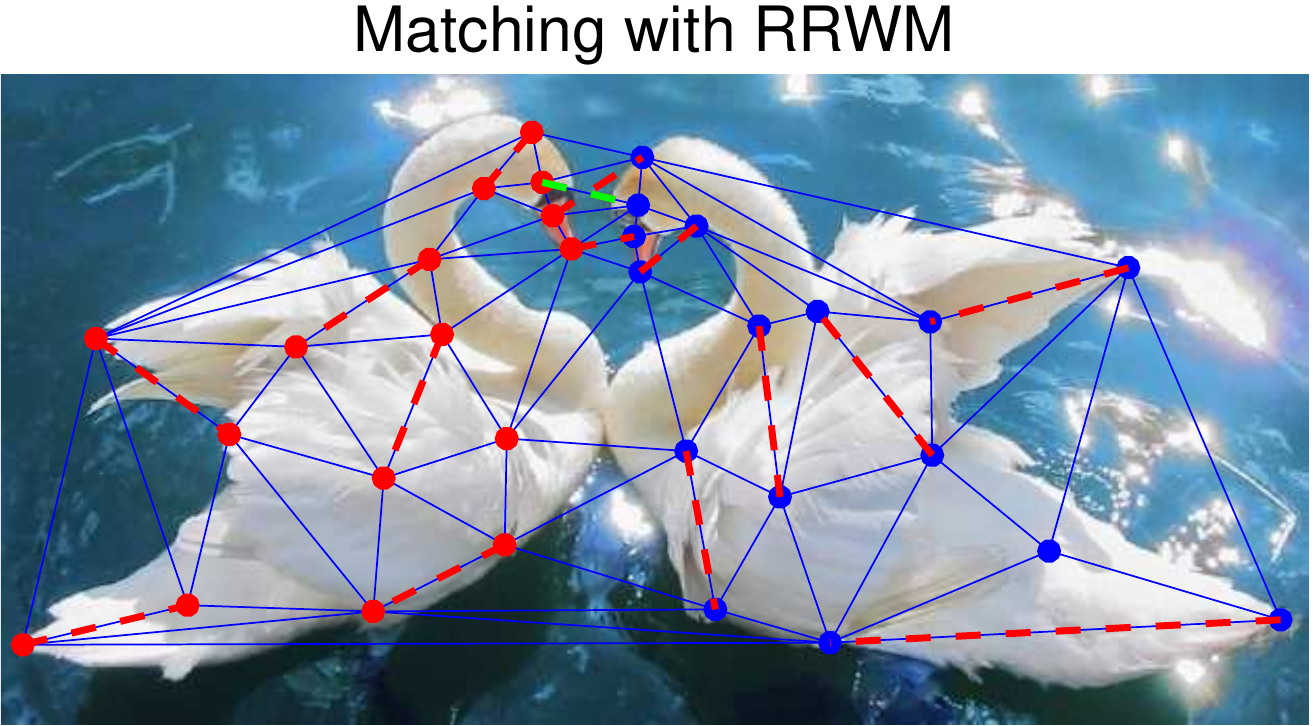}
    \includegraphics[width = 0.195\textwidth]{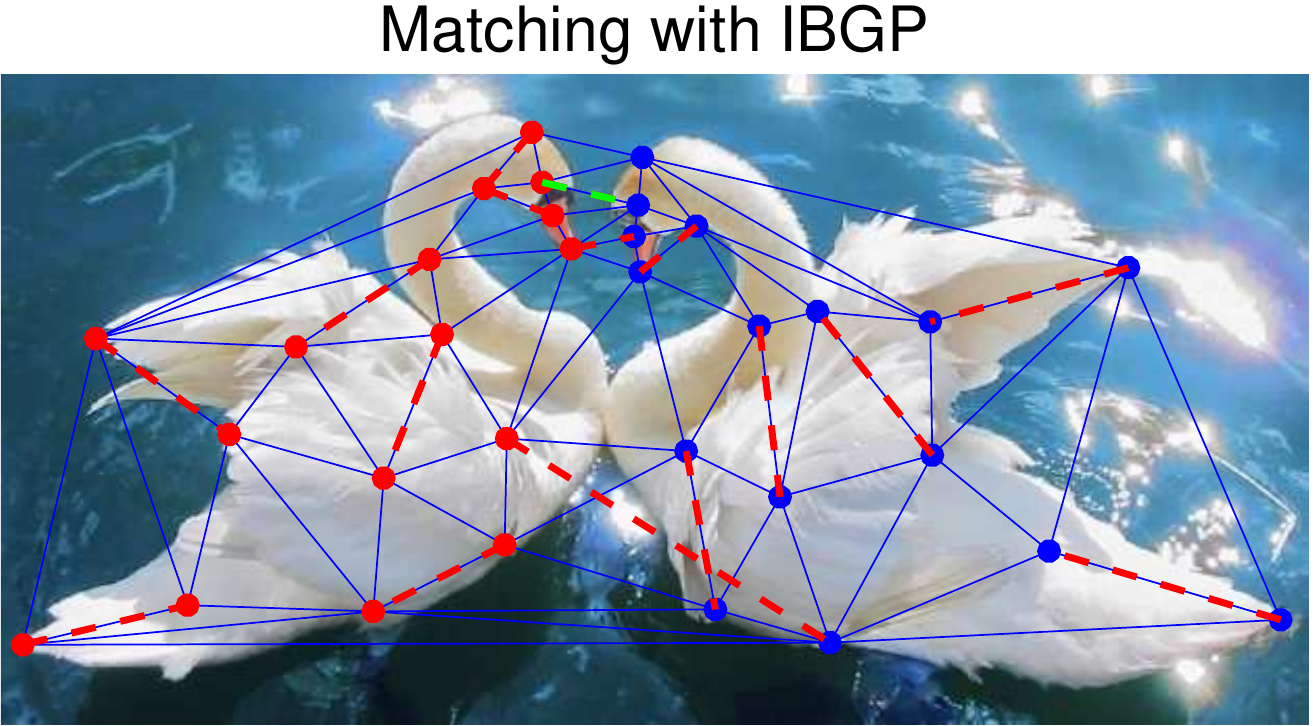}
    \includegraphics[width = 0.195\textwidth]{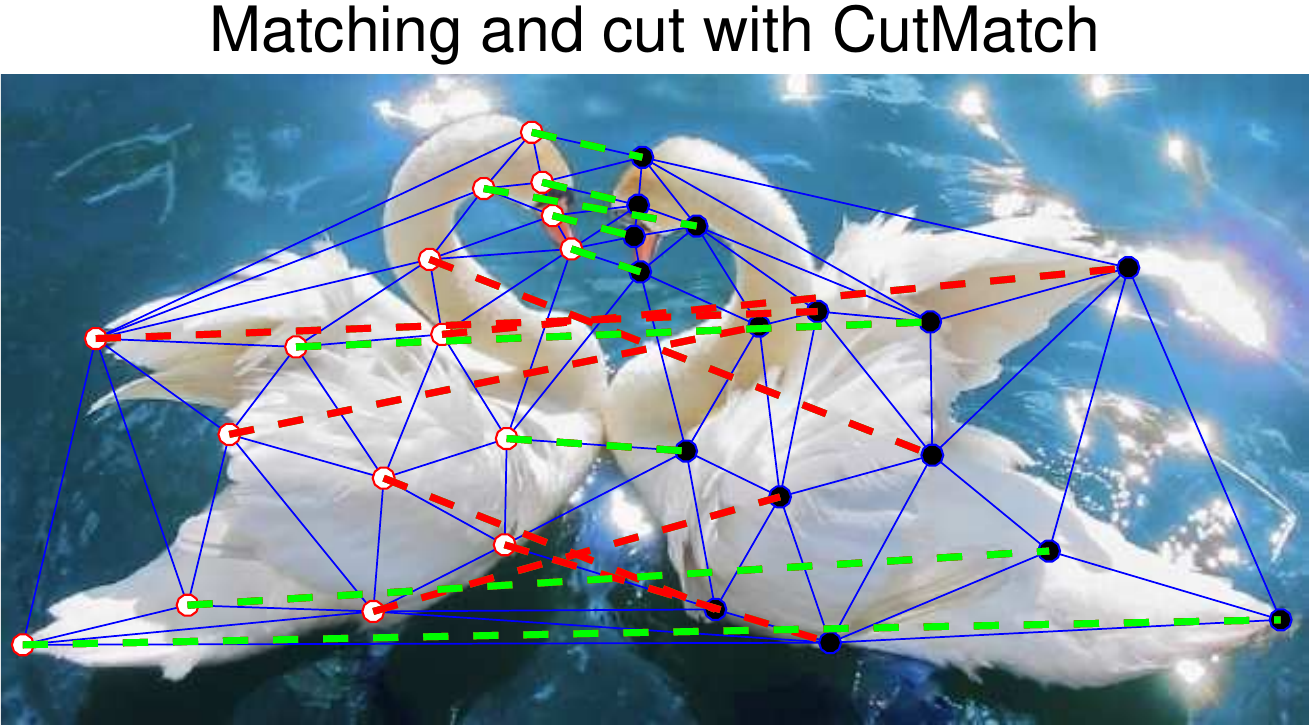}
\end{center}
\vspace{-10pt}
\caption{Comparison of the proposed CutMatch v.s. graph cut and graph matching on image with two swans: \textbf{a)} ground truth matching and partition; \textbf{b)} graph cuts which aims to minimize the overall cut loss (here defined as node distance) tends to group nearby nodes into the same partition (two heads are close to each other); \textbf{c)} (modified) graph matching focuses on find similar nodes based the appearance and local structural similarity while gets completely wrong matchings when the partition is unknown; \textbf{d)} CutMatch combines the best of the two by considering both local closeness and overall correspondences and thus produce the best results for similar object partition from a single image; \textbf{e)} Green (red) dashed lines refer to the matchings that agree (disagree) with the ground-truth. Zoom in for better view.}
\label{fig:cutmatch_example}
\end{figure*}

Over the past few decades, exploration on graphs has brought remarkable advances for computer vision. Images often have strong structural correlation and are modeled as connected graphs with nodes and edges. While nodes correspond to specific feature points, edges reveal the spatial relation or interaction between neighboring nodes. In this sense, graph is capable of encoding the local and global structural information, and is a natural representation of visual input. We consider two important tasks associated with graph structure: \textbf{graph cuts (GC)} and \textbf{graph matching (GM)}. Applications of these tasks can be found in stereo \cite{tappen2003comparison}, video synthesis \cite{kwatra2003graphcut}, image segmentation \cite{shi2000normalized} (for graph cut/partition) and object categorization \cite{DuchenneICCV11}, action recognition \cite{YaoECCV12}, and feature matching \cite{TorresaniECCV08}, etc.

\textbf{Graph cuts}, as the name suggests, is a task to cut one graph into two or more partitions, so as to aggregate the nodes with higher similarities and separate the ones without. Conventionally, the edges is associated with flow weights measuring the transportation capacity between end nodes. Thus cutting the graph can be casted as finding a collection of edges, whose end nodes form a bipartite separation such that the overall transportation on these edges is minimized. For computer vision, pixels and their neighboring relation are often regarded as nodes and edges, respectively. In this fashion, graph cuts is equivalent to grouping the pixels into two clusters taking into account their appearance and adjacency properties. With different weights or transportation measurements, various objectives are proposed.

\textbf{Graph matching} aims to find correspondence among graphs. Both unary and second-order (or higher-order) affinity are considered for matching. Since graph matching (GM) typically involves structural information beyond unary/point-wise similarity, it is in general robust against noise and local ambiguities. However, compared with linear assignment with polynomial-time global optimum solvers such as the Hungarian method \cite{munkres1957algorithms}, GM in general involves quadratic assignment which is known NP-hard.

This paper is motivated primarily by the interest of exploring how the above two key tasks on graphs may be jointly considered so as to attain certain optimality not possible with each task considered independently. There are real applications of image segmentation where graph cut alone would have difficulty, while GM seems to be a natural part of the problem, and hence a joint approach is called for.

Fig. \ref{fig:cutmatch_example} illustrates the key potential benefits of such a joint approach (more explanations in the caption): (balanced) graph cuts \cite{andreev2004balanced} fails to produce correct partition for the image containing two swans, even though the size balance between the partition is considered. This can be explained by the fact that GC encourages the clustered points (also called local connectivity in this paper) to be assigned to the same partition\footnote{We follow a traditional setting by using the point distance as the cut cost whereby the goal is to minimize the overall cut cost.}. On the other hand, if we enforce existing graph matching solvers such as reweighed random walk (RRWM) \cite{cho2010reweighted} or our proposed IBGP (a side product of our main method, see more details in Sect. 4) with a brute-force modification to make its input compatible with only one input graph, the result would be completely meaningless.

We draw important observation from the above example: In many scenarios, separating identical/similar objects in an image requires a new mechanism to assign two correspondence nodes (e.g., those similar in appearance and local structure) into two different partitions. This conflicts with the general assumption of graph cut whereby similar nodes should be grouped into the same partition. Importantly, we posit that such a misalignment can be to some extent alleviated by adopting graph matching as it encourages similar nodes to be assigned into separate partitions by finding their correspondence.  In general, by combining cut and matching, one task may be facilitated through the extra information/constraint provided by the other (e.g., nearby nodes shall be with the same group while two nodes with similar local structure should be with different partitions), and thus algorithms may be developed to achieve cut and matching alternately in reaching a balanced optima. This inspires us to take a joint approach to achieve the best of the two worlds. To our best knowledge, this is a new problem that has not been addressed.

\textbf{Contribution} This paper makes two-fold contributions: 1) a novel model for jointly incorporating and solving graph cuts and graph matching in a unified framework; 2) a novel approach for the joint cut and matching problem whose key component is an effective optimization algorithm named Iterative Bregman Gradient Projection (IBGP). IBGP is theoretically ensured to find a stationary solution. One side-product of IBGP is its simplified version that can be adapted as a new approach for standard two-graph matching.

\section{Related Work}
\textbf{Graph partitions and cuts} Graph partition and cuts have been a long-standing research topic. The general idea is to obtain high similarity among nodes within partition and low across partitions. The minimum weight $k$-cut problem aims to divide the graph into $k$ disjoint non-empty partitions such that the cut metric is minimized, with no balance constraint enforced. As shown in \cite{goldschmidt1994polynomial}, this problem can be solved optimally in $O(n^{k^2})$ when $k$ is given. Without knowing $k$, the problem becomes NP-complete. When the (roughly) equal size constraint is involved, early work have proved it is NP-complete. Extended theoretical study is presented in \cite{andreev2004balanced}. To limit the scope, we focus in this paper on two-cut partition and forgo a more generalized treatment of (relaxed) $k$-cut problem for future study.

Apart from the above general theoretical studies, more efficient algorithms are devised in practice. To the computer vision community, graph cut based approaches are popular due to their empirically observed promising performance as well as solid theoretical foundations \cite{kolmogorov2004energy}.

\textbf{Graph matching} Graph matching is mostly considered in the two-graph setting. Since the problem can in general be reformulated as a quadratic assignment task which is NP-hard, different approximate solvers \cite{gold1996graduated,leordeanu2009integer,cho2010reweighted} are devised with empirical success. A line of work \cite{PachauriNIPS13,YanICCV13,YanECCV14,ChenICML14,ShiCVPR16} moves from the two-graph setting to a collection of graphs. While all these work assume each graph is known and focus on finding the node correspondences among graphs. There are also end-to-end approaches \cite{ChoCVPR12,CollinsECCV14} on joint node detection (in their respective candidate sets) and matching in an iterative fashion. However, none of these work involves graph partition as they still assume the two graphs are separate in advance. Another orthogonal area is graph structure learning \cite{CaetanoPAMI09,ChoICCV13} whereby the edge weights on each separate graph is learned to improve the matching accuracy. Readers are referred to \cite{YanICMR16} for a comprehensive survey.

Though there have been a large amount of literature on graph cuts and graph matching, we are unable to identify any prior work for combing the both lines of research.
\section{Problem Formulation}
\label{sec:problem}
\textbf{Notations} Lower-case bold $\mathbf{x}$ and upper-case bold $\mathbf{X}$ represent a vector and a matrix, respectively. Lower-case letter such as $n$ corresponds to scalar. Specifically $\mathbf{x}_i$ and $\mathbf{X}_{ij}$ returns the scalar values of $i$-th element of $\mathbf{x}$ and element $(i,j)$ of $\mathbf{X}$, respectively. Calligraphic letters $\mathcal{G}$ and $\mathcal{E}$ represent set of nodes and edges. $\mathbb{R}$ and $\mathbb{S}^{+}$ denote the real-number domain and $n$-th order non-negative symmetric matrices. Function $\mathbf{K}=\text{diag}(\mathbf{x})$ spans a vector into a matrix such that $\mathbf{K}_{ij}=\mathbf{x}_i$ if $i=j$, and $\mathbf{K}_{ij}=0$ otherwise. Denote $\mathbf{I}$ and $\mathbf{O}$ ($\mathbf{0}$) the identity matrix and all-zeros matrix (vector), respectively. $[\cdot]_+$ is the element-wise ramp function, which keeps the input if it is positive, and $0$ otherwise.

\textbf{Preliminaries on graph cuts} Consider graph $\mathcal{G}$ associated with weight matrix $\mathbf{W}\in\mathbb{R}^{n\times n}$, where $n$ is the number of nodes in the graph and $\mathbf{W}_{ij}$ corresponds to the weight (similarity) assigned to edge $(i,j)\in\mathcal{E}$. Graph cuts amounts to finding binary partition $\{\mathcal{P}_i\}_{i\in\{1,2\}}$ with $\mathcal{P}_1\cup\mathcal{P}_2=\mathcal{N}$ and $\mathcal{P}_1\cap\mathcal{P}_2=\emptyset$, such that the energy $\sum_{i\in\mathcal{P}_1,j\in\mathcal{P}_2}\mathbf{W}_{ij}$ is minimized. By introducing variable $\mathbf{y}\in\{-1,1\}^n$ as indication vector, such that $\mathbf{y}_i=-1$ if $i\in\mathcal{P}_1$ and $\mathbf{y}_i=1$ for $i\in\mathcal{P}_2$, this problem can be casted as minimizing $\sum_{i,j}\mathbf{W}_{ij}(\mathbf{y}_i-\mathbf{y}_j)^2$ \cite{shi2000normalized}. This energy function encourages the node pair $(i,j)$ belonging to the same partition if the corresponding edge weight is large.

Optimizing this energy is NP-hard. Hence continuous relaxation is often adopted by letting $\mathbf{y}\in[-1,1]^n$. A popular reformulation of this problem in the context of spectral theory is $\mathbf{y}^T \mathbf{L}^g\mathbf{y}=\sum_{i,j}\mathbf{W}_{ij}(\mathbf{y}_i-\mathbf{y}_j)^2$, where $\mathbf{L}^g=\mathbf{D}-\mathbf{W}$ and $\mathbf{D}=\text{diag}(\sum_{i}\mathbf{W}_i)$ \cite{shi2000normalized}. Here $\mathbf{L}^g$ is conventionally called graph Laplacian in terms of weight $\mathbf{W}$. This reformulation yields the following problem:
\begin{small}
\begin{equation}
\label{eq:cut_original}
\min_{\mathbf{y}}\frac{\mathbf{y}^T\mathbf{L}^g\mathbf{y}}{\mathbf{y}^T\mathbf{y}}
\end{equation}
\end{small}
where the imposed normalization $\lVert\mathbf{y}\rVert^2_2=1$ is encoded into the objective by dividing the denominator $\mathbf{y}^T\mathbf{y}$. Graph cuts based methods such as Ratio Cut \cite{wei1989towards} and NCut (normalized cut) \cite{shi2000normalized} are popular solvers to this problem, which are devised based on the Rayleigh-Ritz theorem \cite{NLA97}.

\textbf{Adapting standard graph matching to one input graph setting} Standard, namely conventional graph matching (GM) usually involves two given graphs for establishing their node correspondence. In this paper, we consider a more challenging case aiming to find matching between two \emph{implicit} partitions with equal size $m$ from a whole graph of size $n=2m$\footnote{We use the term `implicit' to denote the partitions are unknown before matching, and leave the more ill-posed and challenging case of unequal sizes of partitions and matching for future work.}. Consider a graph $\mathcal{G}=\langle\mathcal{N},\mathcal{E}\rangle$ with $n=2m$ nodes, where $\mathcal{N}$ and $\mathcal{E}$ are the nodes and edges, respectively. We establish the node-to-node correspondences within this graph using a matrix $\mathbf{X}\in\{0,1\}^{n\times n}$, where $\mathbf{X}_{ij}=1$ implies there is a matching between node $i$ and $j$, and $\mathbf{X}_{ij}=0$ otherwise. We set $\mathbf{X}_{ii}=0$ by eliminating the matching feasibility between a node $i$ and itself. Given matrix $\mathbf{A}\in\mathbb{R}^{n^2\times n^2}$ encoding the first (on diagonal) and second order (off diagonal) affinity, graph matching can be formulated as finding binary solution to maximize the overall score $\text{vec}(\mathbf{X})^T\mathbf{A}\text{vec}(\mathbf{X})$ \cite{YanICMR16}, where $\text{vec}(\mathbf{X})$ is the vectorized replica of $\mathbf{X}$. This problem, however, is notoriously NP-hard with combinatorial complexity. In line with standard graph matching methods, we relax $\mathbf{X}$ into the continuous interval $[0,1]$ and derive the following model:
\begin{small}
\begin{equation}
\label{eq:GM}
\begin{split}
&\max_{\mathbf{X}}\text{vec}\left(\mathbf{X}\right)^T \mathbf{A}\text{vec}\left(\mathbf{X}\right) \\
&\text{s.t. }\sum_{i}\mathbf{X}_{ij}=\mathbf{1},\sum_{j}\mathbf{X}_{ij}=\mathbf{1}^T,\mathbf{X}_{ii}=0,\mathbf{X}\in\mathbb{S}^{+}
\end{split}
\end{equation}
\end{small}
where $\mathbf{1}$ is a vector with all $1$ values. The first two constraints, ensuring $\mathbf{X}$ to be doubly stochastic, indicate one-to-one matching in line with the widely used (relaxed) formulation for two-graph matching \cite{cho2010reweighted}.

In fact, matching two partitions from one input graph is more challenging compared with standard two-graph matching: i) there are additional constraints i.e. $\mathbf{X}_{ii}=0$, $\mathbf{X}\in\mathbb{S}^{+}$ as the model seeks the matching between two partitions from \emph{a single input graph}; ii) the involved variables in Eq. \ref{eq:GM} for single graph's partition matching are of larger size. In fact, when the two partitions are given for standard two-graph matching, the affinity matrix and matching matrix become $\mathbf{A}^{standard}_{gm}\in\mathbb{R}^{m^2\times m^2}$ and $\mathbf{X}^{standard}_{gm}\in\{0,1\}^{m\times m}$ for $n=2m$. Hence the new problem cannot be addressed by existing graph matching solvers e.g. \cite{gold1996graduated,cho2010reweighted,leordeanu2009integer}.

\textbf{Joint cuts and partition matching in one graph} Our model aims to unify cut and matching, to cut a graph into two components and establish their correspondence simultaneously. As discussed in Section \ref{sec:intro} our method is based on the observation that both \textbf{closeness} (\textbf{for graph cuts}) and \textbf{correspondence} (\textbf{for graph matching}), no matter measured by appearance or local structure, shall be considered to find meaningful partitions in particular scenarios, e.g. finding identical objects/structures from an input image. Specifically we assume if $\mathbf{X}_{ij}$ is large for matching, then partition $\mathbf{y}_i\neq\mathbf{y}_j$ is more likely to be true. Similar to the graph cuts objective form, the above discussion can be quantified by $\sum_{i,j}\mathbf{X}_{ij}(\mathbf{y}_i-\mathbf{y}_j)^2$. Moreover the doubly stochastic property guarantees that $\sum_{i}\mathbf{X}_{ij}=\mathbf{1}$ is constant during iterative optimization. Note while graph cuts seeks a minimizer, this term is to be maximized. Hence we must have the Laplacian of matching $\mathbf{I}-\mathbf{X}$, and the coupled energy measuring how much a matching and partition agree with each other becomes:
\begin{small}
\begin{equation}
\label{eq:coupling}
\mathbf{y}^T\left(\mathbf{I}-\mathbf{X}\right)\mathbf{y}
\end{equation}
\end{small}

In summary, adding up Eq. (\ref{eq:cut_original}), (\ref{eq:GM}), (\ref{eq:coupling}) and letting $\mathbf{x}=\text{vec}(\mathbf{X})$ for short, we have the joint objective:
\begin{small}
\begin{gather}
\label{eq:CutMatch}
\max_{\mathbf{X},\mathbf{y}}\mathbf{x}^T\mathbf{A}\mathbf{x}-\lambda_1\mathbf{y}^T\mathbf{L}^g\mathbf{y}+\lambda_2\mathbf{y}^T\left(\mathbf{I}-\mathbf{X}\right)\mathbf{y}\\\notag
\text{s.t. }\sum_{i}\mathbf{X}_{ij}=\mathbf{1},\sum_{j}\mathbf{X}_{ij}=\mathbf{1}^T,\mathbf{X}_{ii}=0, \mathbf{X}\in\mathbb{S}^{+},\lVert\mathbf{y}\rVert^2_2=1
\end{gather}
\end{small}
where $\lambda_1$ and $\lambda_2$ balance the cut energy and strength of coupling, respectively. Denote $\mathcal{X}$ the set of variables satisfying the four constraints in Eq. (\ref{eq:CutMatch}). Though Normalized Cut \cite{shi2000normalized} can produce more balanced partitions, it is not suitable in our case. This is because Laplacian $\lambda_2(\mathbf{I}-\mathbf{X})-\lambda_1\mathbf{L}^g$ is not necessarily positive semidefinite. Thus the normalization matrix may have negative diagonal elements.

\section{Proposed Solver}
We devise an optimization procedure which involves updating graph cuts and matching alternatively.

\textbf{Initialization} There are two variables for initialization: $\mathbf{X}^{\text{init}}$ and $\mathbf{y}^{\text{init}}$. ii) For initial cut $\mathbf{y}^{\text{init}}$, it is obtained by performing Rayleigh Quotient over the graph Laplacian; ii) For initial matching $\mathbf{X}^{\text{init}}$, firstly we employ on-the-shelf graph matching solver e.g. Reweighted Random Walk Matching (RRWM) \cite{cho2010reweighted} to obtain the raw matching $\mathbf{X}^{\text{raw}}\in\{0,1\}^{n\times n}$ with respect to the affinity matrix $\mathbf{A}\in\mathbb{R}^{n^2\times n^2}$, then the symmetry constraint is fulfilled by averaging $\mathbf{X}^{\text{raw}}$ with its transpose: $\mathbf{X}^{\text{raw}}=\frac{\mathbf{X}^{\text{raw}}+\mathbf{X}^{\text{raw}T}}{2}$. Then we continue to set the matchings $\mathbf{X}^{\text{raw}}_{ij}=0$ if $i$ and $j$ are in the same partition according to $\mathbf{y}$. As the obtained matching $\mathbf{X}^{\text{raw}}$ may violate the constraint $\mathbf{X}_{ii}=0$, we further project it onto the convex set $\mathcal{X}$. To this end, we draw inspiration from the Bregmanian Bi-Stochastication algorithm \cite{wang2010learning} and devise a modified version to obtain $\mathbf{X}^{\text{init}}$ satisfying $\mathbf{X}_{ii}=0$. Different from the algorithm in \cite{wang2010learning} to find a bi-stochastic projection, the modified version also seeks to obtain the solution with diagonal elements all $0$s. The modified projection algorithm denoted by $\mathfrak{P}(\cdot)$ is depicted in Algorithm \ref{alg:BregProj}. Note the projection is performed regarding with Euclidian distance. We show how to update $\mathbf{y}$, $\mathbf{X}$ alternatively until converge.

\textbf{Update cuts $\mathbf{y}$} Given current matching $\mathbf{X}^{\text{cur}}$, by peeling off the terms involving $\mathbf{y}$ from Eq. (\ref{eq:CutMatch}), we have:
\begin{small}
\begin{equation}
\max_{\mathbf{y}}\mathbf{y}^T\left\{\lambda_2(\mathbf{I}-\mathbf{X}^{\text{cur}})-\lambda_1(\mathbf{D}-\mathbf{W})\right\}\mathbf{y}
\end{equation}
\end{small}

Optimizing this objective is straightforward. Firstly the graph Laplacian, which is the second term within the curly braces, is positive semidefinite by its definition. For the first term in the curly braces, we notice that as $\mathbf{X}^{\text{cur}}$ is a doubly stochastic matrix with the largest eigenvalue no larger than $1$, then $\mathbf{I}-\mathbf{X}^{\text{cur}}$ must also be positive semidefinite.
Thus solving $\mathbf{y}$ yields to calculate the eigenvector corresponding to the largest non-zero algebraic eigenvalue.

\textbf{Update matching $\mathbf{X}$} One natural idea on updating matching $\mathbf{X}$ is to employ a standard graph matching solver. However, this is not suitable for our case. This fact is caused by the difficulty to encode the linear part coupling cut and matching $\mathbf{y}^T\left(\mathbf{I}-\mathbf{X}\right)\mathbf{y}$ into the quadratic term. Instead we devise a gradient projection based method to obtain stationary solution as described in the following.

First, we compute the partial derivative of $\mathbb{E}$ w.r.t. $\mathbf{X}$:
\begin{small}
\begin{equation}
\label{eq:deriv_x}
[\nabla\mathbb{E}]_{ij}=\left[\left(\mathbf{A}+\mathbf{A}^T\right)\mathbf{x}\right]_{\tau(i,j)}-\lambda_2\left[\mathbf{y}\mathbf{y}^T\right]_{ij}
\end{equation}
\end{small}
where $\tau(i,j)$ is a mapping from matrix subscript $(i,j)$ to its corresponding vector index -- elements in two terms on the right hand side are from vector and matrix respectively.

\begin{algorithm}[tb!]
\begin{small}
\caption{Bregmanian Projection with Zero Constraint}
\label{alg:BregProj}
\begin{algorithmic}[1]
        \Require{$\mathbf{X}^{\text{raw}}$ by a standard GM solver e.g. RRWM \cite{cho2010reweighted}}
        \State $\mathbf{X}=\mathbf{X}^{\text{raw}}$
        \Repeat
            \State $\mathbf{X}\leftarrow\left[\mathbf{X}+\frac{\mathbf{1}\mathbf{1}^T-\mathbf{X}\mathbf{1}\mathbf{1}^T-\mathbf{1}\mathbf{1}^T\mathbf{X}}{n}+\frac{\mathbf{1}\mathbf{1}^T\mathbf{X}\mathbf{1}\mathbf{1}^T}{n^2}\right]_+$
            \State $\mathbf{X}_{jj}\leftarrow 0$
        \Until{Converge}
        \State \Return $\mathbf{X}^{\text{init}}\leftarrow\mathbf{X}$
\end{algorithmic}
\end{small}
\end{algorithm}
\begin{algorithm}[tb!]
\begin{small}
\caption{Iterative Bregman Gradient Projection \textbf{IBGP}}
\label{alg:BregDirection}
\begin{algorithmic}[1]
        \Require{$\mathbf{X}^{\text{prev}}$, $\mathbf{y}^{\text{prev}}$, $\mathbf{A}$, $\epsilon$}
        \Repeat
            \State Compute partial derivative w.r.t. $\mathbf{X}^{\text{prev}}$ by Eq. (\ref{eq:deriv_x})
            \State $\mathbf{V}\leftarrow\frac{\nabla\mathbb{E}+\nabla\mathbb{E}^T}{2}$ // \emph{remove for standard GM}
            \Repeat
                \State $\mathbf{V}\leftarrow\mathbf{V}-\frac{1}{n}\mathbf{V}\mathbf{1}\mathbf{1}^T-\frac{1}{n}\mathbf{1}\mathbf{1}^T\mathbf{V}+\frac{1}{n^2}\mathbf{1}\mathbf{1}^T\mathbf{V}\mathbf{1}\mathbf{1}^T$
                \State $\mathbf{V}_{jj}\leftarrow 0$ // \emph{remove for standard GM}
                \If{\emph{solving CutMatch}}
                    \If{$\mathbf{V}_{jk}<\max\{-\frac{\mathbf{X}^{\text{prev}}_{jk}}{\epsilon},-\frac{\mathbf{X}^{\text{prev}}_{kj}}{\epsilon}\}$}
                        \State $\mathbf{V}_{jk}\leftarrow\max\{-\frac{\mathbf{X}^{\text{prev}}_{jk}}{\epsilon},-\frac{\mathbf{X}^{\text{prev}}_{kj}}{\epsilon}\}$
                    \EndIf
                    \If{$\mathbf{V}_{jk}>\min\{\frac{1-\mathbf{X}^{\text{prev}}_{jk}}{\epsilon},\frac{1-\mathbf{X}^{\text{prev}}_{kj}}{\epsilon}\}$}
                        \State $\mathbf{V}_{jk}\leftarrow\min\{\frac{1-\mathbf{X}^{prev}_{jk}}{\epsilon},\frac{1-\mathbf{X}^{\text{prev}}_{kj}}{\epsilon}\}$
                    \EndIf
                \ElsIf{\emph{solving standard GM}}
                    \If{$\mathbf{V}_{jk}<-\frac{\mathbf{X}^{\text{prev}}_{jk}}{\epsilon}$}
                        \State $\mathbf{V}_{jk}\leftarrow -\frac{\mathbf{X}^{\text{prev}}_{jk}}{\epsilon}$
                    \EndIf
                    \If{$\mathbf{V}_{jk}>\frac{1-\mathbf{X}^{\text{prev}}_{jk}}{\epsilon}$}
                        \State $\mathbf{V}_{jk}\leftarrow\frac{1-\mathbf{X}^{\text{prev}}_{jk}}{\epsilon}$
                    \EndIf
                \EndIf
            \Until{Converge}
            \State $\mathbf{X}^{\text{prev}}\leftarrow\mathbf{X}^{\text{prev}}+\epsilon\mathbf{V}$
        \Until{Converge}
        \State \Return $\mathbf{X}^{\text{cur}}\leftarrow\mathbf{X}^{\text{prev}}$
\end{algorithmic}
\end{small}
\end{algorithm}
Given $\mathbf{X}^{\text{prev}}$ obtained in previous iteration, using gradient $\nabla\mathbb{E}$ can not guarantee the updated $\mathbf{X}^{\text{cur}}$ lying in the feasible convex set $\mathcal{X}$. To address this issue, we develop an objective to find an optimal updating direction:
\begin{small}
\begin{equation}
\label{eq:GradUpdate}
\begin{split}
\min_{\nabla\mathbf{X}}&\lVert\nabla\mathbf{X}-\nabla\mathbb{E}\rVert_F^2 \\
\text{s.t. }&\nabla\mathbf{X}\mathbf{1}=\mathbf{0},\nabla\mathbf{X}=\nabla\mathbf{X}^T,\nabla\mathbf{X}_{ii}=0\\
&\mathbf{X}^{\text{prev}}_{ij}+\epsilon\nabla\mathbf{X}_{ij}\ge 0,\mathbf{X}^{\text{prev}}_{ij}+\epsilon\nabla\mathbf{X}_{ij}\leq 1
\end{split}
\end{equation}
\end{small}
where $\nabla\mathbf{X}$ is the optimal update direction starting from $\mathbf{X}^{\text{prev}}$, and $\epsilon>0$ is a pre-defined step length. The convex objective $\min_{\nabla\mathbf{X}}\lVert\nabla\mathbf{X}-\nabla\mathbb{E}\rVert_F^2$ is to find an optimal ascending direction within the feasible set, which is defined by the constraints in Eq. (\ref{eq:GradUpdate}) claiming the updated $\mathbf{X}^{\text{cur}}=\mathbf{X}^{\text{prev}}+\epsilon\nabla\mathbf{X}$ still falling into $\mathcal{X}$. The verification of this claim is trivial. To obtain the solution to Eq. (\ref{eq:GradUpdate}), we perform \emph{Dykstra Algorithm} \cite{dykstra1983algorithm}, whereby we first split the constraints into two sets $\mathcal{C}_1$ and $\mathcal{C}_2$:
\begin{small}
\begin{equation}
\begin{split}\notag
&\mathcal{C}_1:\nabla\mathbf{X}\mathbf{1}=\mathbf{0},\nabla\mathbf{X}=\nabla\mathbf{X}^T\\\notag
&\mathcal{C}_2:\nabla\mathbf{X}_{ii}=0, \mathbf{X}^{\text{prev}}_{ij}+\epsilon\nabla\mathbf{X}_{ij}\ge 0,\mathbf{X}^{\text{prev}}_{ij}+\epsilon\nabla\mathbf{X}_{ij}\leq 1
\end{split}
\end{equation}
\end{small}
where $\mathcal{C}_1$ is a subspace and $\mathcal{C}_2$ is a bounded convex set. By denoting $\mathbf{V}=\nabla\mathbf{X}$ and $\mathbf{U}=\nabla\mathbb{E}$ for short, we first optimize the objective with constraints in $\mathcal{C}_1$:
\begin{small}
\begin{equation}
\label{eq:Gradient_C1}
\min_{\mathbf{V}}\lVert\mathbf{V}-\mathbf{U}\rVert_F^2, \quad
\text{s.t. } \mathbf{V}\mathbf{1}=\mathbf{0},\mathbf{V}=\mathbf{V}^T
\end{equation}
\end{small}

The Lagrangian function of this problem is:
\begin{small}
\begin{equation}
\mathcal{L}=\Tr(\mathbf{V}^T\mathbf{V}-2\mathbf{V}^T\mathbf{U}+\mathbf{U}^T\mathbf{U})-\bm{\mu}^T\mathbf{V}\mathbf{1}-\bm{\mu}^T\mathbf{V}^T\mathbf{1}
\end{equation}
\end{small}
where $\bm{\mu}\in\mathbb{R}^{n}$ is the corresponding Lagrangian multiplier. There is only one multiplier because $\mathbf{V}$ is symmetric. Taking the partial derivative with respect to $\mathbf{V}$ and letting $\partial\mathcal{L}/\partial\mathbf{V}=0$, we have:
\begin{small}
\begin{equation}
\label{eq:V_multiplier}
\mathbf{V}=\mathbf{U}+\frac{1}{2}\bm{\mu}\mathbf{1}^T+\frac{1}{2}\mathbf{1}\bm{\mu}^T
\end{equation}
\end{small}

After right multiplying $\mathbf{1}$ on both sides and noticing the constraint $\mathbf{V}\mathbf{1}=\mathbf{0}$ in Eq. (\ref{eq:Gradient_C1}), we have:
\begin{small}
\begin{equation}
\mathbf{0}=\mathbf{U}\mathbf{1}+\frac{n}{2}\bm{\mu}+\frac{1}{2}\mathbf{1}\mathbf{1}^T\bm{\mu}
\end{equation}
\end{small}

Applying Woodbury formula \cite{hager1989updating} for the inverse, we can derive a closed form of multiplier:
\begin{small}
\begin{equation}
\label{eq:multiplier_closed}
\bm{\mu}=-\frac{2}{n}\left(\mathbf{I}-\frac{1}{2n}\mathbf{1}\mathbf{1}^T\right)\mathbf{U}\mathbf{1}
\end{equation}
\end{small}

Substituting Eq. (\ref{eq:multiplier_closed}) back to Eq. (\ref{eq:V_multiplier}) we obtain:
\begin{small}
\begin{equation}
\label{eq:update_V}
\mathbf{V}=\mathbf{U}-\frac{1}{n}\mathbf{U}\mathbf{1}\mathbf{1}^T-\frac{1}{n}\mathbf{1}\mathbf{1}^T\mathbf{U}+\frac{1}{n^2}\mathbf{1}\mathbf{1}^T\mathbf{U}\mathbf{1}\mathbf{1}^T
\end{equation}
\end{small}
\begin{algorithm}[tb]
\begin{small}
\caption{CutMatch}
\label{alg:CutMatch}
\begin{algorithmic}[1]
        \Require{$\mathbf{X}^{\text{init}}$ by Algorithm \ref{alg:BregProj}, $\mathbf{y}^{\text{init}}$, $\mathbf{A}$, $\mathbf{D}$, $\mathbf{W}$, $\lambda_1$, $\lambda_2$}
        \State $\mathbf{X}^{\text{cur}}\leftarrow\mathbf{X}^{\text{init}}$, $\mathbf{y}^{\text{cur}}\leftarrow\mathbf{y}^{\text{init}}$
        \State $\mathbf{E}\leftarrow\lambda_1\mathbf{D}+\lambda_2\mathbf{I}$
        \Repeat
            \State $\mathbf{L}^{\text{cur}}\leftarrow\lambda_2\left(\mathbf{I}-\mathbf{X}^{\text{cur}}\right)-\lambda_1\left(\mathbf{D}-\mathbf{W}\right)$
            \State // \emph{update cut:}
            \State $\mathbf{y}\leftarrow\arg\min_{\mathbf{y}}\frac{\mathbf{y}^T\mathbf{L}^{\text{cur}}\mathbf{y}}{\mathbf{y}^T \mathbf{y}}$
            \If{First iteration}
                \State // \emph{correlate starting point of matching:}
                \State $\mathbf{X}^{\text{cur}}_{ij}\leftarrow 0$ if $\text{sign}(\mathbf{y}_i)=\text{sign}(\mathbf{y}_j)$
                \State $\mathbf{X}^{\text{cur}}\leftarrow\mathfrak{P}(\mathbf{X}^{\text{cur}})$ using Algorithm \ref{alg:BregProj}
            \EndIf
            \State // \emph{update matching:}
            \State $\mathbf{X}^{\text{cur}}\leftarrow\arg\max_{\mathbf{X}}\mathbb{E}$ according to Algorithm \ref{alg:BregDirection}
        \Until{Converge}
        \State \Return $\mathbf{X}^{\text{opt}}\leftarrow\mathbf{X}^{\text{cur}}$, $\mathbf{y}^{\text{opt}}\leftarrow\mathbf{y}$
\end{algorithmic}
\end{small}
\end{algorithm}

We then consider the partial objective involving $\mathcal{C}_2$:
\begin{small}
\begin{equation}
\label{eq:Gradient_C2}
\begin{split}
\min_{\mathbf{V}}&\lVert \mathbf{V}-\mathbf{U}\rVert^2_F \\
\text{s.t. }&\mathbf{V}_{ii}=0,\mathbf{X}^{\text{prev}}_{ij}+\epsilon\mathbf{V}_{ij}\geq 0,\mathbf{X}^{\text{prev}}_{ij}+\epsilon\nabla\mathbf{X}_{ij}\leq 1
\end{split}
\end{equation}
\end{small}

This problem can be readily solved by letting $\mathbf{V}_{ii}=0$, truncating $\mathbf{V}_{ij}=-\mathbf{X}^{\text{prev}}_{ij}/\epsilon$ if $\mathbf{V}_{ij}<-\mathbf{X}^{\text{prev}}_{ij}/\epsilon$ and $\mathbf{V}_{ij}=(1-\mathbf{X}^{\text{prev}}_{ij})/\epsilon$ if $\mathbf{V}_{ij}>(1-\mathbf{X}^{\text{prev}}_{ij})/\epsilon$. By alternating the procedure between Eq. (\ref{eq:Gradient_C1}) and Eq. (\ref{eq:Gradient_C2}), the optimal direction $\nabla\mathbf{X}$ can be found. Hence the update rule over $\mathbf{X}$ is (recall $\mathbf{V}=\nabla\mathbf{X}$ by definition):
\begin{small}
\begin{equation}
\label{eq:update_matching}
\mathbf{X}^{\text{cur}}=\mathbf{X}^{\text{prev}}+\epsilon\nabla\mathbf{X}
\end{equation}
\end{small}

Fixing $\mathbf{y}$ and repeating the update rule until convergence, one can reach the stationary $\mathbf{X}$ to Problem (\ref{eq:CutMatch}).

We call the above algorithm \textbf{Iterative Bregman Gradient Projection} (\textbf{IBGP}) and summarize it in Algorithm \ref{alg:BregDirection}. \textbf{IBGP} can easily adapt to standard graph matching problem (with or without linear part) by removing constraint $\nabla\mathbf{X}_{ii}=0$  and $\nabla\mathbf{X}=\nabla\mathbf{X}^T$ from Eq. (\ref{eq:GradUpdate}) -- fortunately the corresponding update rule is still as Eq. \ref{eq:update_V}. See supplementary material for derivation. Furthermore, IBGP can be integrated with popular non-convex optimization framework, such as convex-concave relaxation \cite{zhou2012factorized} and path following \cite{wang2016branching}. We leave this to future work. The following proposition ensures the convergence of the update strategy.
\begin{prop}
\label{th:OptGradient}
The optimal solution $\nabla\mathbf{X}^{\text{opt}}$ to Eq. (\ref{eq:GradUpdate}) must be a non-decreasing direction.
\end{prop}
\begin{proof}
As objective \ref{eq:GradUpdate} is convex, the global optima is reachable. First note that matrix $\mathbf{O}$ is in the feasible set of \ref{eq:GradUpdate} as $0\leq \mathbf{X}_{ij}^{\text{prev}}\leq 1$. Thus if an optimal solution $\mathbf{V}$ is with $\Tr(\mathbf{V}^T\nabla\mathbb{E})<0$, we must have $\lVert\mathbf{V}-\nabla\mathbb{E}\rVert_F^2=\Tr(V^T V)-2\Tr(\mathbf{V}^T\nabla\mathbb{E})+\Tr(\nabla\mathbb{E}^T\nabla\mathbb{E})>0+0+\Tr(\nabla\mathbb{E}^T\nabla\mathbb{E})=\lVert\mathbf{O}-\nabla\mathbb{E}\rVert_F^2$. Then $\mathbf{V}$ cannot be optimal. This implies that for any optimal $\nabla\mathbf{X}^{\text{opt}}$ we have $\Tr(\nabla\mathbb{E}^T\nabla\mathbf{X}^{\text{opt}})\geq 0$, thus $\nabla\mathbf{X}^{\text{opt}}$ must be a non-decreasing direction. QED.
\end{proof}

We summarize \textbf{Initialization}, \textbf{Update cut} and \textbf{Update matching} in Algorithm \ref{alg:CutMatch} which is termed as \textbf{CutMatch}.

\textbf{Discretization} We apply Hungarian method \cite{bruff2005assignment} over $\mathbf{X}$ to calculate the discrete solution to graph matching sub-problem. To obtain the discrete solution to partition, we simply refer to the sign of each element in $\mathbf{y}$. Though more complicated discretization method can be devised to avoid the conflict between matching $\mathbf{X}$ and $\mathbf{y}$, an easier way is adopted as we observe that the aforementioned discretization methods yield high performance in experiments.

\textbf{Convergence} According to the four constraints in Eq. (\ref{eq:CutMatch}), $\mathbf{X}$ lies in a closed convex set. Similarly, under the latter three constraints, $\mathbf{y}$ is in a closed ellipsoid with dimension at most $m-1$, which is also convex. Therefore the objective in Eq. (\ref{eq:CutMatch}) should be the upper bounded. The updating rule in Algorithm \ref{alg:CutMatch}, on the other hand, ensures that the energy $\mathbb{E}$ ascends at each iteration. In summary, Algorithm \ref{alg:CutMatch} is guaranteed to converge to a local stationary point.

\section{Experiment and Discussion}
\subsection{Graph matching on synthetic data}
\begin{figure*}
\begin{center}
    \includegraphics[width = 0.28\textwidth]{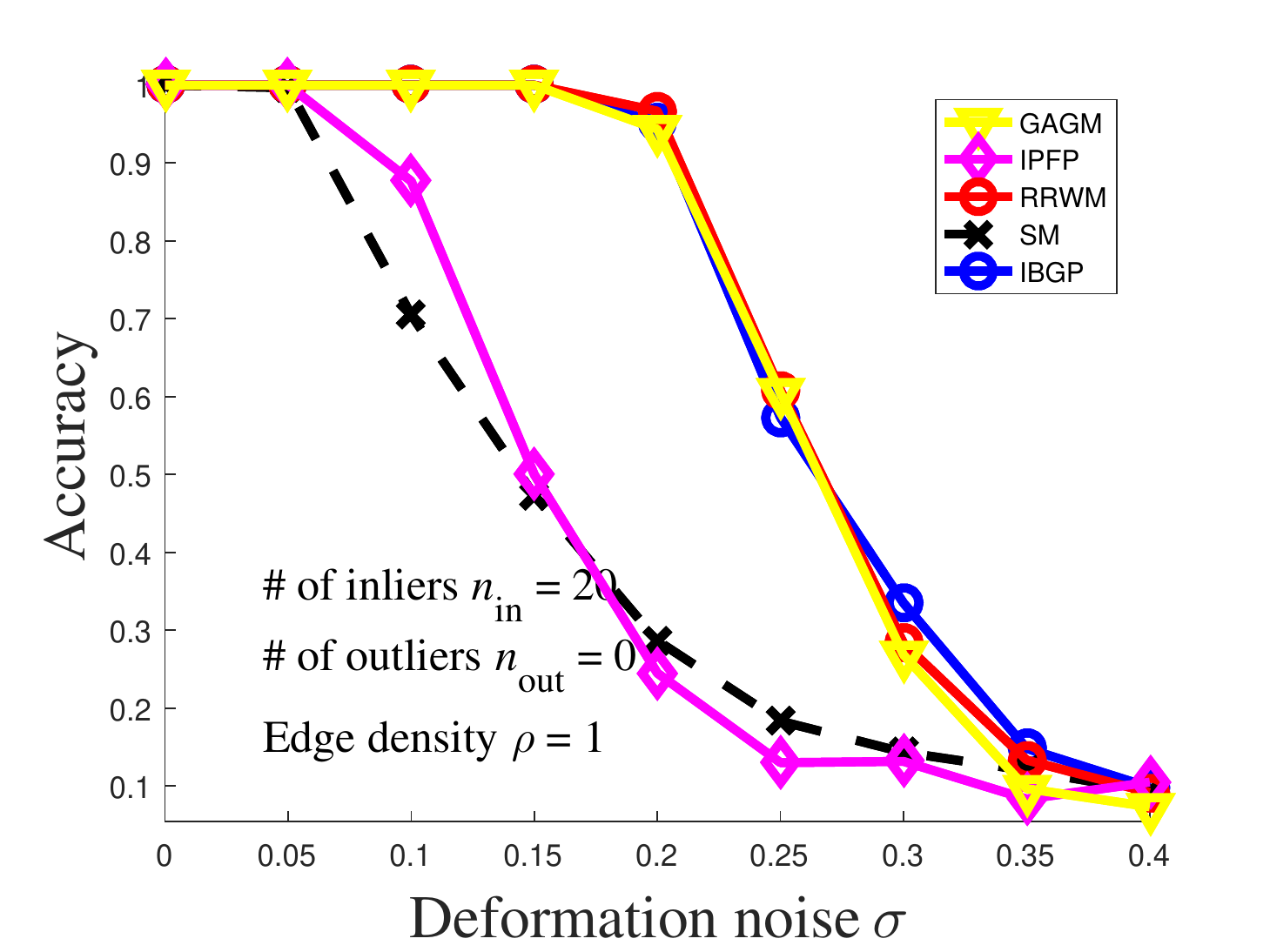}
    \includegraphics[width = 0.28\textwidth]{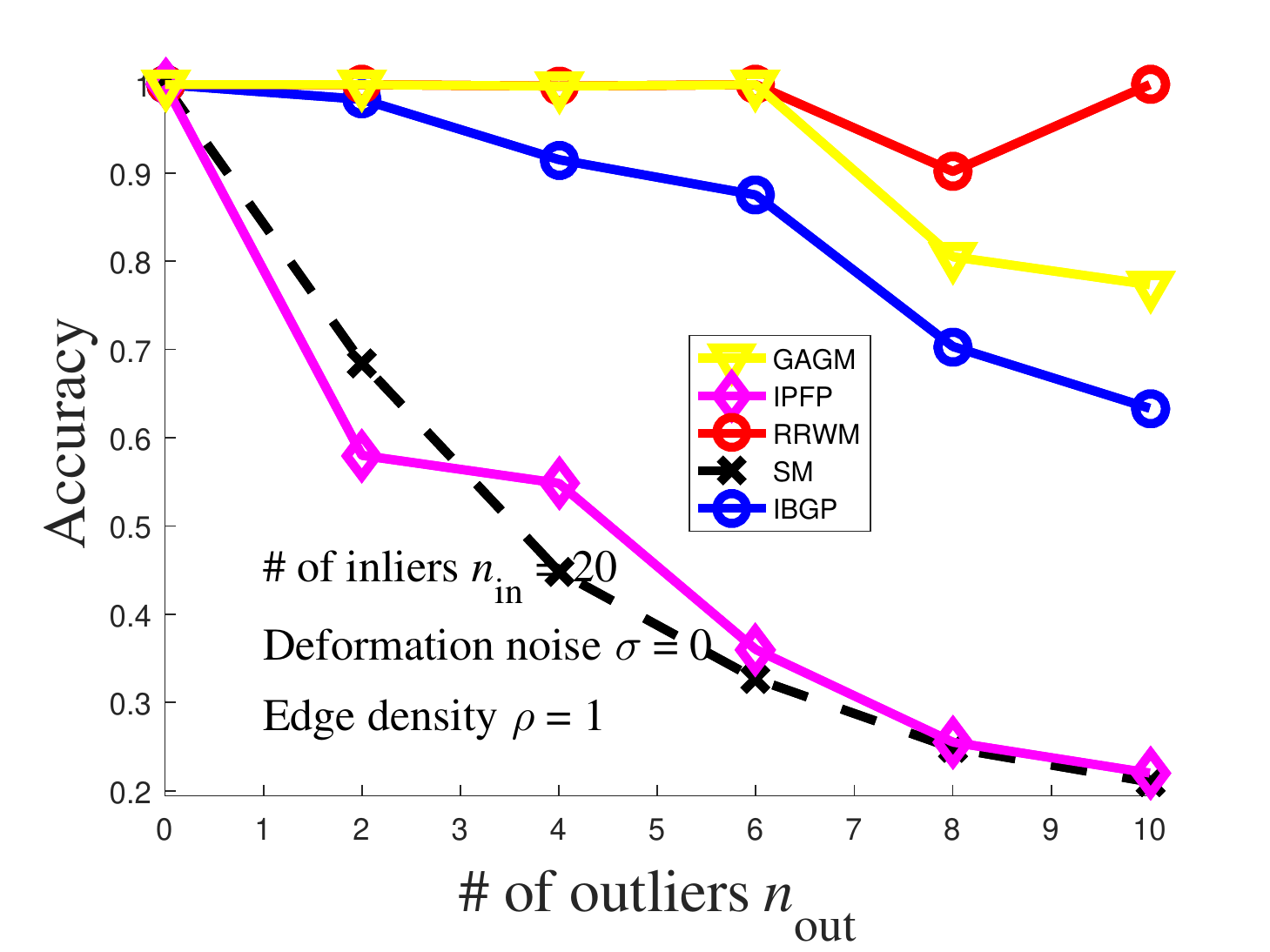}
    \includegraphics[width = 0.28\textwidth]{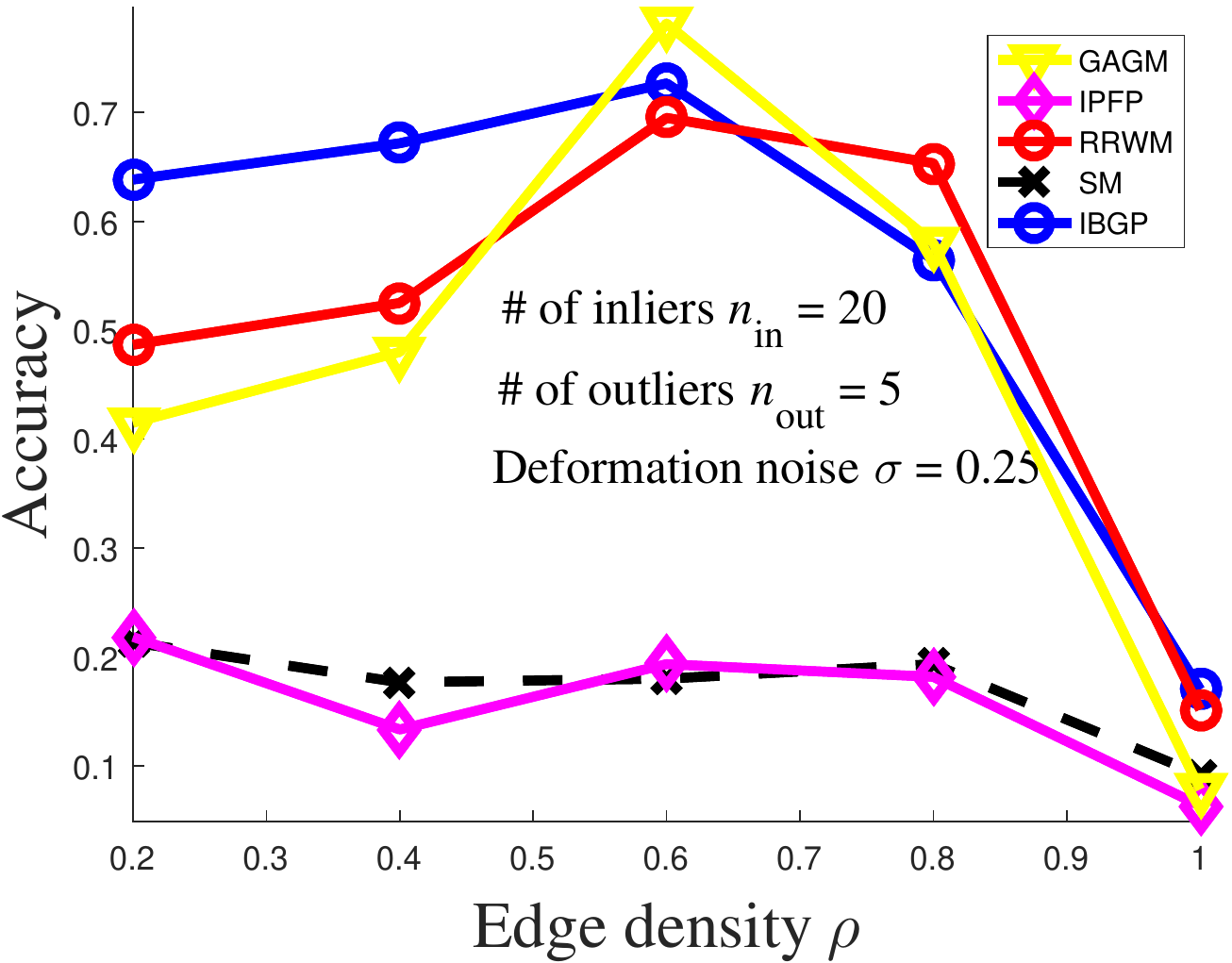} \\
    \includegraphics[width = 0.28\textwidth]{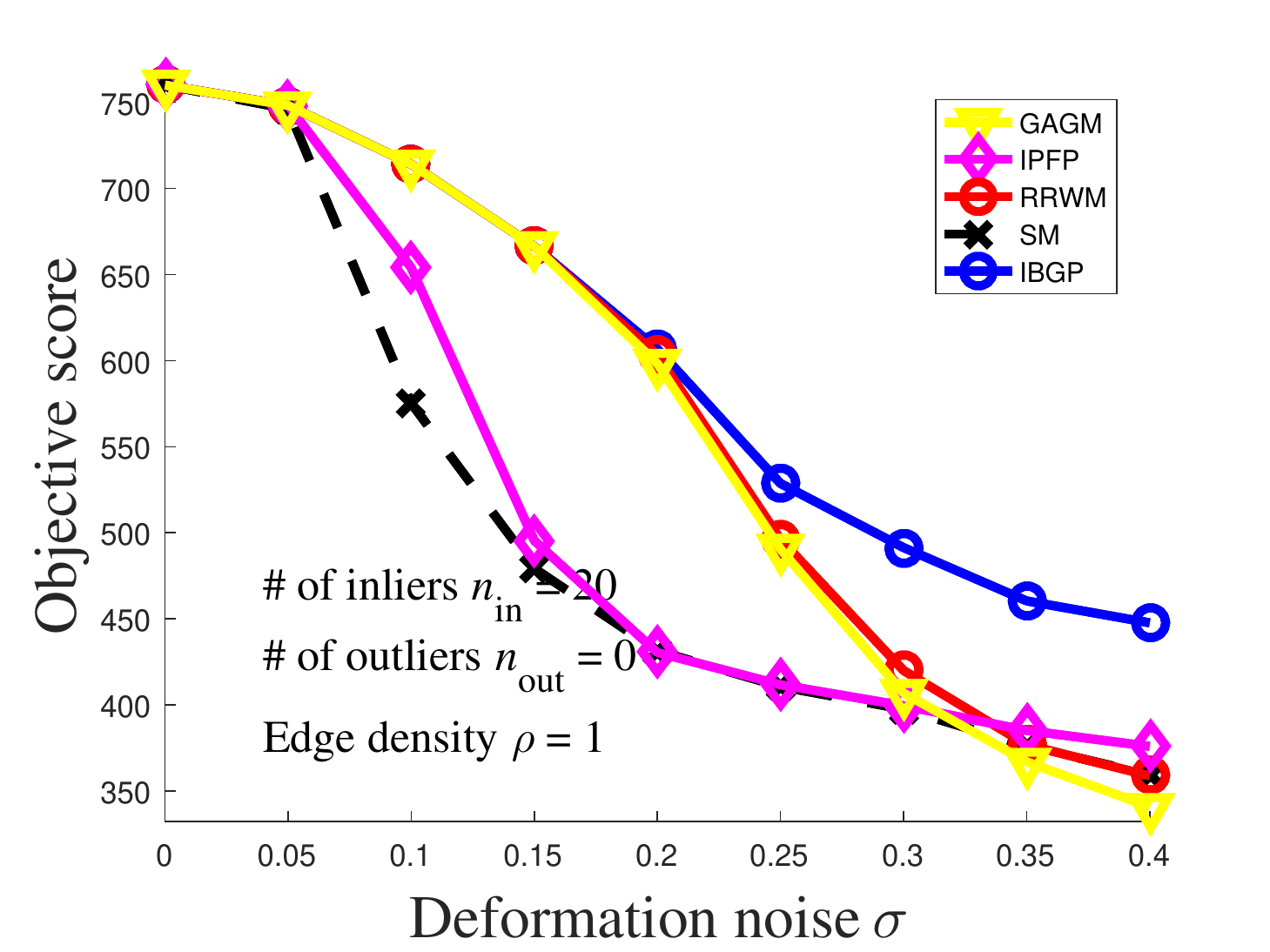}
    \includegraphics[width = 0.28\textwidth]{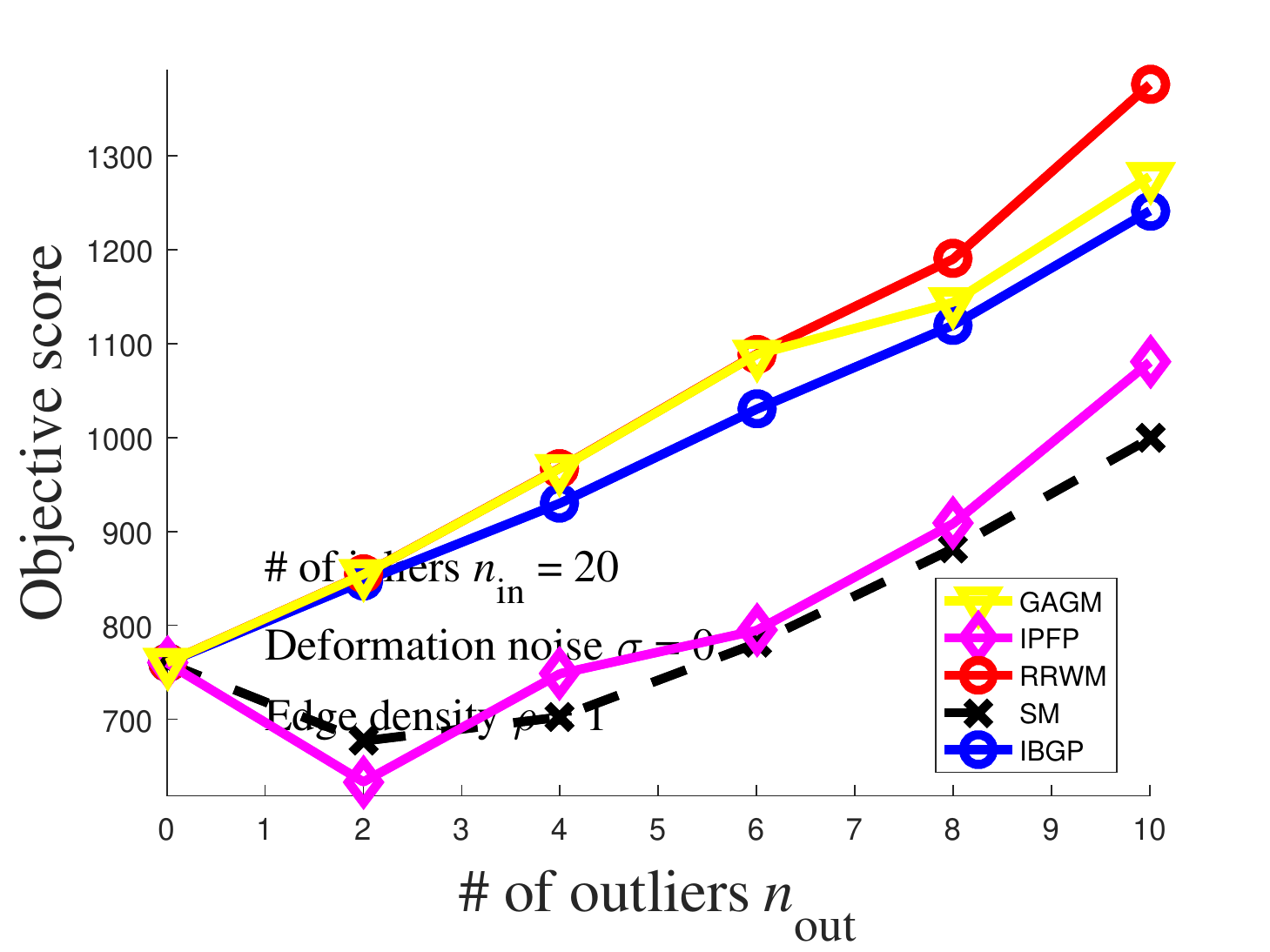}
    \includegraphics[width = 0.28\textwidth]{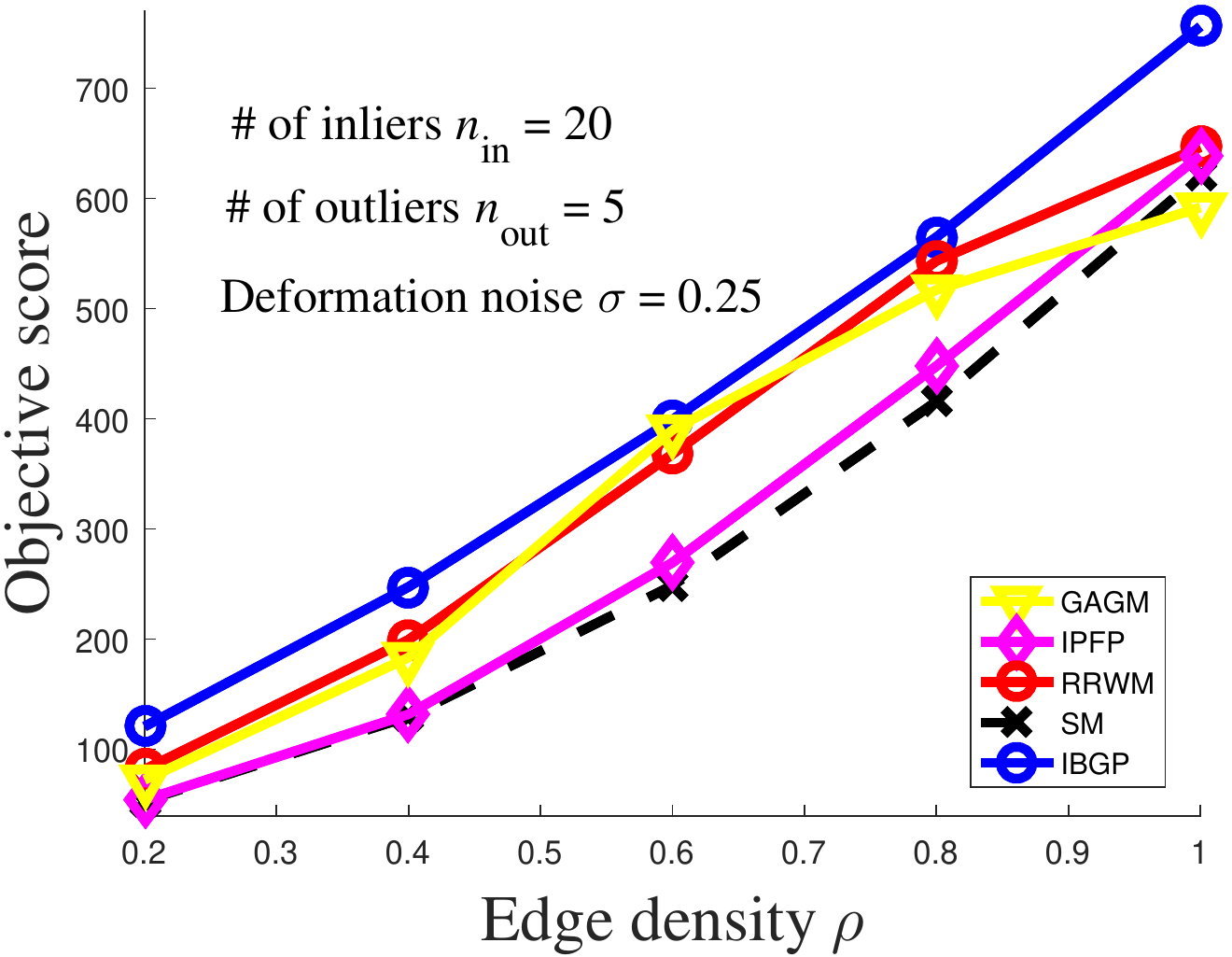}
\end{center}
\vspace{-10pt}
\caption{Evaluation of the proposed IBGP and peer methods on synthetic data for standard two-graph matching problem.}
\label{fig:exp_gm_random}
\end{figure*}

Since we have devised a new graph matching solver -- IBGP along with CutMatch (see Algorithm \ref{alg:BregProj}), we first test it on synthetic data by the protocol in \cite{cho2010reweighted}. Four popular graph matching solvers are compared, including graduated assignment (GAGM) \cite{gold1996graduated}, integer projection fixed point (IPFP) \cite{leordeanu2009integer}, reweighed random walk (RRWM) \cite{cho2010reweighted} and spectral matching (SM) \cite{leordeanu2005spectral}. The four peer methods and IBGP are all initialized with a uniform doubly stochastic matrix. Various levels of deformation, outlier and density change are randomly generated and  casted to the affinity matrix. Figure \ref{fig:exp_gm_random} demonstrates the experimental results. In deformation test, we generate $30$ pairs of graphs with $20$ nodes, then varying noise from $0$ to $0.4$ with increment $0.05$ is added to the affinity. In outlier test, up to $10$ outlier points by increments $2$ are further generated to disturb the matching. In density test, we test the matching performance by varying density from $0.2$ to $1$ as well as noise $0.25$ and $5$ extra outlier points. We see that in deformation and density test, IBGP performs competitively compared to state-of-the-art algorithms in most settings in terms of accuracy and affinity score. Though IBGP is relatively sensitive to outlier, we notice that it has high stability in combinations of various disturbance. From numerical perspective, we further observe that when affinity matrix is negative definite (or contains many negative elements), SM, RRWM and GAGM always report error, due to the spectral limitation.
\subsection{Joint cut and matching on synthetic data}
\label{sec:exp_syn_cutmatch}
\begin{figure*}
\begin{center}
    \includegraphics[width = 0.24\textwidth]{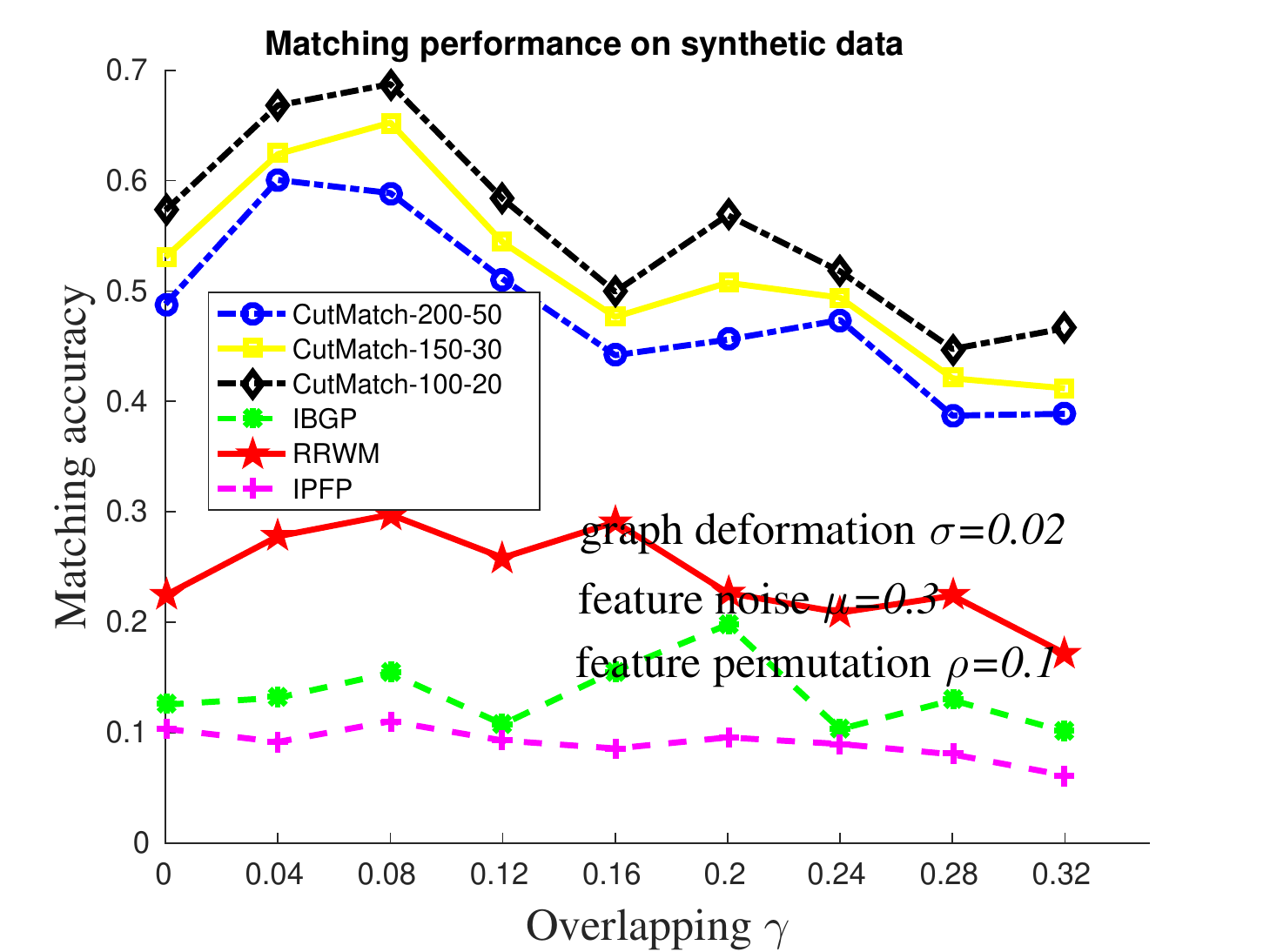}
    \includegraphics[width = 0.24\textwidth]{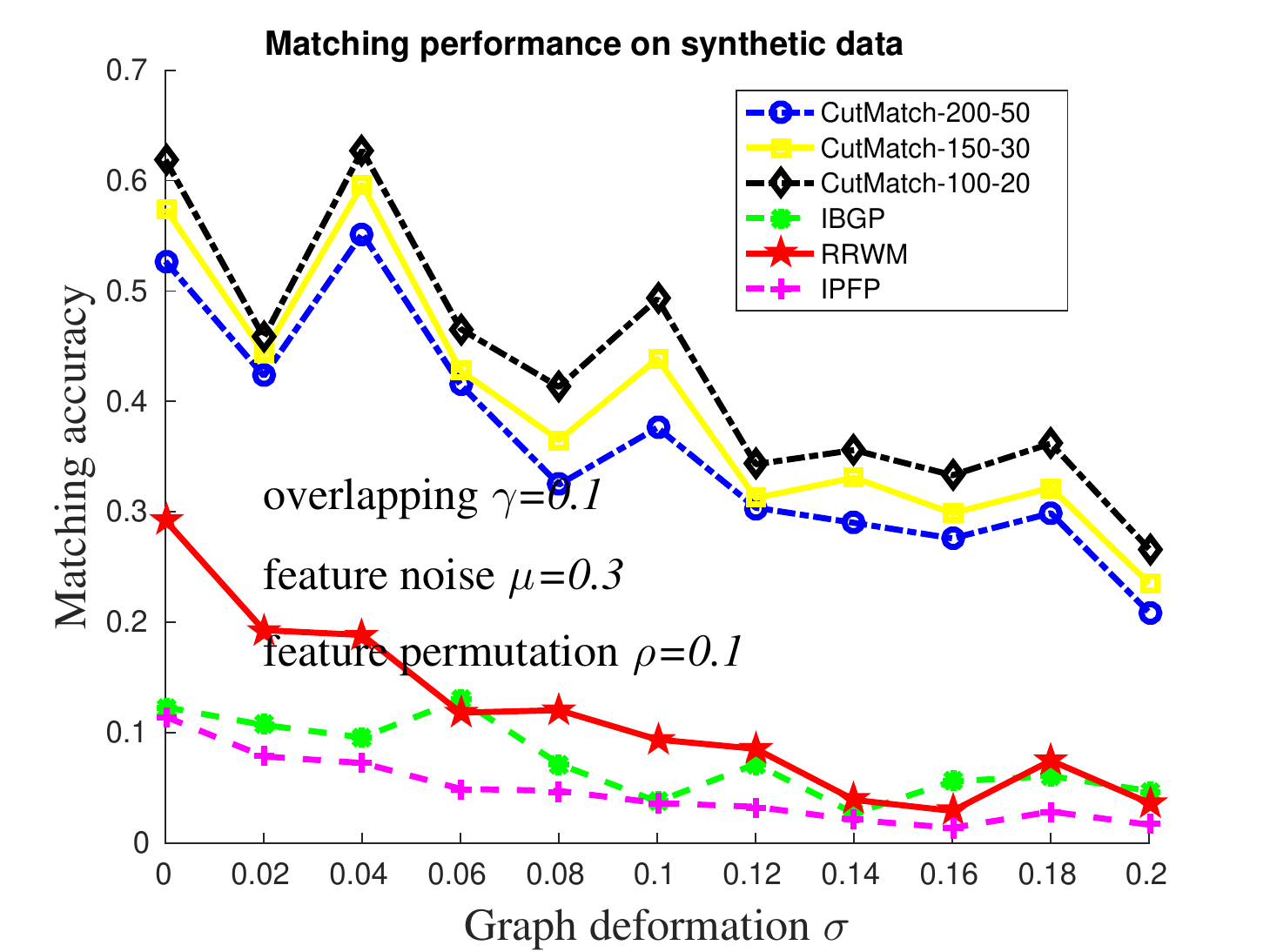}
    \includegraphics[width = 0.24\textwidth]{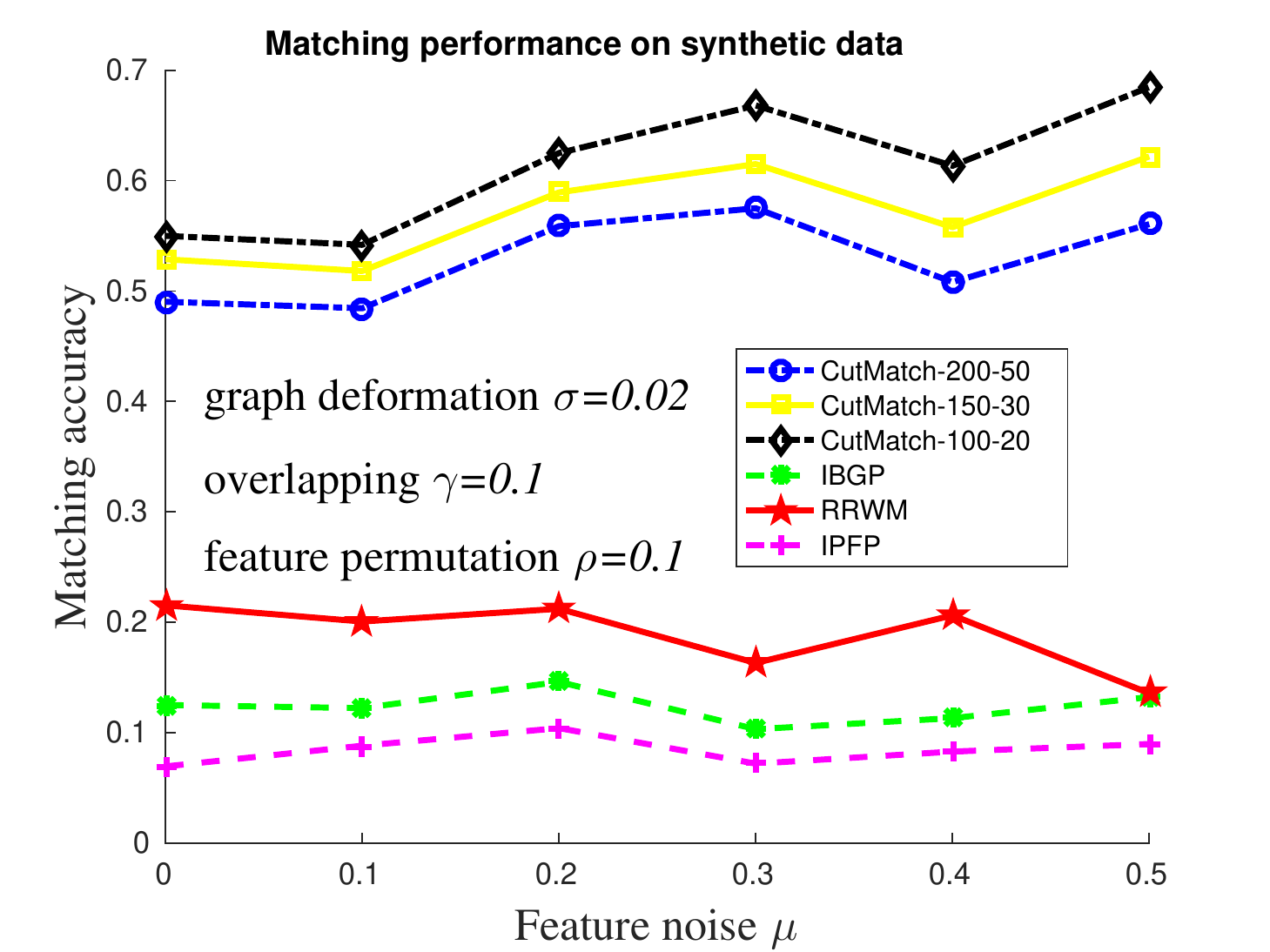}
    \includegraphics[width = 0.24\textwidth]{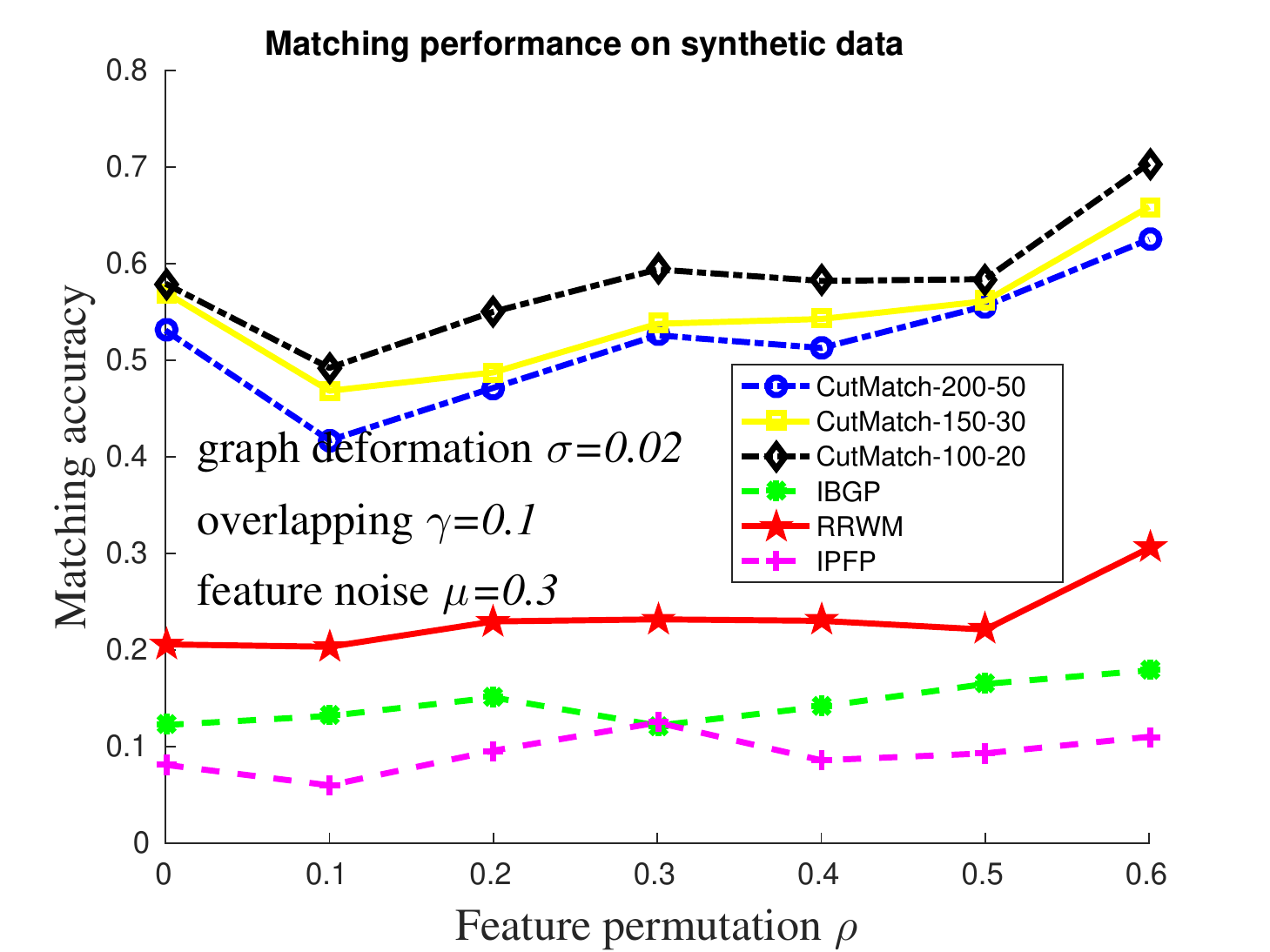} \\
    \includegraphics[width = 0.24\textwidth]{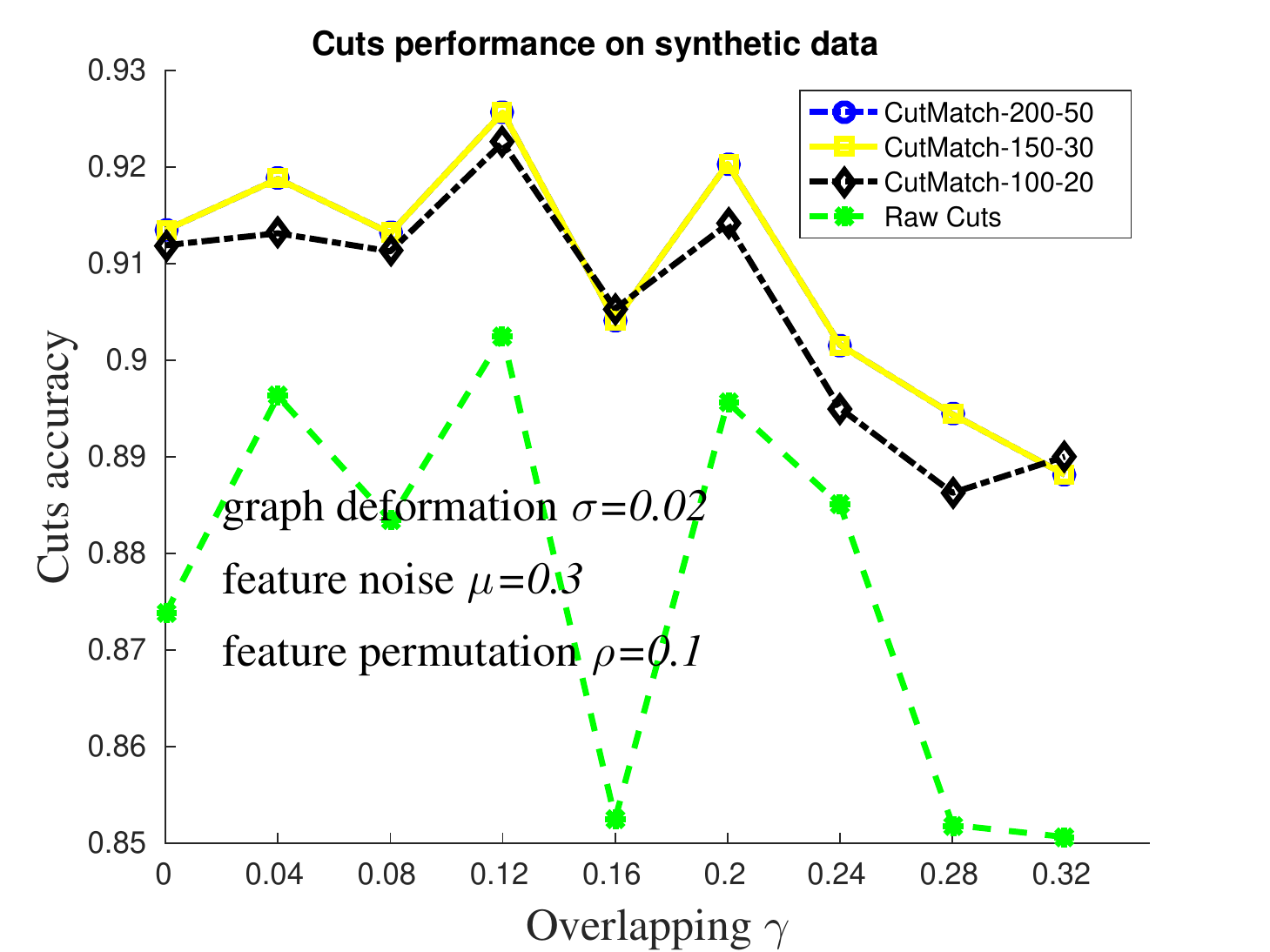}
    \includegraphics[width = 0.24\textwidth]{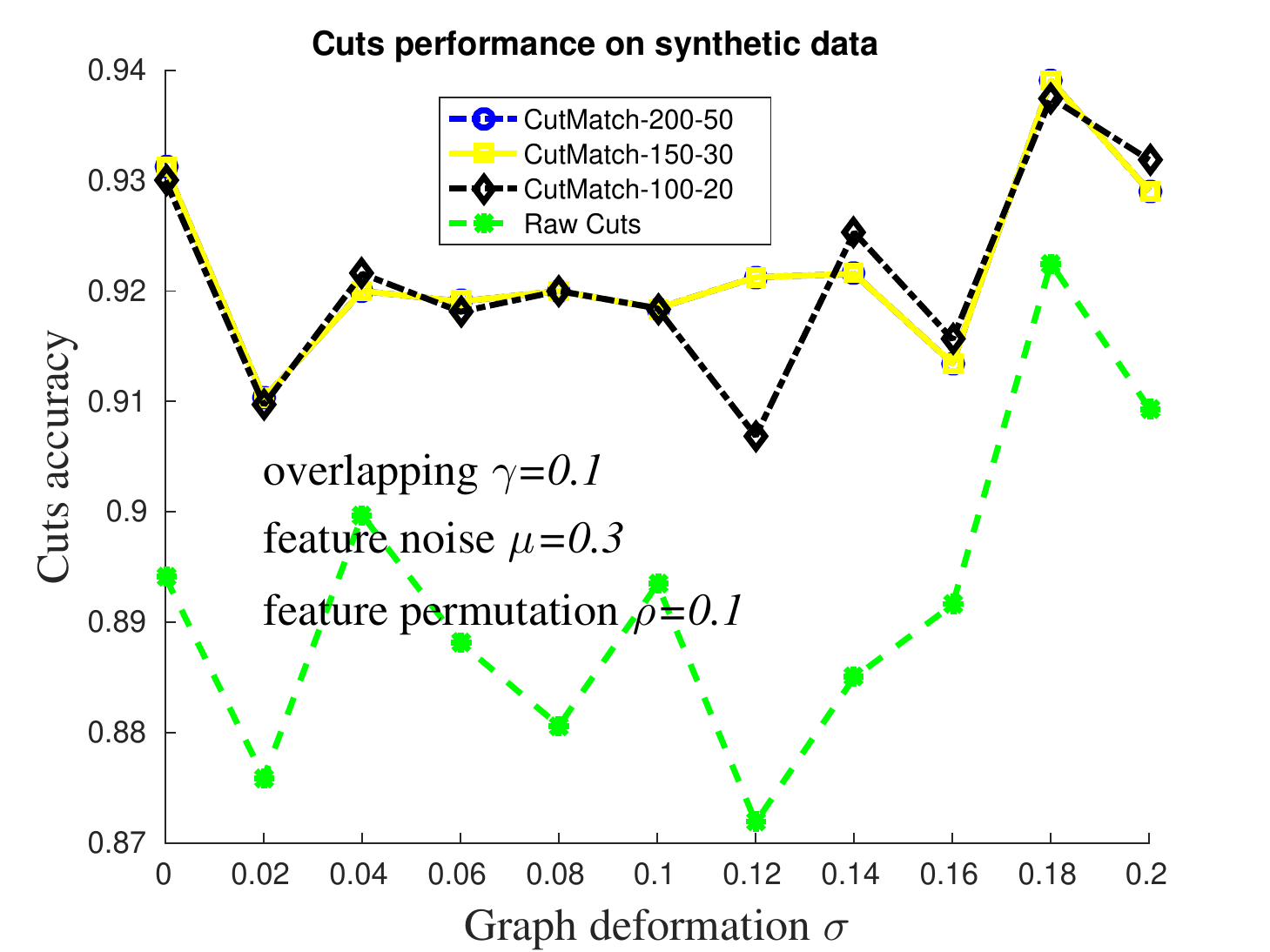}
    \includegraphics[width = 0.24\textwidth]{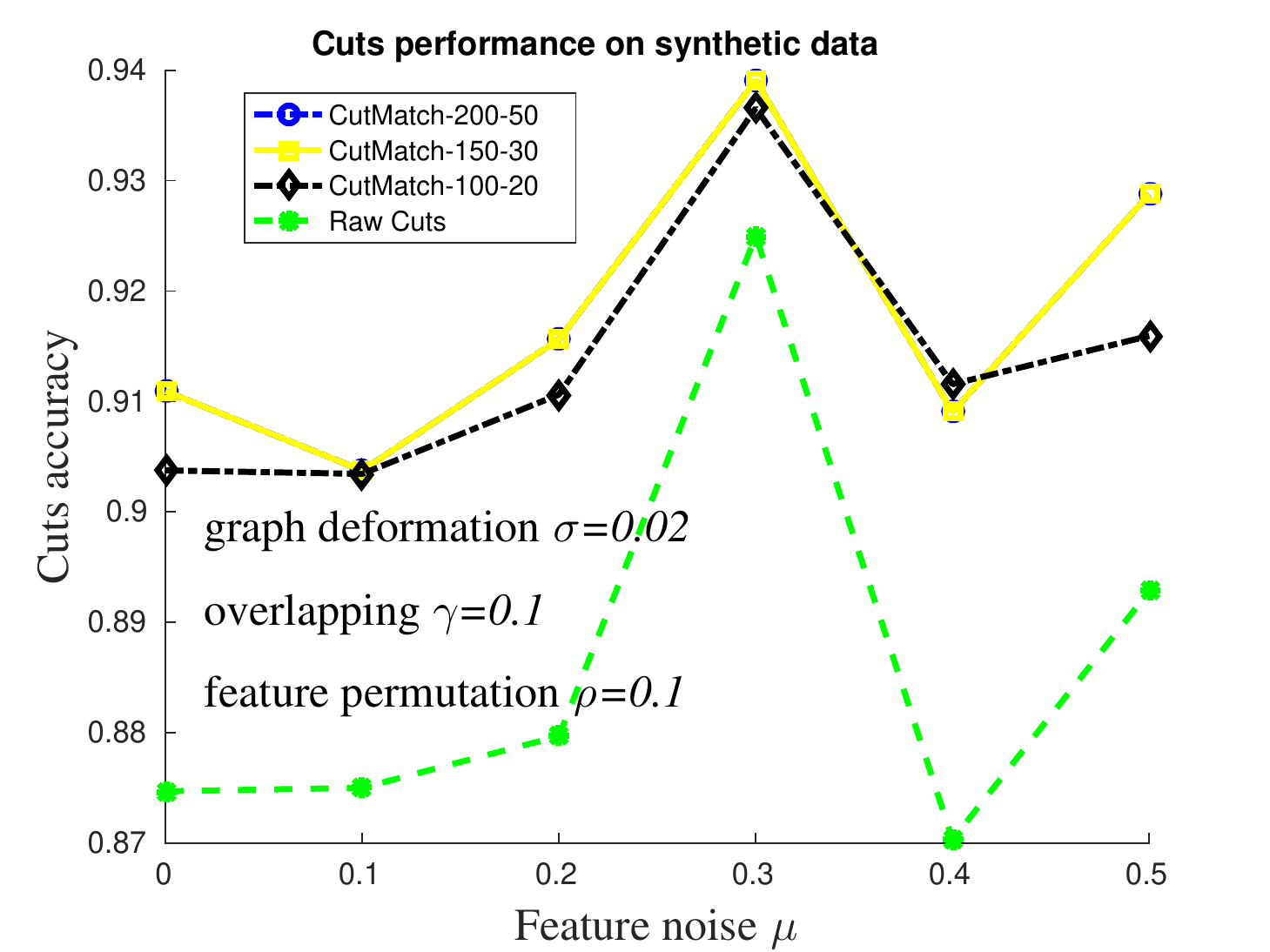}
    \includegraphics[width = 0.24\textwidth]{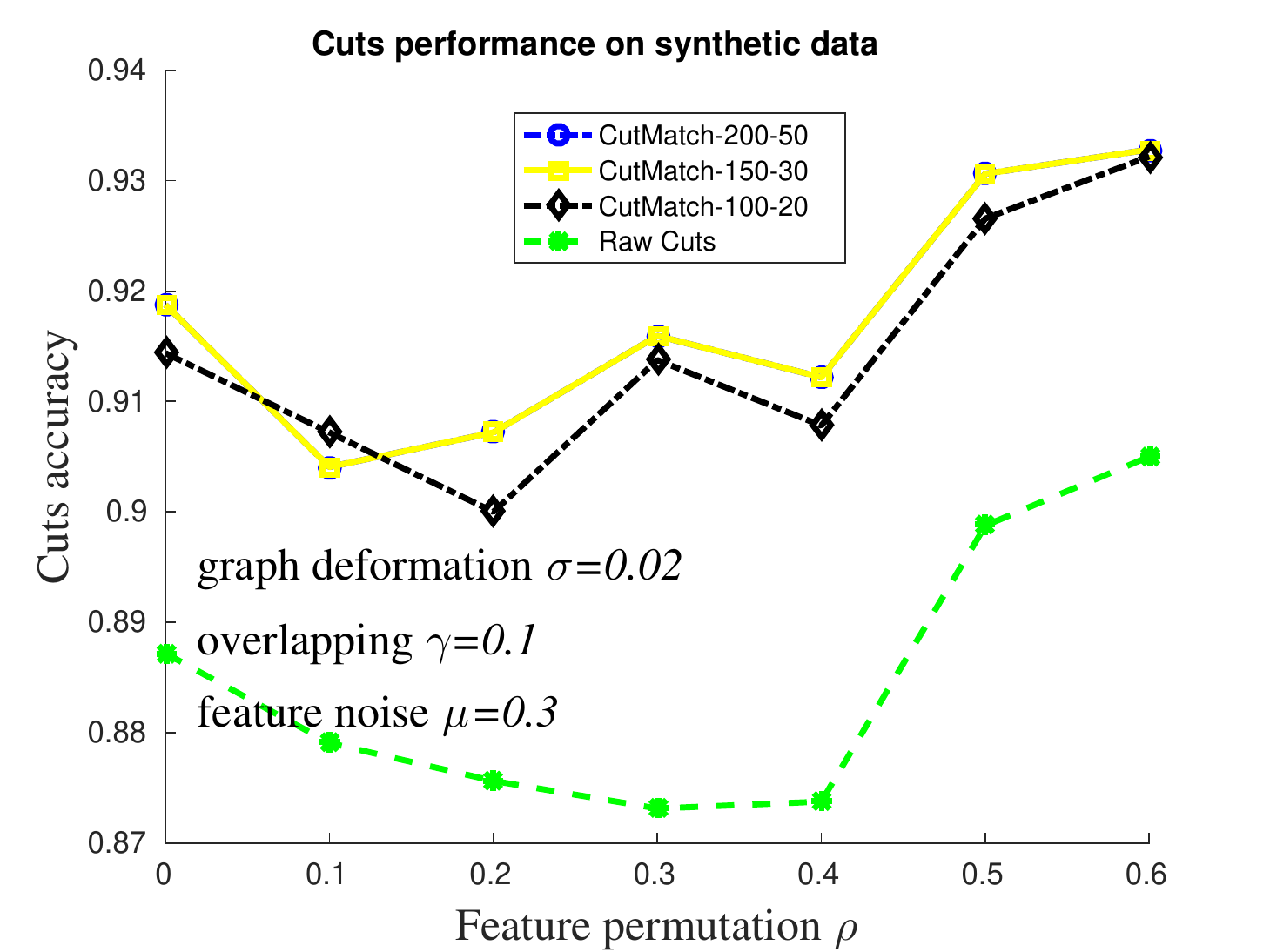}
\end{center}
\vspace{-10pt}
\caption{Evaluation of CutMatch and vanilla graph cuts and graph matching solvers on synthetic data for the joint cut and matching task.}
\label{fig:exp_cutmatch_syn}
\end{figure*}

\textbf{Dataset} We construct the benchmark by randomly generating synthetic graphs with different types of disturbances including noise, randomly displacement and feature permutation. For each partition, there are $20$ nodes from normal distribution, thus $40$ nodes in total for the final test graph. Specifically we first generate partition $\mathcal{P}_1$ and translate it to obtain partition $\mathcal{P}_2$, while the translation retains overlapping $\gamma$ measuring the horizontal distance between the most left point of $\mathcal{P}_2$ and the most right point of $\mathcal{P}_1$. We further add level $\sigma$ Gaussian noise to partition $\mathcal{P}_2$. Having obtained the nodes, we perform Delaunay triangulation \cite{guibas1985primitives} to obtain the edges. For an element $(ia:jb)$ in affinity matrix (for \textbf{Graph Matching}) $\mathbf{A}$, we calculate it as $\mathbf{A}_{ia:jb}=\exp(-(d_{ij}-d_{ab})^2/\delta_1)$, where $d_{ij}$ is the Euclidean distance between node $i$ and $j$. For each corresponding node pair $i$ and $j$ in $\mathcal{P}_1$ and $\mathcal{P}_2$, respectively, we assign a randomly generated $128$-D feature $\mathbf{f}_i=\mathbf{f}_j$ (to simulate SIFT feature) from normal distribution and further add level $\mu$ of Gaussian noise to $\mathbf{f}_j$. $\rho$ level of random permutation is also performed to disturb the order of the features in the whole graph. For any node $i\in\mathcal{P}_1\cup\mathcal{P}_2$, we denote its $2$-D location $\mathbf{l}_i$. Thus similarity matrix (for \textbf{Graph Cuts}) $\mathbf{W}_{ij}=\exp(-\lVert\mathbf{f}_i-\mathbf{f}_j\rVert_2^2/\delta_2)+\exp(-\lVert\mathbf{l}_i-\mathbf{l}_j\rVert_2^2/\delta_3)$ and we set $\delta_1=0.5$, $\delta_2=5$, $\delta_3=0.5$ in all tests.

\textbf{Evaluation metrics} We calculate the average $Accuracy$ for both graph cut and graph matching. For graph cut, $Accuracy$ represents the portion of correctly clustered nodes w.r.t. the sum of all nodes. As the final output of cuts is a $sign$ function, thus given a ground-truth label vector $\mathbf{C}\in\{-1,1\}^{2n}$ and an output label vector $\mathbf{C}_{cut}\in\{-1,1\}^{2n}$, we calculate as $Accuracy=\frac{\max\{\text{number of }\mathbf{C}=\mathbf{C}_{out},\text{number of }\mathbf{C}\neq\mathbf{C}_{cut}\}}{2n}$. For graph matching, $Accuracy$ corresponds to the portion of correctly matched nodes w.r.t. the sum of all nodes. The metric is also adopted in real-image test in Section \ref{sec:exp:image}. Note in the graph cuts experiments, the minimal $Accuracy$ is $0.5$.

\textbf{Results} We evaluate the performance of CutMatch by varying $\gamma$, $\sigma$, $\mu$ and $\rho$. For each value of parameters, we randomly generate $80$ graphs and calculate the average cuts and matching accuracy. Figure \ref{fig:exp_cutmatch_syn} shows the statistical results, and the first and second columns correspond to the matching and the cuts accuracy, respectively. We also present the performance with three pairs of $\lambda_1$ and $\lambda_2$ values: $(200,50)$, $(150,30)$ and $(100,20)$. Specifically, Fig. \ref{fig:cutmatch_example} shows an example with combined disturbance, where $\gamma=0.2$, $\sigma=0.1$, $\mu=0.3$ and $\rho=0.1$. Figure \ref{fig:exp_cutmatch_syn} presents the raw accuracy by vanilla matching: RRWM, IPFP and IBGP, together with the raw cuts accuracy with vanilla spectral graph cuts \cite{shi2000normalized}. One can observe that, with the joint objective, CutMatch significantly outperforms vanilla cuts and matching algorithms. As well as the performance enhancement, CutMatch is also robust to all kinds of disturbances. We also found that matching nodes (RRWM, IPFP and IBGP) within one graph is extremely sensitive to the initial $\mathbf{X}$. The CutMatch approach, however, is capable of correlating the starting points of cuts and matching alternatively in each iteration. In all the experiments, we observe that $5$ iterations are sufficient to yield satisfactory performance.

\subsection{Joint cut and matching on real images}
\label{sec:exp:image}
\textbf{Dataset} We collect $15$ images from internet containing two similar objects with the same category (\textbf{see supplemental material for the dataset}). For each image, we manually select $10$ to $20$ landmark nodes and establish the ground-truth correspondences. To obtain the connectivity, we first apply Delaunay triangulation on each partition, and then all the landmarks. The union of the edges in the three triangulations is regarded as the baseline connectivity. We further calculate the SIFT feature of each node. The same strategy on generating affinity $\mathbf{A}$ and similarity $\mathbf{W}$ is applied. For generating matrix $\mathbf{W}$, we set $\lambda_2=2\times 10^5$ and $\lambda_3=100$, which is to balance the un-normalized SIFT features. Because we observe that sometimes the deformation contained in real images is so large and arbitrary that local connectivity obtained from triangulation changes drastically, we test the performance under varying edge sampling rate $\eta$, where $\eta=1$ and $\eta=0$ correspond to fully connected and baseline graphs, respectively. As we believe that real images already contain natural contaminations (e.g., noises, deformation and outliers), we don't add extra degradations in this experiment as in section \ref{sec:exp_syn_cutmatch}. This dataset will be published.

\begin{figure}
\centering
    \includegraphics[width=.235\textwidth]{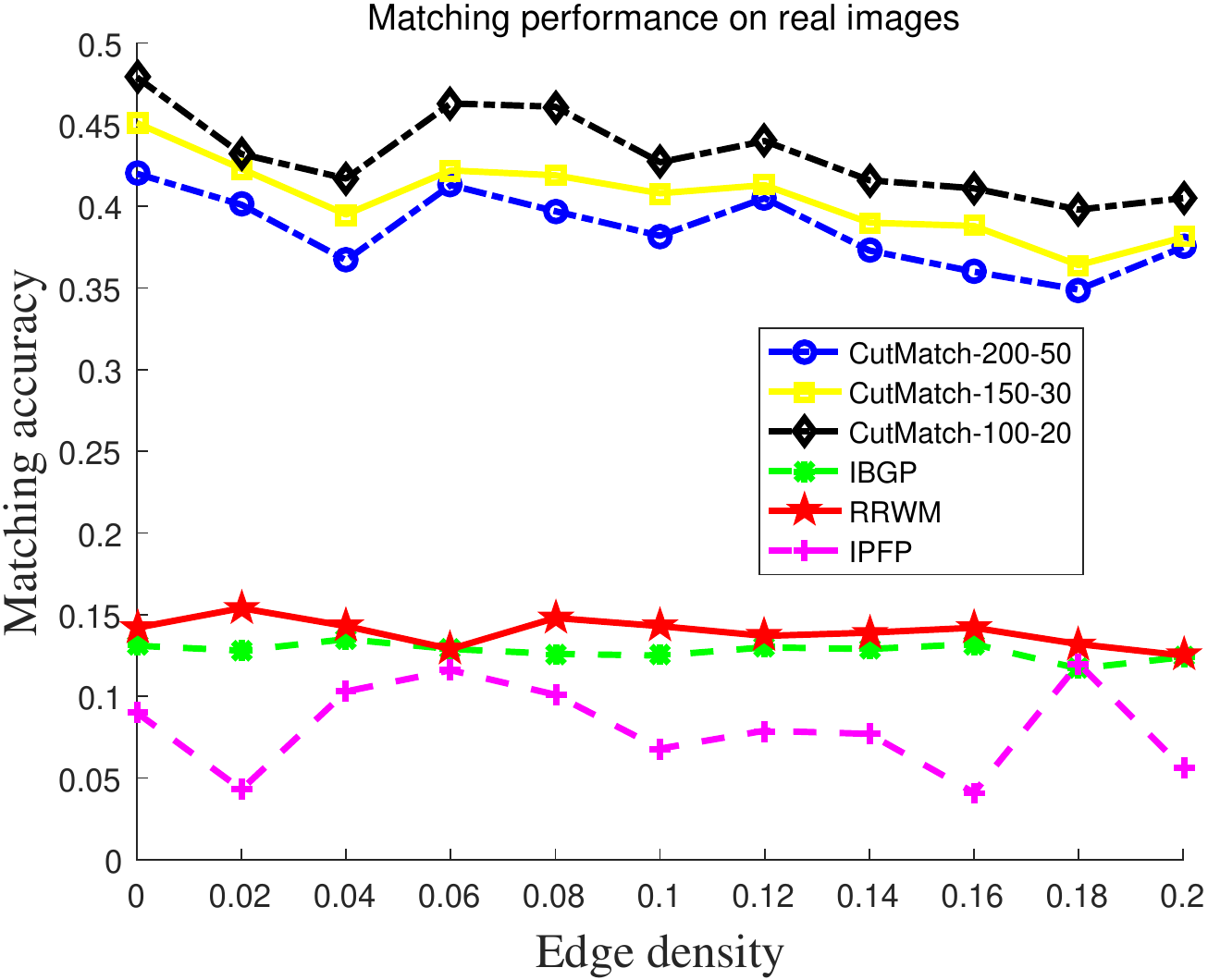}
    \includegraphics[width=.235\textwidth]{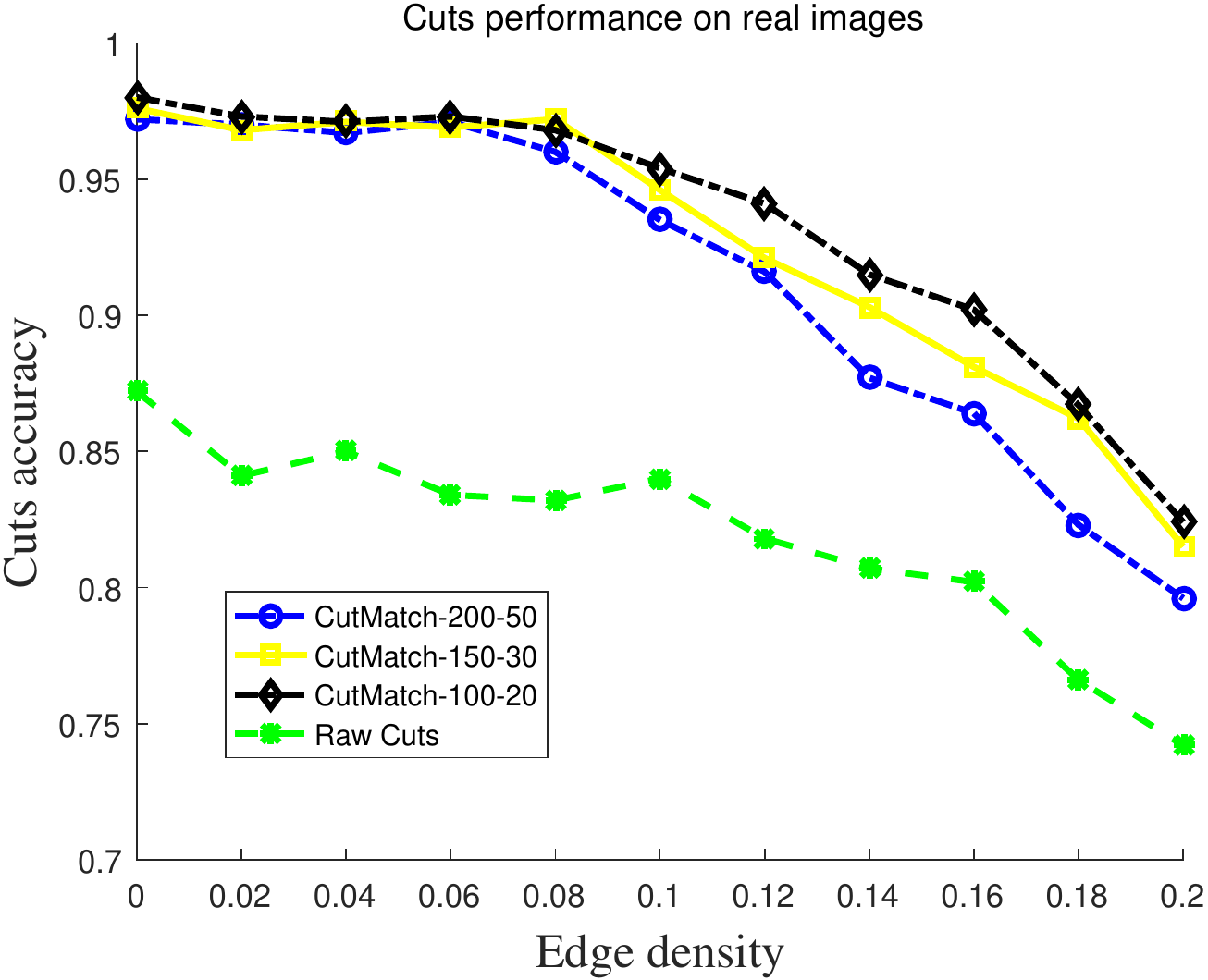}
\caption{Joint cuts and matching results on real-world images. \textbf{Left}: CutMatch with different parameter settings $(\lambda_1, \lambda_2) = \{(200, 50), (150, 30), (100, 20)\}$ vs. three vanilla graph matching solver IBGP, RRWM \cite{cho2010reweighted}, IPFP \cite{leordeanu2009integer}. \textbf{Right}: CutMatch $(\lambda_1, \lambda_2) = \{(200, 50), (150, 30), (100, 20)\}$ vs. vanilla graph cuts \cite{shi2000normalized}.}
\label{fig:image_result}
\end{figure}

\begin{figure*}
\centering
{   \begin{minipage}{0.18\textwidth}
        \subfigure[Ground-truth: \textbf{20/20}]{\includegraphics[width=1\textwidth]{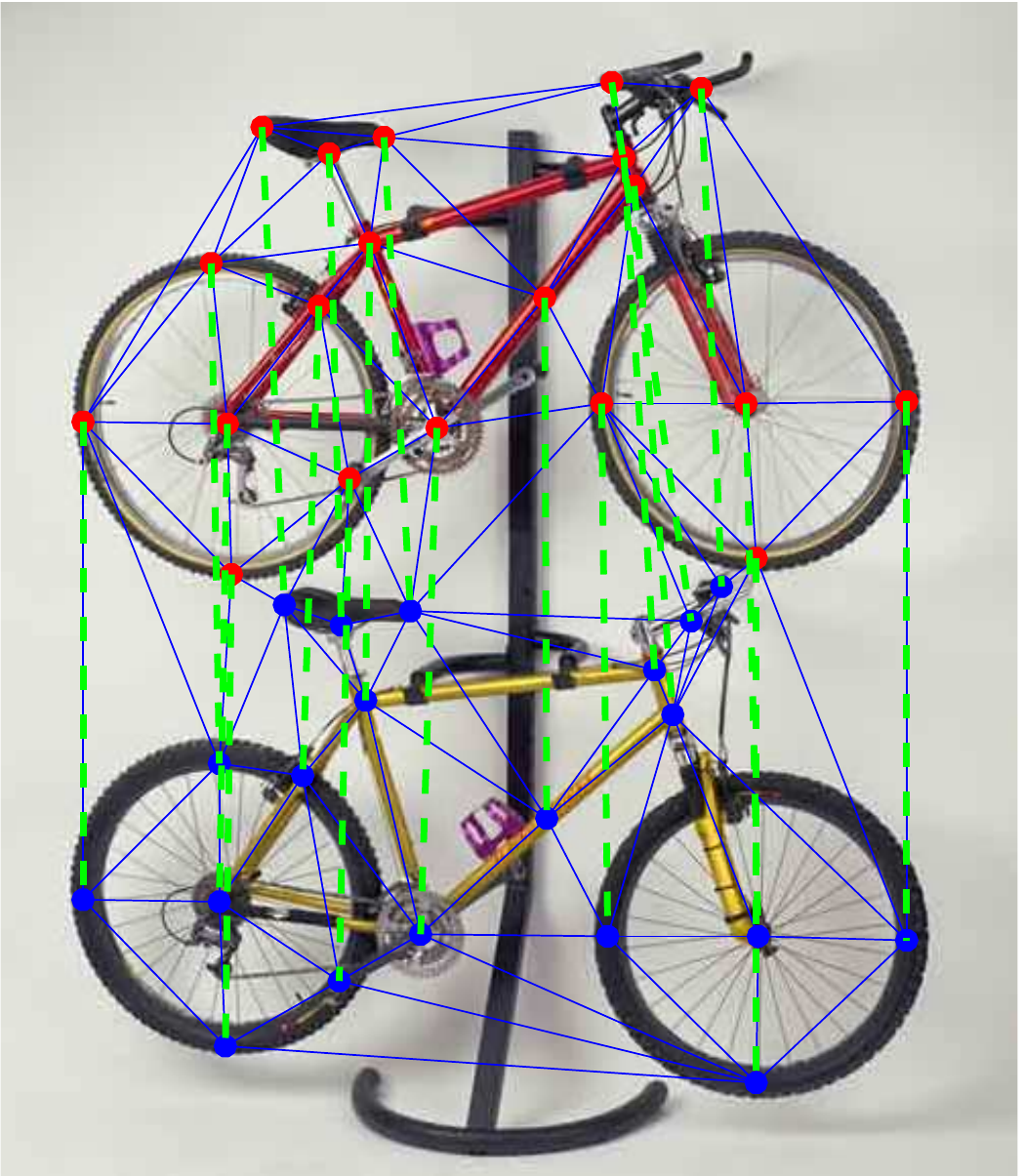}}\\\vspace{-5pt}
        \subfigure[Ground-truth: \textbf{25/25}]{\includegraphics[width=1\textwidth]{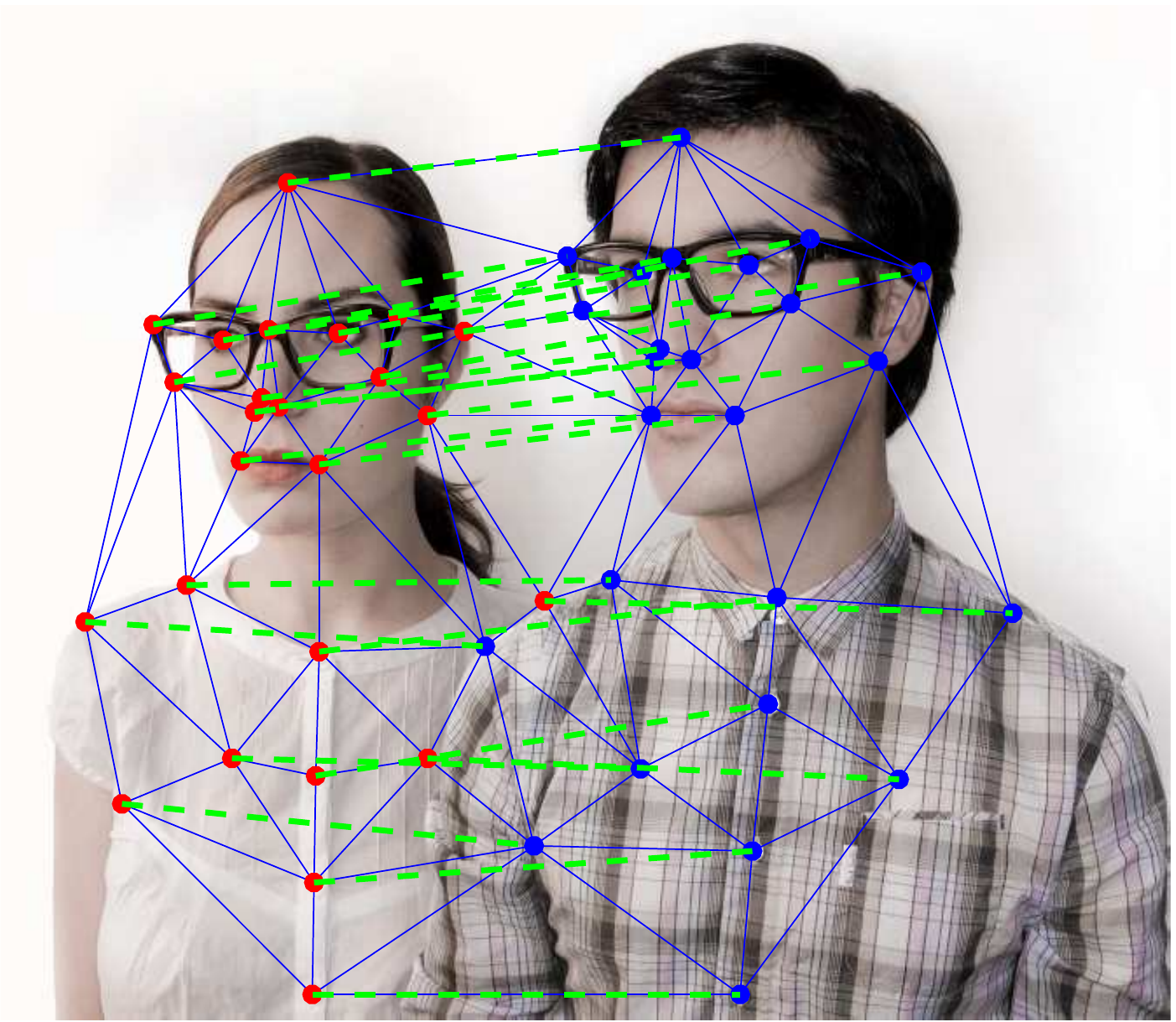}} \\\vspace{-5pt}
        \subfigure[Ground-truth: \textbf{15/15}]{\includegraphics[width=1\textwidth]{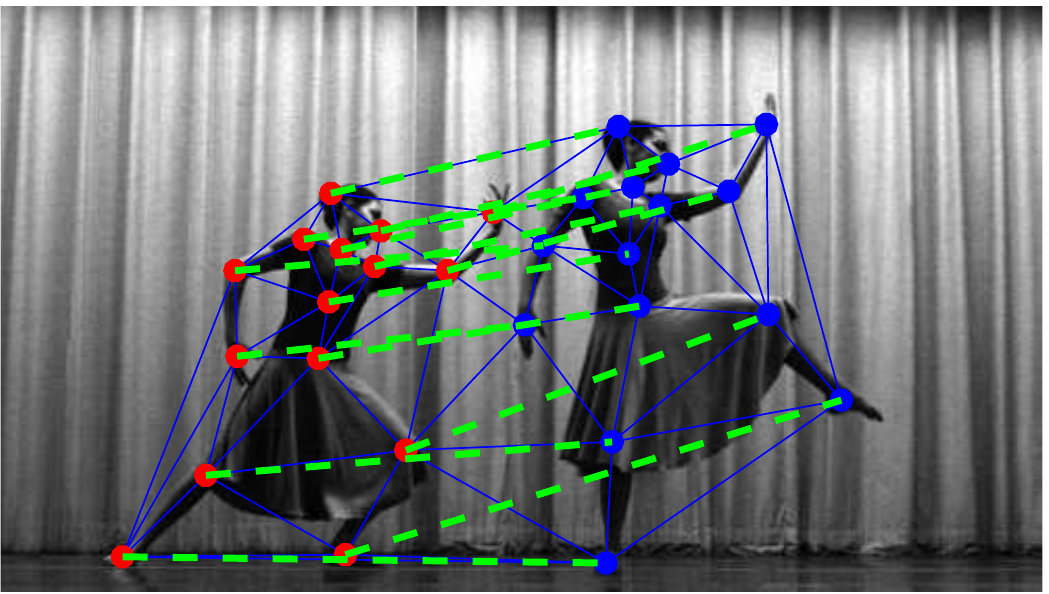}} \\\vspace{-5pt}
        \subfigure[Ground-truth: \textbf{18/18}]{\includegraphics[width=1\textwidth]{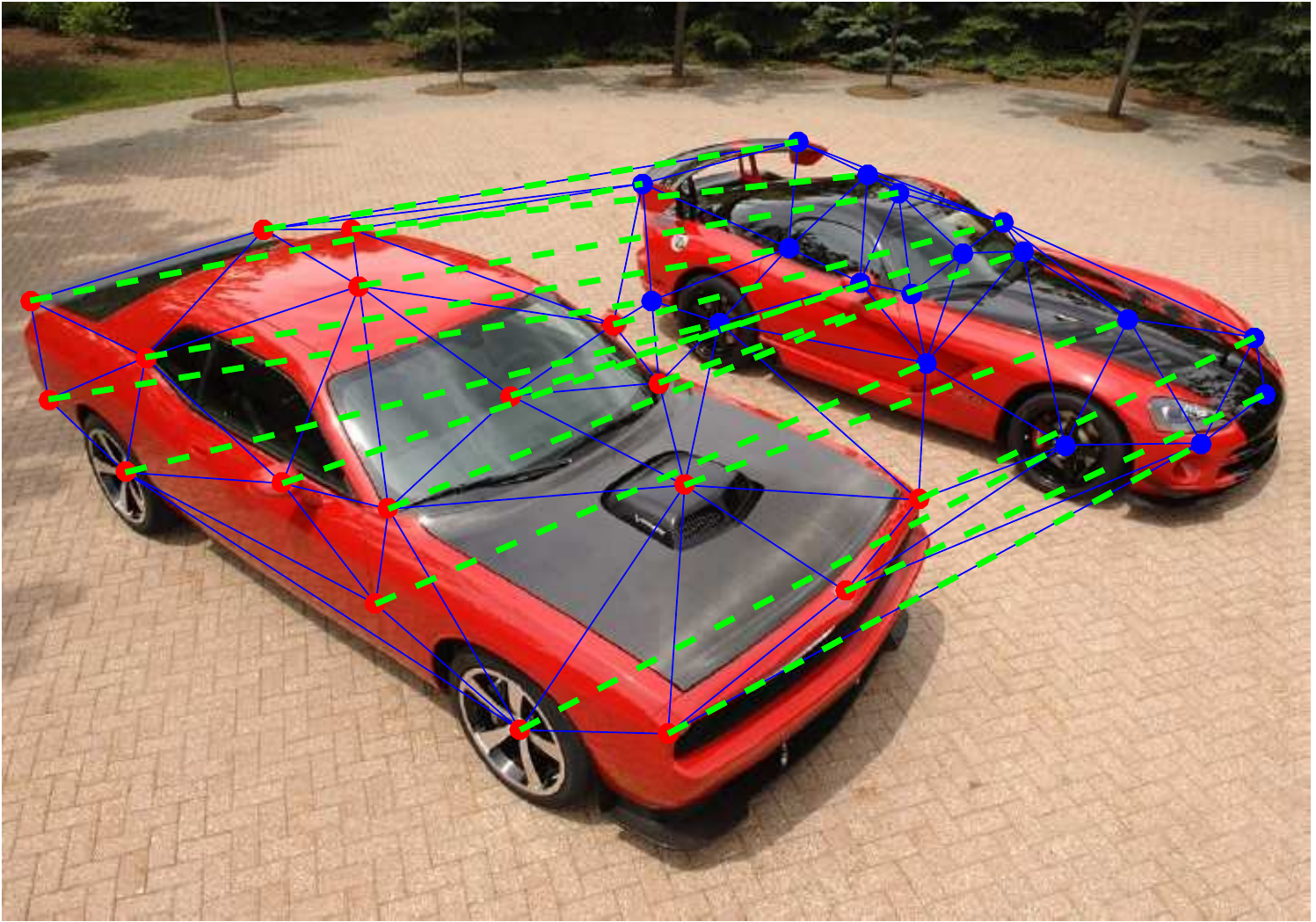}}
    \end{minipage}
}{
    \begin{minipage}{0.18\textwidth}
        \subfigure[Standard cuts: \textbf{16/20}]{\includegraphics[width=1\textwidth]{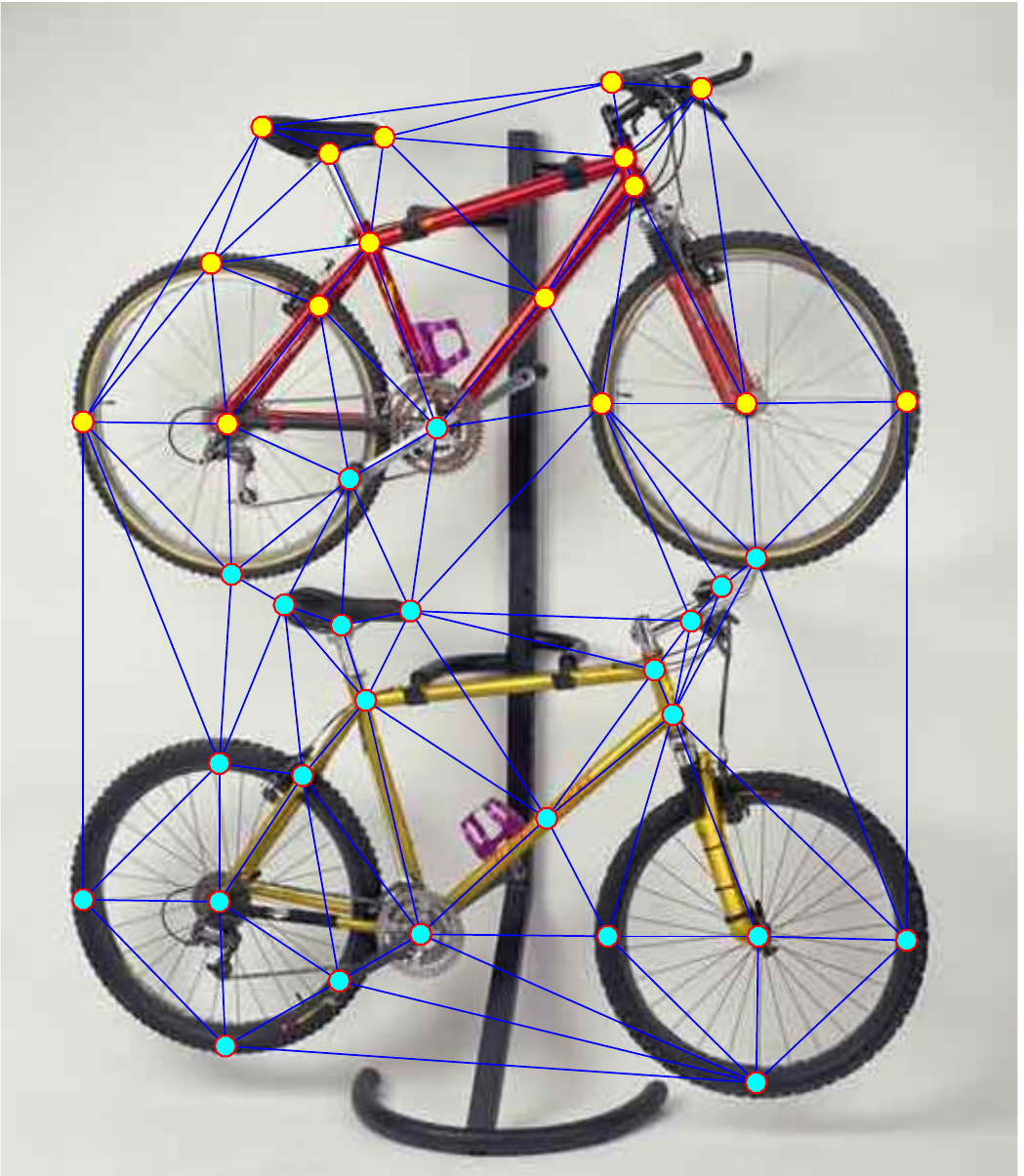}} \\\vspace{-5pt}
        \subfigure[Standard cuts: \textbf{21/25}]{\includegraphics[width=1\textwidth]{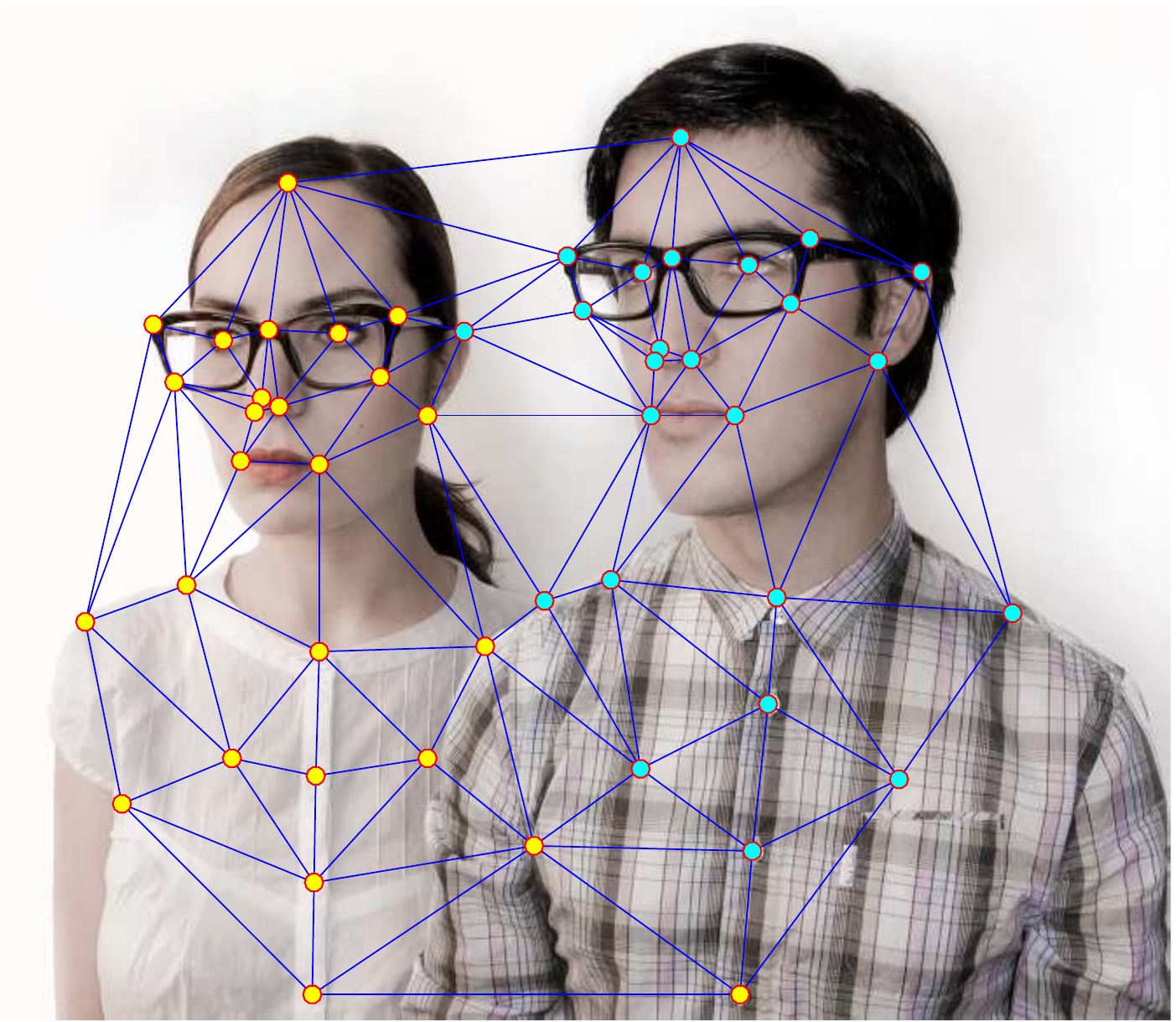}} \\\vspace{-5pt}
        \subfigure[Standard cuts: \textbf{13/15}]{\includegraphics[width=1\textwidth]{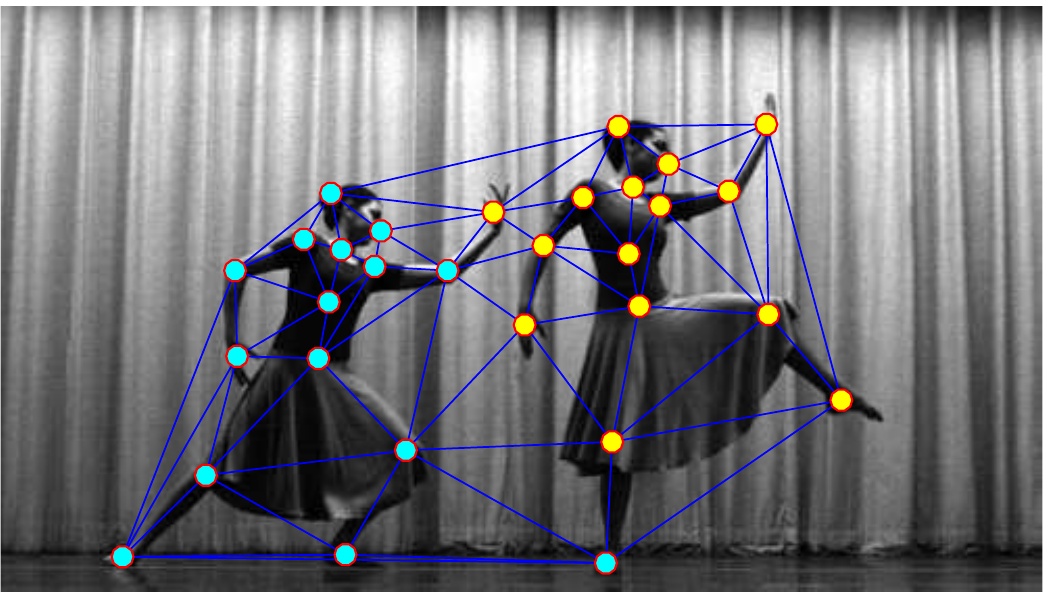}} \\\vspace{-5pt}
        \subfigure[Standard cuts: \textbf{15/18}]{\includegraphics[width=1\textwidth]{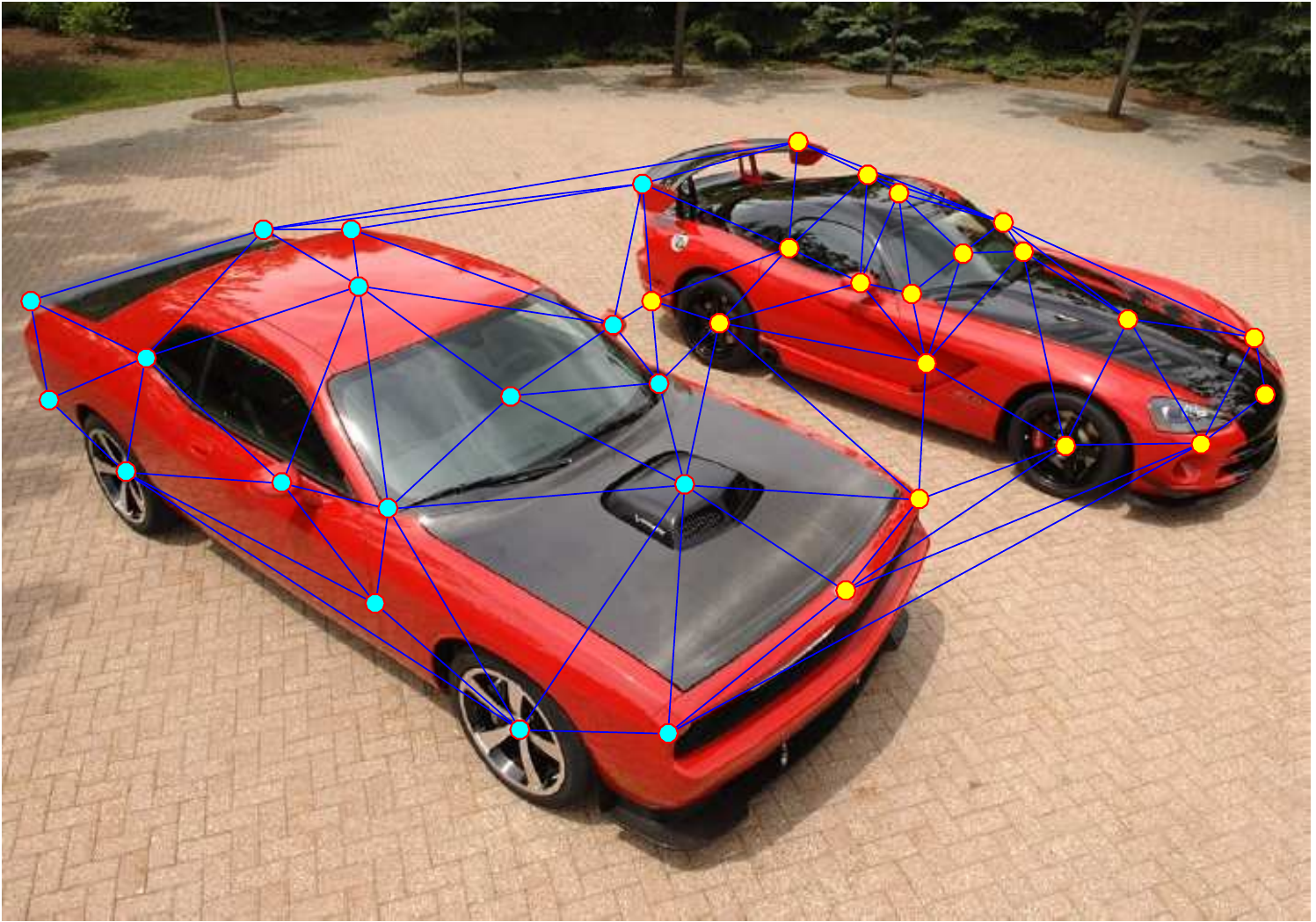}}
    \end{minipage}
}
{
    \begin{minipage}{0.18\textwidth}
        \subfigure[Match by IBGP: \textbf{3/20}]{\includegraphics[width=1\textwidth]{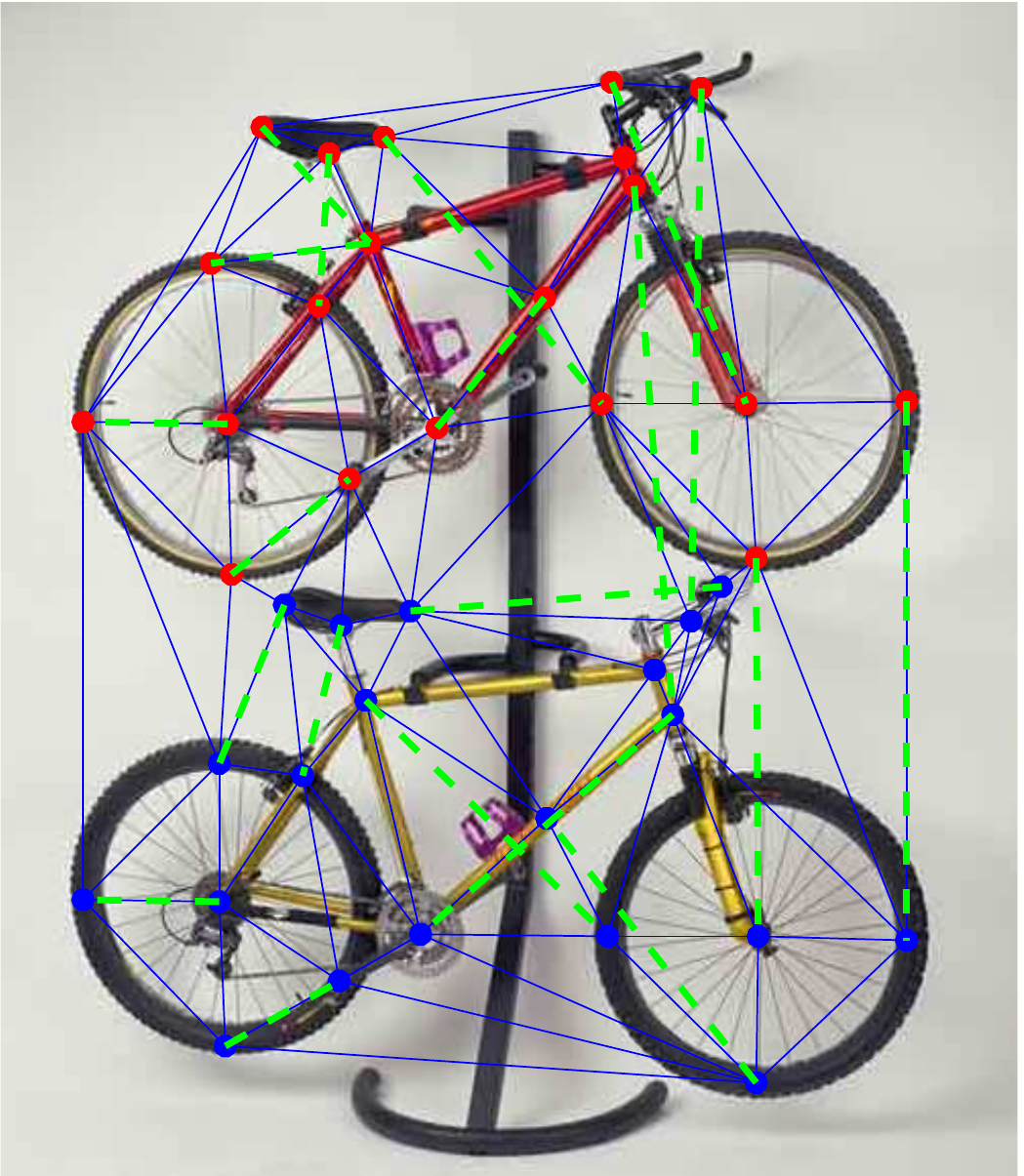}}\\\vspace{-5pt}
        \subfigure[Match by IBGP: \textbf{5/25}]{\includegraphics[width=1\textwidth]{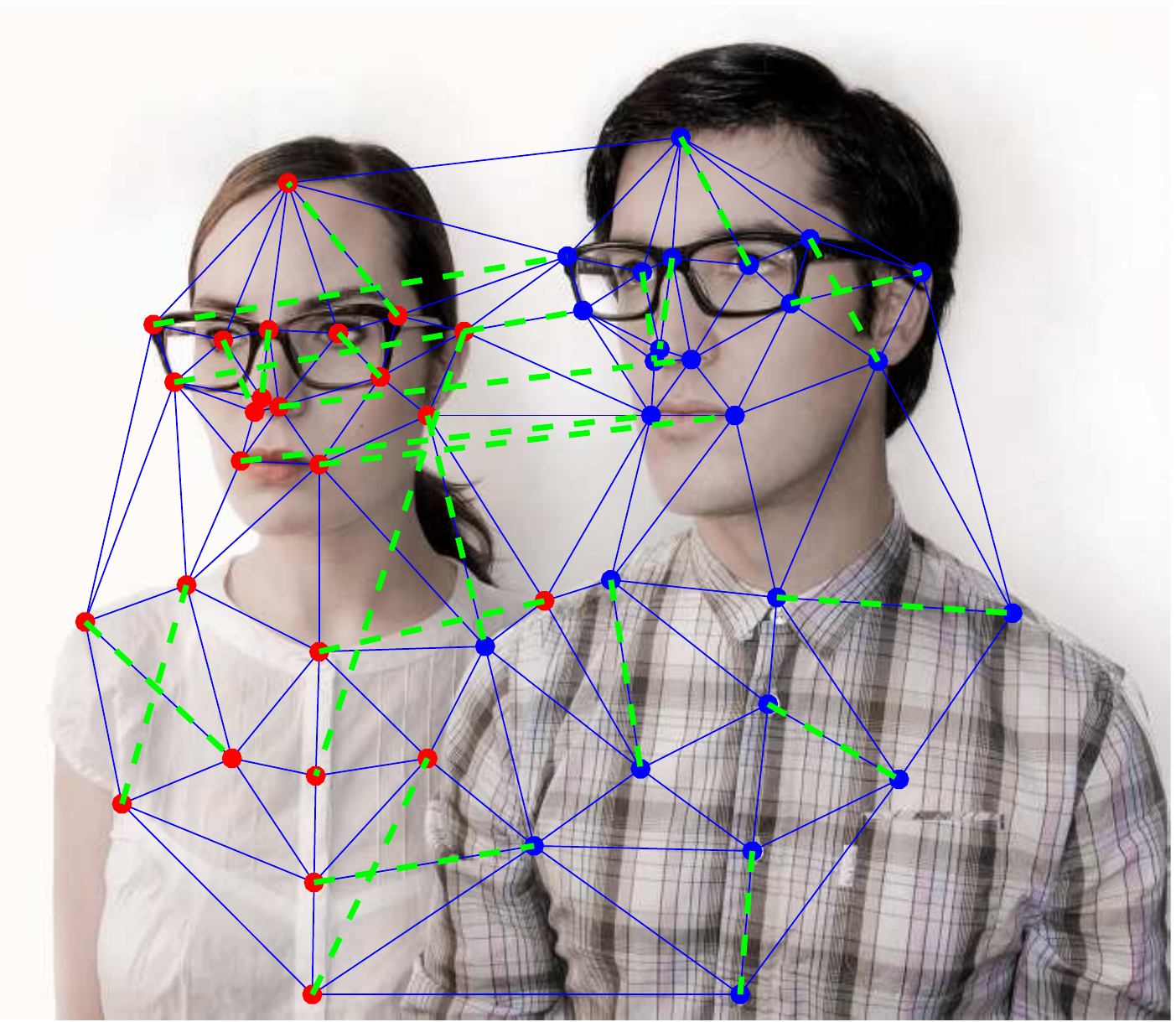}} \\\vspace{-5pt}
        \subfigure[Match by IBGP: \textbf{1/15}]{\includegraphics[width=1\textwidth]{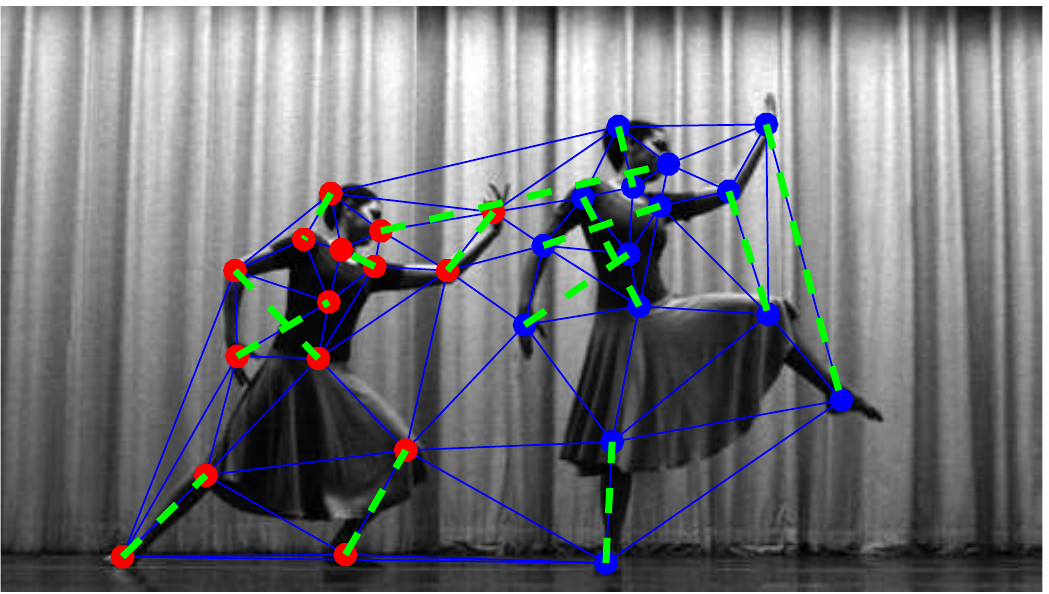}} \\\vspace{-5pt}
        \subfigure[Match by IBGP: \textbf{0/18}]{\includegraphics[width=1\textwidth]{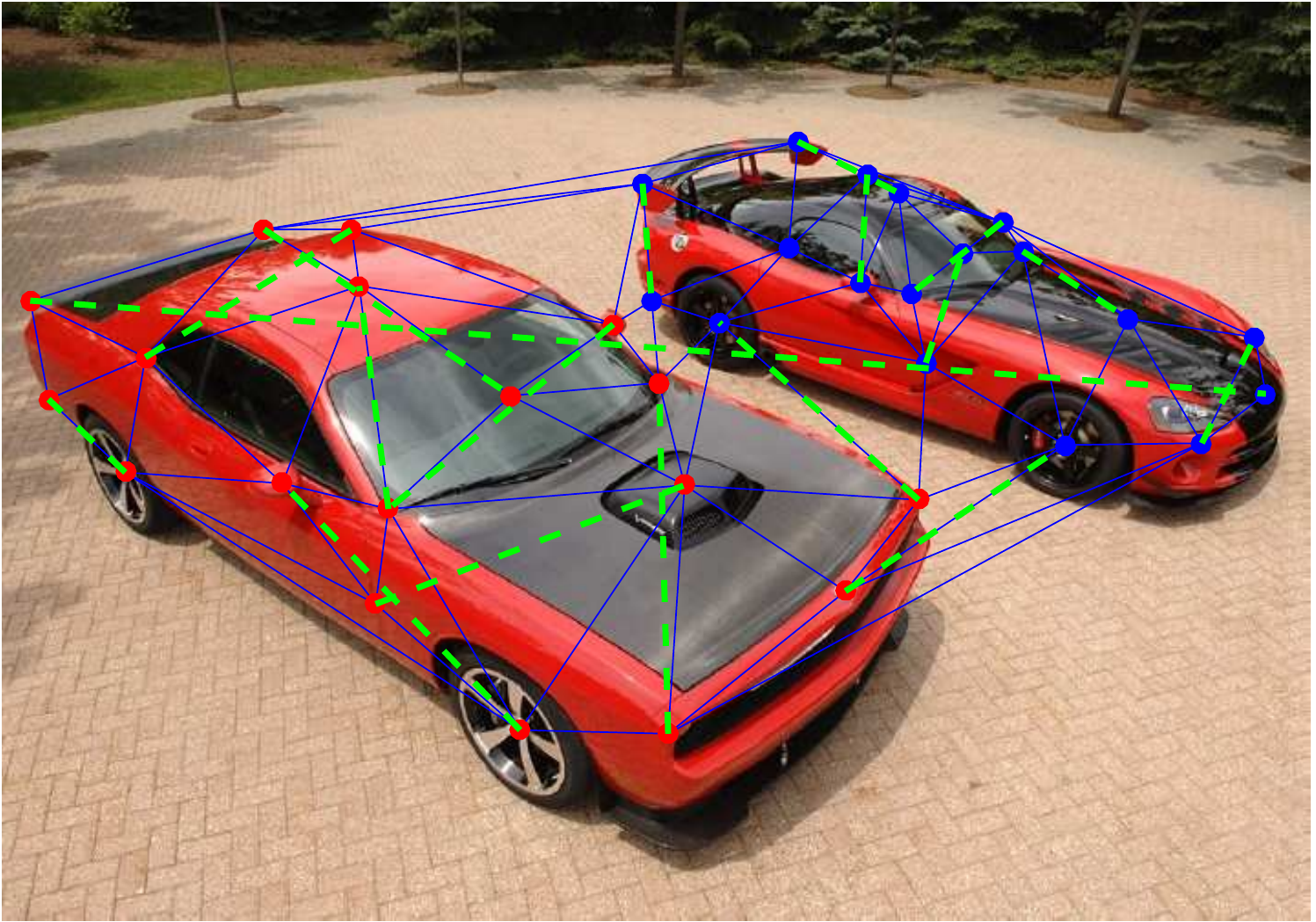}}
    \end{minipage}
}
{
    \begin{minipage}{0.18\textwidth}
        \subfigure[CutMatch-cuts: \textbf{19/20}]{\includegraphics[width=1\textwidth]{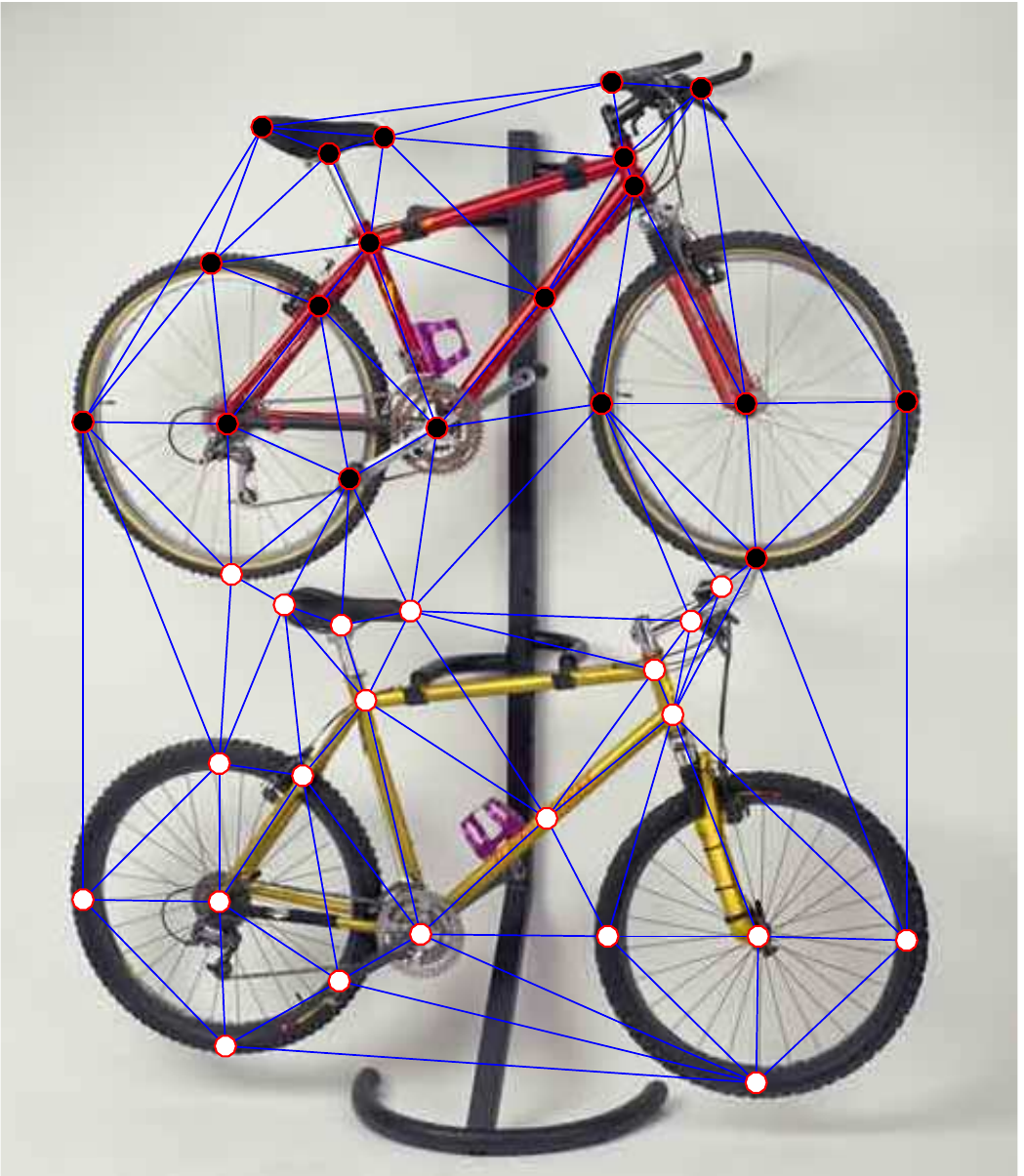}} \\\vspace{-5pt}
        \subfigure[CutMatch-cuts: \textbf{25/25}]{\includegraphics[width=1\textwidth]{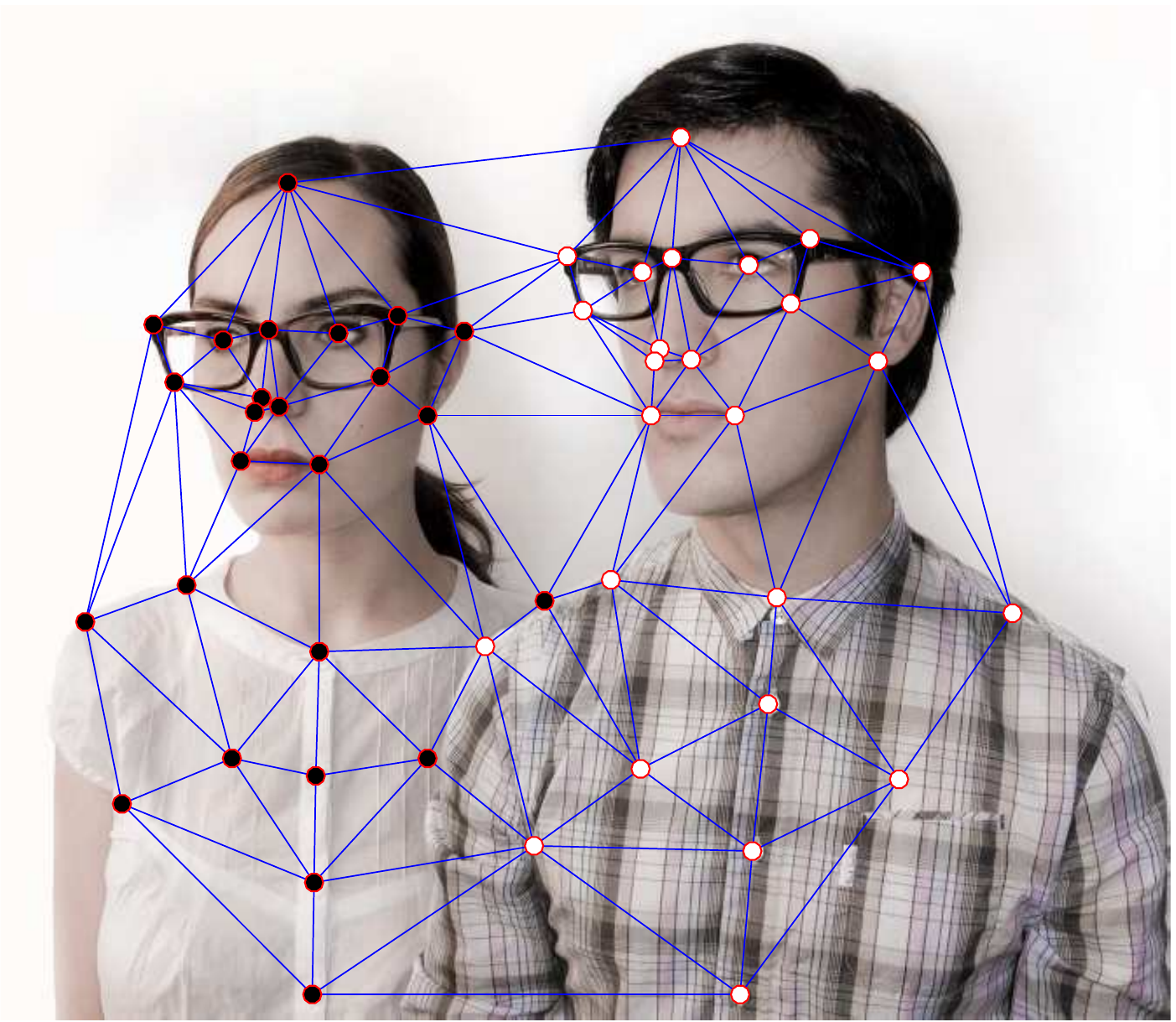}} \\\vspace{-5pt}
        \subfigure[CutMatch-cuts: \textbf{15/15}]{\includegraphics[width=1\textwidth]{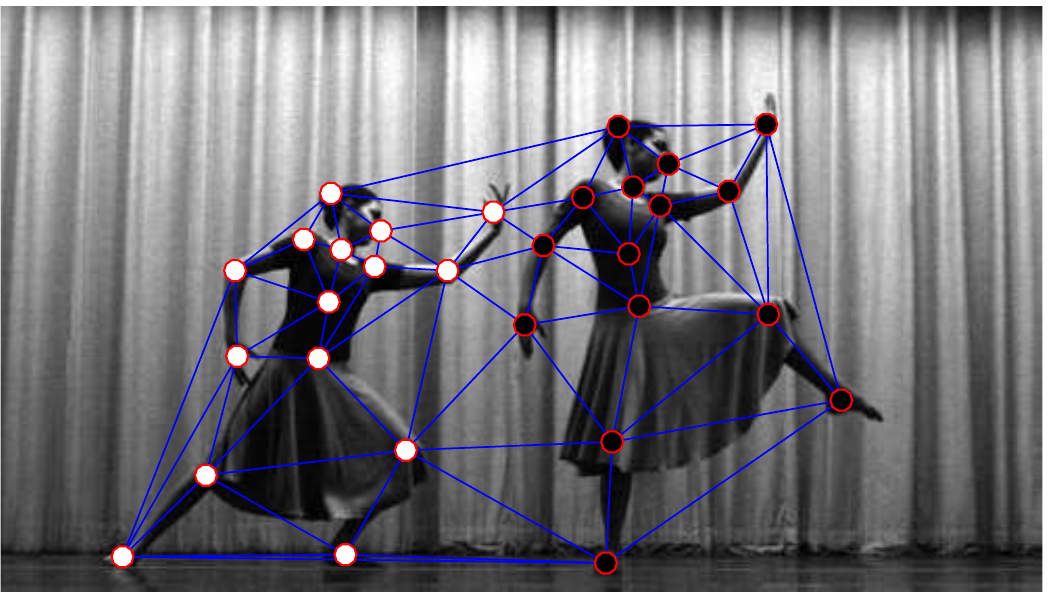}} \\\vspace{-5pt}
        \subfigure[CutMatch-cuts: \textbf{15/18}]{\includegraphics[width=1\textwidth]{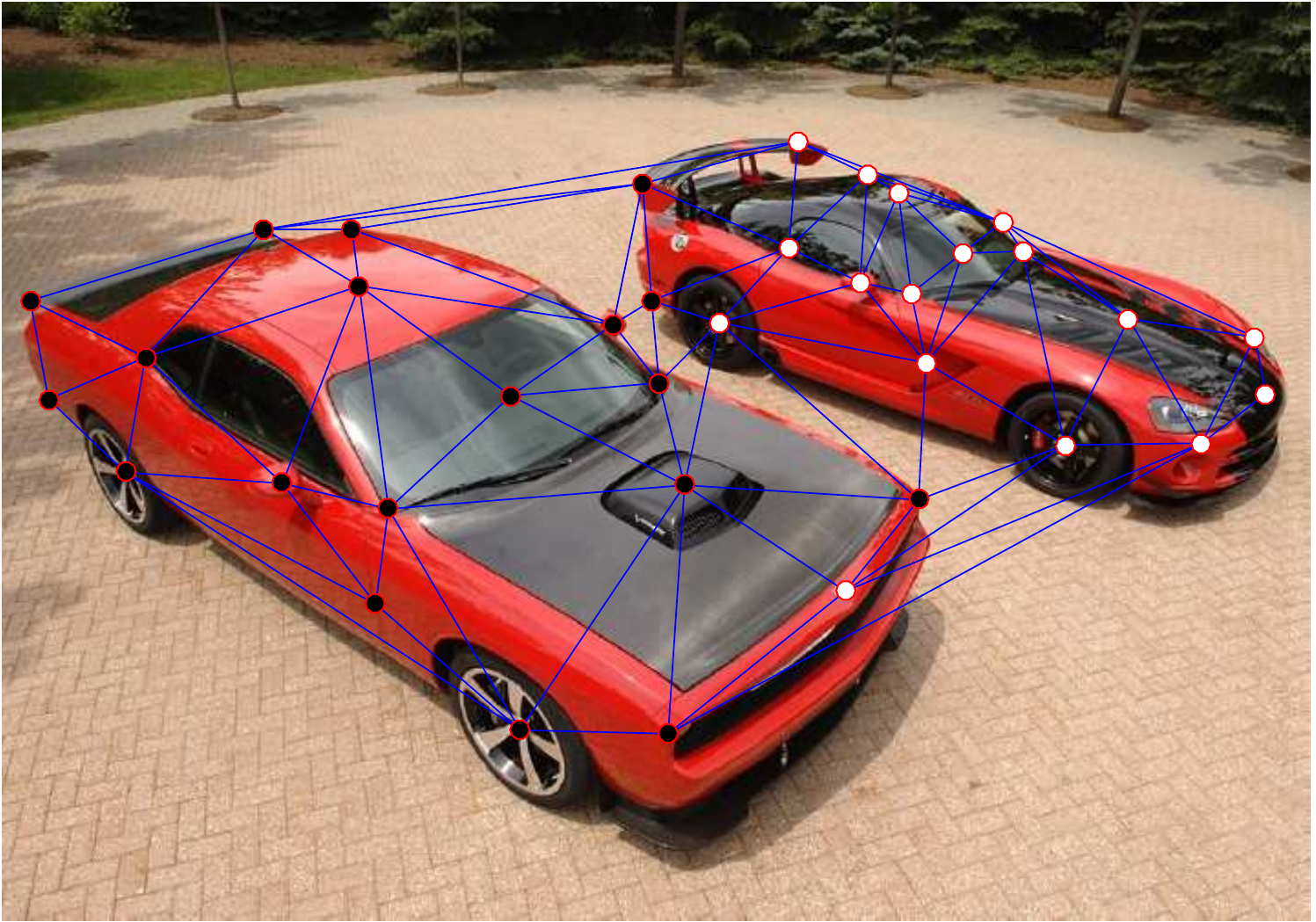}}
    \end{minipage}
    }
{
    \begin{minipage}{0.18\textwidth}
        \subfigure[CutMatch-match: \textbf{20/20}]{\includegraphics[width=1\textwidth]{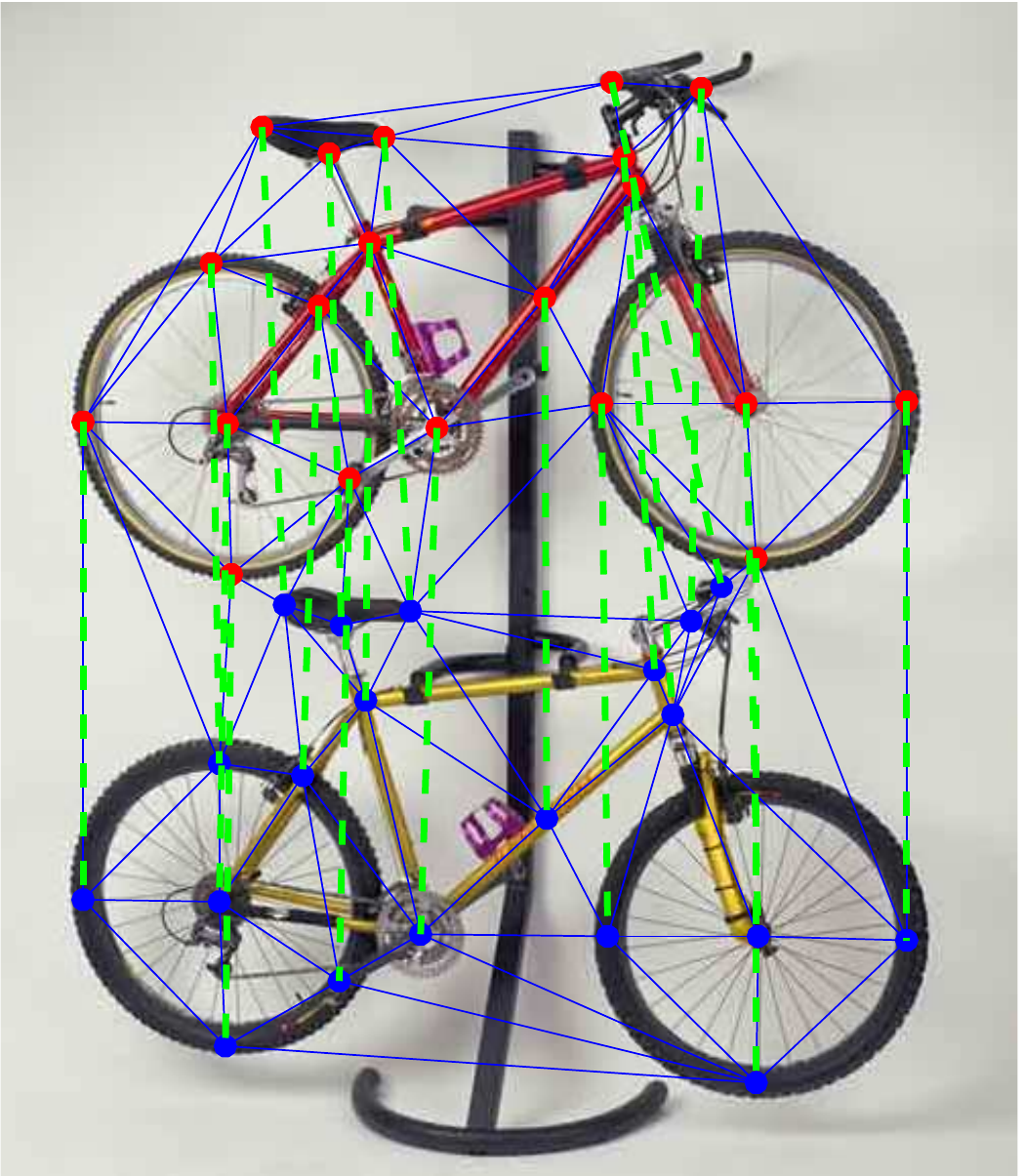}} \\\vspace{-5pt}
        \subfigure[CutMatch-match: \textbf{16/25}]{\includegraphics[width=1\textwidth]{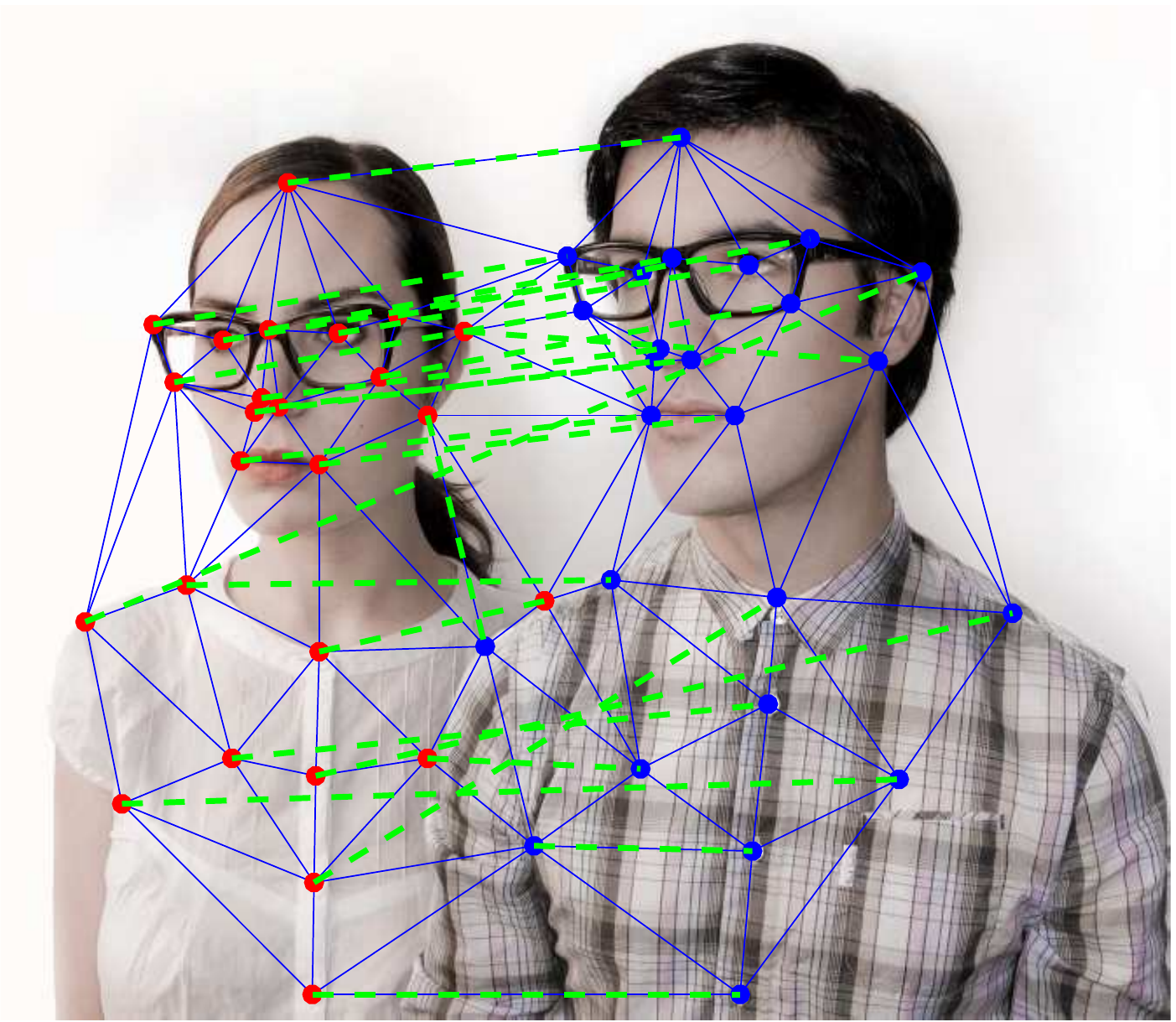}} \\\vspace{-5pt}
        \subfigure[CutMatch-match: \textbf{7/15}]{\includegraphics[width=1\textwidth]{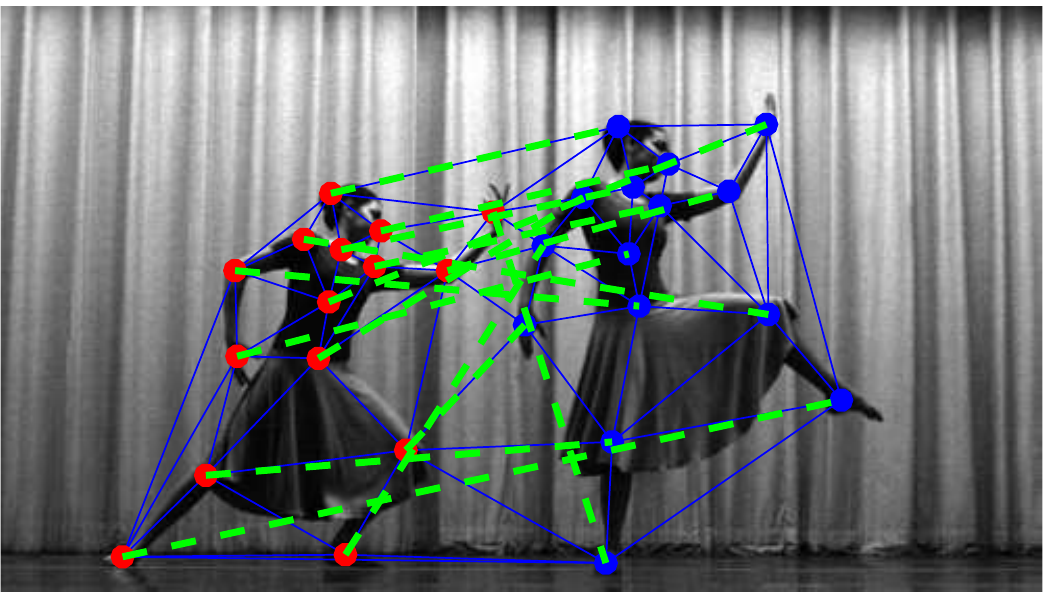}} \\\vspace{-5pt}
        \subfigure[CutMatch-match: \textbf{4/18}]{\includegraphics[width=1\textwidth]{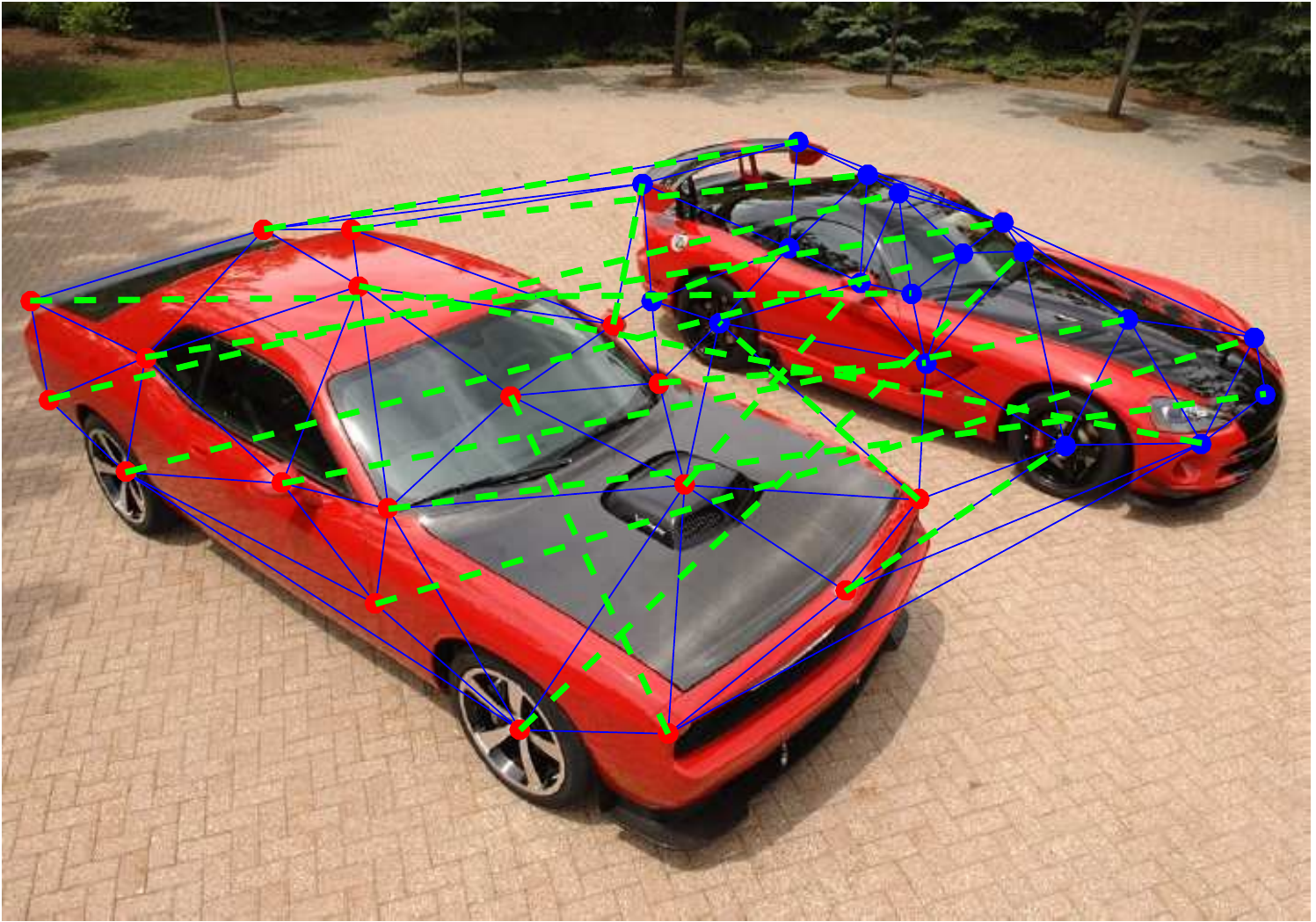}}
    \end{minipage}
}
\caption{Results on real-world image by joint CutMatch and vanilla graph cuts or graph matching alone.}
\label{fig:exp_image_results}
\end{figure*}

\textbf{Results} To this end, we randomly generate edges corresponding to varying edge density from $0$ to $0.2$ with step $0.02$. For each density value other than $0$, $10$ graphs are generated to avoid the bias, and the average performance is calculated regarding all the graphs. Fig. \ref{fig:image_result} demonstrates the cuts and matching results. IBGP, RRWM and raw cuts are selected one again for comparison. As CutMatch shows its parametric stability in synthetic test, we fix parameters $\lambda_1=200$ and $\lambda_2=50$. We also let $\delta_1=150$ and $\delta_2=400,000$ to generate affinity and flow in proper range. As can be seen from Fig. \ref{fig:image_result}, though the task on real images is challenging, CutMatch is superior to all the selected counterparts in terms of cuts and matching accuracy. In general, matching partitions is relatively more difficult than cuts, thus needs more concentration in our future work. Fig. \ref{fig:exp_image_results} shows some results on real-world images. The cuts performance is observed to be enhanced by using CutMatch. When the connectivity is stable, e.g. in the first row (for rigid bike), the matching with CutMatch is reliable and very close to the ground-truth. When objects contain salient partial deformation, e.g. in the second and the third row (for deformable human body), CutMatch reaches significant matching performance compared to vanilla IBGP for graph matching. However, when disturbance is severe as in the last row (more large transformation), the matching will mostly fail. In either case, the performance of raw IBGP matching is extremely low.

\section{Conclusion and Future Work}
This paper presents a solver to the joint graph cuts and matching problem on a single input graph, in order to better account for both the closeness and correspondences in the presence of two similar objects appearing in one image. Despite the illustrated potential of CutMatch, there is space for future work: i) CutMatch can possibly trap into a local optima. This issue can be possibly addressed with annealing strategies; ii) the large amount of variables and gradient-based update scheme involved in MatchCut makes it currently not scalable for dealing with dense correspondences. We empirically observe that CutMatch tends to deliver sparse solution $\mathbf{X}$, hence we believe MatchCut can be accelerated especially considering affinity $\mathbf{A}$ is also sparse; iii) the generalized case: cut graph into $k$ partitions and establish their correspondence.

\section*{Appendix}
\subsection*{Update of gradient for GM}
\begin{figure*}
\begin{center}
    \includegraphics[width = 0.24\textwidth]{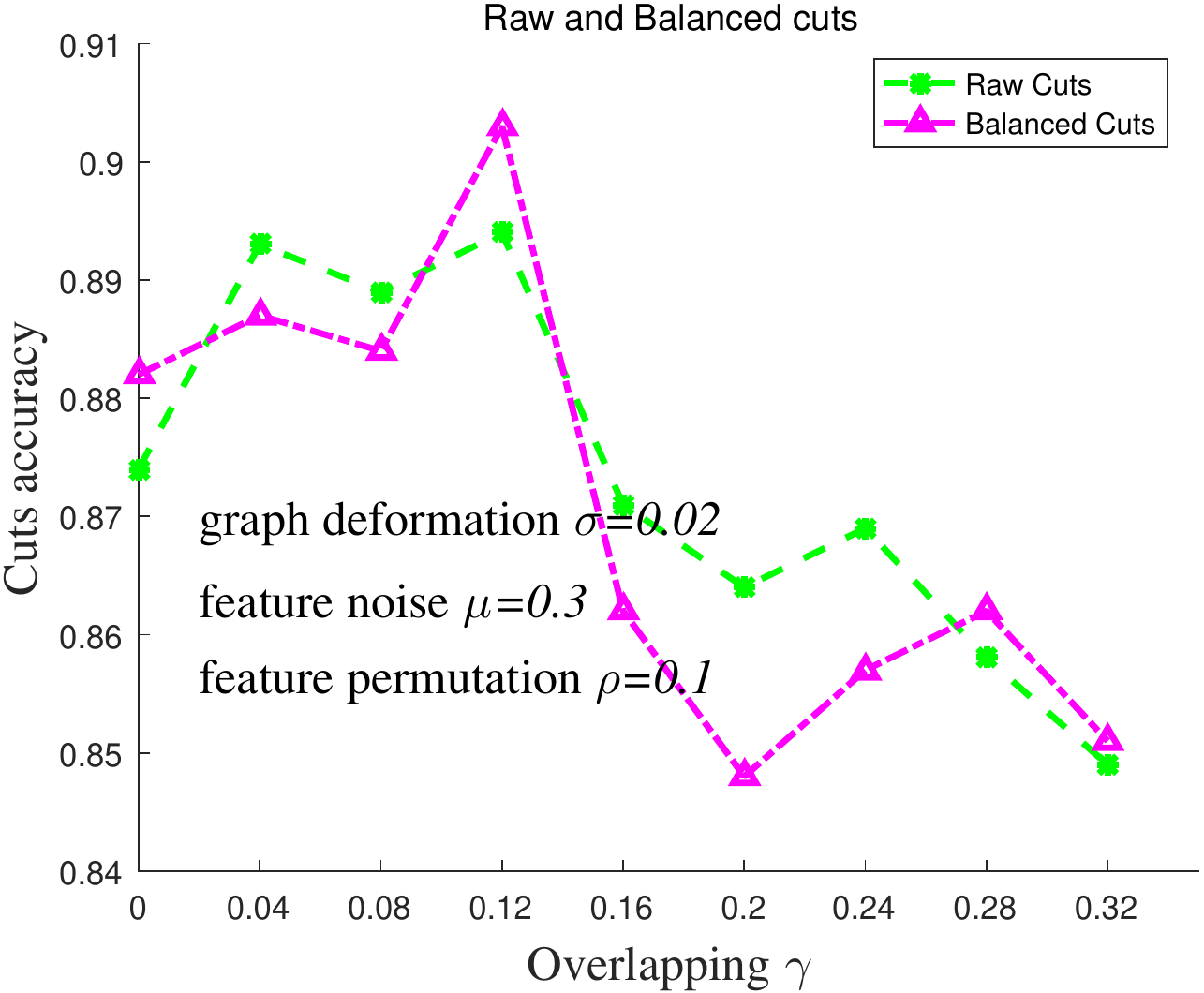}
    \includegraphics[width = 0.24\textwidth]{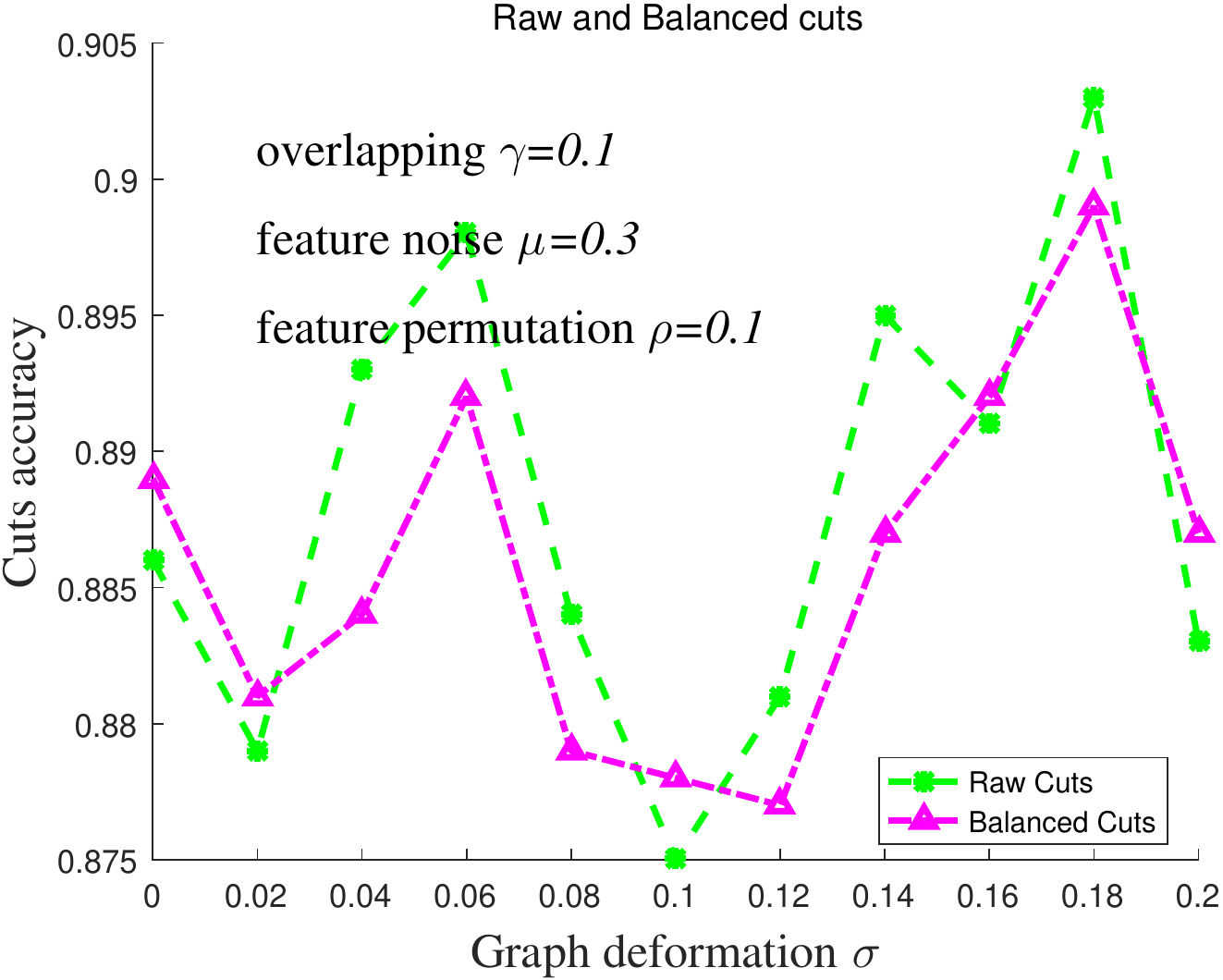}
    \includegraphics[width = 0.24\textwidth]{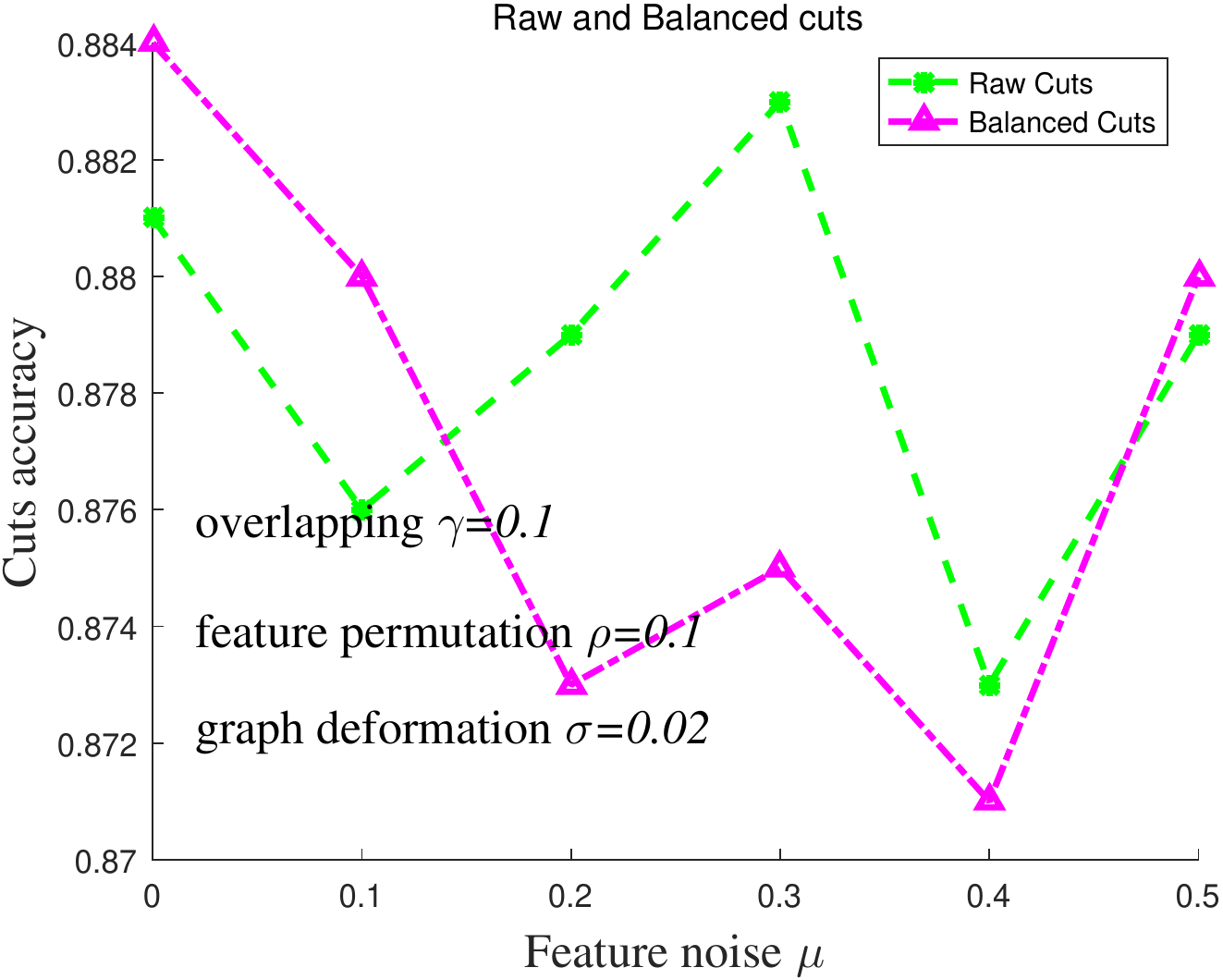}
    \includegraphics[width = 0.24\textwidth]{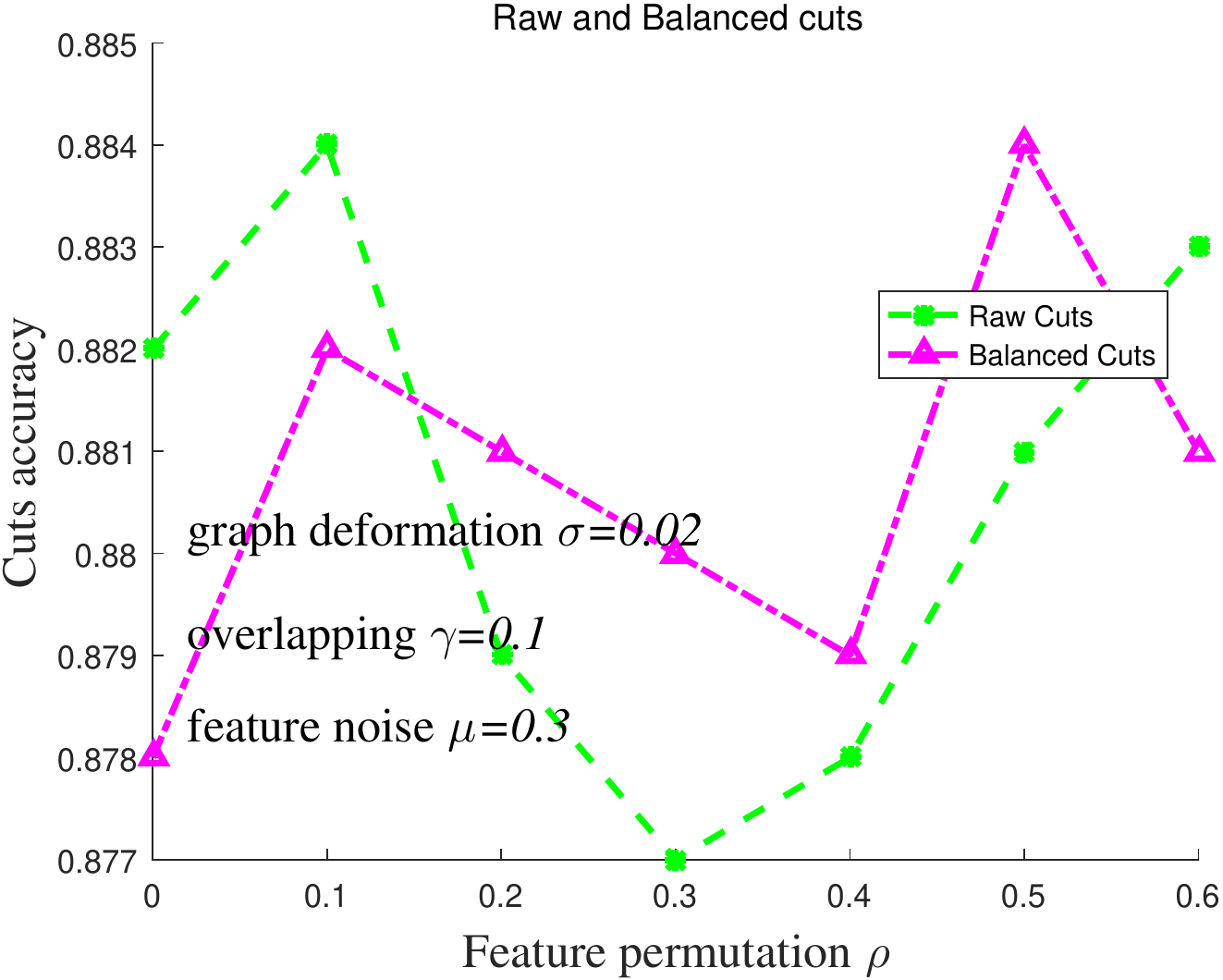}
\end{center}
\vspace{-10pt}
\caption{Performance of raw cuts and balanced cuts.}
\label{fig:balanced_cut}
\end{figure*}
For standard graph matching, we seek to solve:
\begin{equation}
\begin{split}
\min_{\mathbf{V}}&\lVert\mathbf{V}-\mathbf{U}\rVert_F^2 \\
\text{s.t. }&\mathbf{V}\mathbf{1}=\mathbf{0},\mathbf{V}^T\mathbf{1}=\mathbf{0}
\end{split}
\end{equation}

\noindent where $\mathbf{V}$ and $\mathbf{U}$ are the optimal update direction and the objective gradient, respectively. Refer to the main body for the CutMatch version. Similar to the previous procedure, the Lagrangian function is:
\begin{equation}
\mathcal{L}=\lVert\mathbf{V}-\mathbf{U}\rVert_F^2-\mathbf{a}^T\mathbf{V}\mathbf{1}-\mathbf{b}^T\mathbf{V}^T\mathbf{1}
\end{equation}

Zeroing the partial derivative w.r.t. $\mathbf{V}$ we have:
\begin{equation}
\frac{\partial\mathcal{L}}{\partial\mathbf{V}}=2(\mathbf{V}-\mathbf{U})-\mathbf{a}\mathbf{1}^T-\mathbf{1}\mathbf{b}^T=\mathbf{O}
\end{equation}

Then we have:
\begin{equation}
\label{eq:sup_V}
\mathbf{V}=\mathbf{U}+\frac{1}{2}\mathbf{a}\mathbf{1}^T+\frac{1}{2}\mathbf{1}\mathbf{b}^T
\end{equation}

Right multiplying $\mathbf{1}$, we have:
\begin{equation}
\label{eq:gm_1}
\begin{split}
\mathbf{0}&=\mathbf{U}\mathbf{1}+\frac{1}{2}\mathbf{a}\mathbf{1}^T\mathbf{1}+\frac{1}{2}\mathbf{1}\mathbf{b}\mathbf{1}^T \\
&=\mathbf{U}\mathbf{1}+\frac{n}{2}\mathbf{a}+\frac{1}{2}\mathbf{1}\mathbf{1}^T\mathbf{b}
\end{split}
\end{equation}

Transposing Eq. (\ref{eq:sup_V}) and right multiplying $\mathbf{1}$ we have:
\begin{equation}
\label{eq:gm_2}
\begin{split}
\mathbf{0}=\mathbf{U}^T\mathbf{1}+\frac{1}{2}\mathbf{1}\mathbf{1}^T\mathbf{a}+\frac{n}{2}\mathbf{b}
\end{split}
\end{equation}

Then we have:
\begin{align}
&\frac{\mathbf{1}\mathbf{1}^T\cdot\text{Eq.\ref{eq:gm_1}}-\text{\ref{eq:gm_2}}}{n}\\
=&\frac{\mathbf{1}\mathbf{1}^T\mathbf{U}\mathbf{1}}{n}-\mathbf{U}^T\mathbf{1}+\frac{\mathbf{1}\mathbf{1}^T\mathbf{1}\mathbf{1}^T\mathbf{b}}{2n}-\frac{n\mathbf{b}}{2}\\
=&\mathbf{0}
\end{align}

Then,
\begin{equation}
\frac{1}{2}(\mathbf{1}\mathbf{1}^T-n\mathbf{I})\mathbf{b}=\mathbf{U}^T\mathbf{1}-\frac{1}{n}\mathbf{1}\mathbf{1}^T\mathbf{U}\mathbf{1}
\end{equation}

As we have the pseudo inverse (see next section):
\begin{equation}
(n\mathbf{I}-\mathbf{1}\mathbf{1}^T)^{-1}=\frac{1}{n}(\mathbf{I}-\frac{1}{n}\mathbf{1}\mathbf{1}^T)
\end{equation}

We then have:
\begin{equation}
\label{eq:sup_b}
\mathbf{b}=-\frac{2}{n}(\mathbf{U}^T\mathbf{1}-\frac{1}{n}\mathbf{1}\mathbf{1}^T\mathbf{U}^T\mathbf{1})
\end{equation}

Analogously, we can obtain:
\begin{equation}
\label{eq:sup_a}
\mathbf{a}=-\frac{2}{n}(\mathbf{U}\mathbf{1}-\frac{1}{n}\mathbf{1}\mathbf{1}^T\mathbf{U}\mathbf{1})
\end{equation}

Substituting Eq. (\ref{eq:sup_b}) and (\ref{eq:sup_a}) back to (\ref{eq:sup_V}), we obtain:
\begin{equation}
\mathbf{V}=\mathbf{U}-\frac{1}{n}\mathbf{U}\mathbf{1}\mathbf{1}^T-\frac{1}{n}\mathbf{1}\mathbf{1}^T\mathbf{U}+\frac{2}{n^2}\mathbf{1}\mathbf{1}^T\mathbf{U}\mathbf{1}\mathbf{1}^T
\end{equation}
which is the same form as the update rule in the main text.
\subsection*{Derivation of pseudo inverse}
Consider the matrix $\mathbf{K}=\mathbf{I}-\frac{1}{n}\mathbf{1}\mathbf{1}^T$, and the corresponding SVD is $\mathbf{K}=\mathbf{U}\mathbf{\Sigma}\mathbf{V}^T$. As $\mathbf{K}$ is symmetric, we have $\mathbf{U}=\mathbf{V}$. Note that $\mathbf{K}$ corresponds to the Laplacian of a complete graph with $n$ nodes, divided by $n$. Recall that Laplacian of a complete graph must have one $0$ eigenvalue, and $n$ for all the remaining ones. Thus the scaled matrix $\mathbf{K}$ must have one $0$ eigenvalue, and $1$ for the rest. Denoting $\text{pinv}(\cdot)$ the psuedo inverse function, we have:

\begin{equation}
\begin{split}
\text{pinv}(\mathbf{K})&=\text{pinv}(\mathbf{I}-\frac{1}{n}\mathbf{1}\mathbf{1}^T) \\
                       &=\text{pinv}\left(\mathbf{U}\left[
                                                      \begin{array}{cc}
                                                        \mathbf{I} & \mathbf{0} \\
                                                        \mathbf{0}^T & 0 \\
                                                      \end{array}
                                                    \right]
                       \mathbf{U}^T\right) \\
                       &=\left(\mathbf{U}\left[
                                                      \begin{array}{cc}
                                                        \mathbf{I}^{-1} & \mathbf{0} \\
                                                        \mathbf{0}^T & 0 \\
                                                      \end{array}
                                                    \right]
                       \mathbf{U}^T\right) \\
                       &=\left(\mathbf{U}\left[
                                                      \begin{array}{cc}
                                                        \mathbf{I} & \mathbf{0} \\
                                                        \mathbf{0}^T & 0 \\
                                                      \end{array}
                                                    \right]
                       \mathbf{U}^T\right) \\
                       &=\mathbf{K}
\end{split}
\end{equation}
which gives the analytical form of the inverse.

\subsection*{Supplementary experiment on balanced cuts}
We present performance of balanced cuts on synthetic graphs. To enforce each partition has the same size, we calculate the balanced cuts as follows. Given the graph Laplacian $\mathbf{L}^g$, we first obtain the optimal solution to the Rayleigh problem (the same as in the main body):

\begin{equation}
\mathbf{y}^*=\arg\min_{\mathbf{y}}\frac{\mathbf{y}^T\mathbf{L}^g\mathbf{y}}{\mathbf{y}^T\mathbf{y}}
\end{equation}

Then for each element of vector $\mathbf{y}^*$, we find its median value $\mathbf{y}_m$. For an element $\mathbf{y}_i$, we category node $i$ to the first partition if $\mathbf{y}_i\geq\mathbf{y}_m$, and to the second partition if $\mathbf{y}_i<\mathbf{y}_m$. For comparison, we only additionally present the performance of raw graph cuts which is sufficient to illustrate that balanced cuts is no more superior than raw cuts in our settings. In synthetic test, we generate $50$ graphs for each parameter value combination, and calculate the average cuts performance. All the other setting follow the ones from the manuscript. Fig \ref{fig:balanced_cut} summarizes the experimental result, and we can observe that the performance of balanced cuts is very similar to that of raw cuts. Note the range of the accuracy on the vertical axis.
{\small
\bibliographystyle{ieee}
\bibliography{egbib}
}
\end{document}